\RequirePackage{tikz}
\documentclass[final,3p,times,twocolumn]{elsarticle}

\usepackage{hyperref}
\usepackage{tabularx}
\usepackage{amsmath}
\usepackage{multirow}
\usepackage{booktabs}
\usepackage{rotating}
\usepackage{lipsum,adjustbox}
\usetikzlibrary{trees}
\usepackage{dcolumn}

\usepackage{amssymb}
\newcolumntype{d}[1]{D{.}{.}{#1}}
\newcommand{\mc}[1]{\multicolumn{1}{c}{#1}}

\makeatletter
\def\ps@pprintTitle{%
	\let\@oddhead\@empty
	\let\@evenhead\@empty
	\let\@oddfoot\@empty
	\let\@evenfoot\@oddfoot
}
\makeatother

\begin{document}

\begin{frontmatter}



\title{Transformers in  Single Object Tracking: An Experimental Survey}


\author[1,2]{Janani Thangavel\corref{cor1}\fnref{fn1}}
\ead{jananitha@univ.jfn.ac.lk}

\author[3]{Thanikasalam Kokul\corref{cor1}\fnref{fn1}}
\ead{kokul@univ.jfn.ac.lk}

\author[3]{Amirthalingam Ramanan}
\ead{a.ramanan@univ.jfn.ac.lk}

\author[1]{Subha Fernando}
\ead{subhaf@uom.lk}

\cortext[cor1]{Corresponding author}
\fntext[fn1]{These authors contributed equally to this work.}

\affiliation[1]{organization={Department of Computational Mathematics, University of Moratuwa},
	country={Sri Lanka}}
\affiliation[2]{organization={Department of Interdisciplinary Studies, University of Jaffna},
	country={Sri Lanka}}
\affiliation[3]{organization={Department of Computer Science, University of Jaffna},
	country={Sri Lanka}}

\begin{abstract}
	Single-object tracking is a well-known and challenging research topic in computer vision. Over the last two decades, numerous researchers have proposed various algorithms to solve this problem and achieved promising results. Recently, Transformer-based tracking approaches have ushered in a new era in single-object tracking by introducing new perspectives and achieving superior tracking robustness. In this paper, we conduct an in-depth literature analysis of Transformer tracking approaches by categorizing them into CNN-Transformer based trackers, Two-stream Two-stage fully-Transformer based trackers, and One-stream One-stage fully-Transformer based trackers.  In addition, we conduct experimental evaluations to assess their tracking robustness and computational efficiency using publicly available benchmark datasets. Furthermore, we measure their performances on different tracking scenarios to identify their strengths and weaknesses in particular situations. Our survey provides insights into the underlying principles of Transformer tracking approaches, the challenges they encounter, and the future directions they may take.
\end{abstract}


\end{frontmatter}


\section{Introduction}\label{sec1}
Visual object tracking (VOT)  algorithms are intended to estimate the state (i.e., spatial location and size) of an object in a given video sequence. Given the initial state of the target in the first frame of a video sequence, these algorithms tracks the target's states in the remaining frames. VOT can be classified as single and multi-object tracking \cite{ciaparrone2020deep,kalake2021analysis} approaches based on the number of targets to be tracked. In single-object tracking, a single instance of an object class is tracked, while in multi-object tracking, multiple instances of an object class are tracked. 

Single object tracking  algorithms became popular and gained interest in recent years  because of its wide range of applications in computer vision, including video surveillance \cite{bib1}, augmented reality \cite{bib2}, automated driving \cite{bib3}, mobile robotics \cite{bib4}, traffic monitoring \cite{bib5}, sports video analysis \cite{bib6}, scene understanding \cite{bib7}, and human computer interaction \cite{bib8}. Single object VOT approaches captures the target's appearance features in the first frame of a video sequence and then use it to locate the target in the remaining frames. Although many appearance-based approaches have been proposed in VOT, it is still far from reaching the tracking robustness of humans with real-time speed  due to many challenges such as appearance and pose variations, occlusions, motion blurring, background clutter, similar object distractors, and deformation.

\begin{figure*}[t]%
	\centering
	\includegraphics[width=0.98\textwidth,height=8cm]{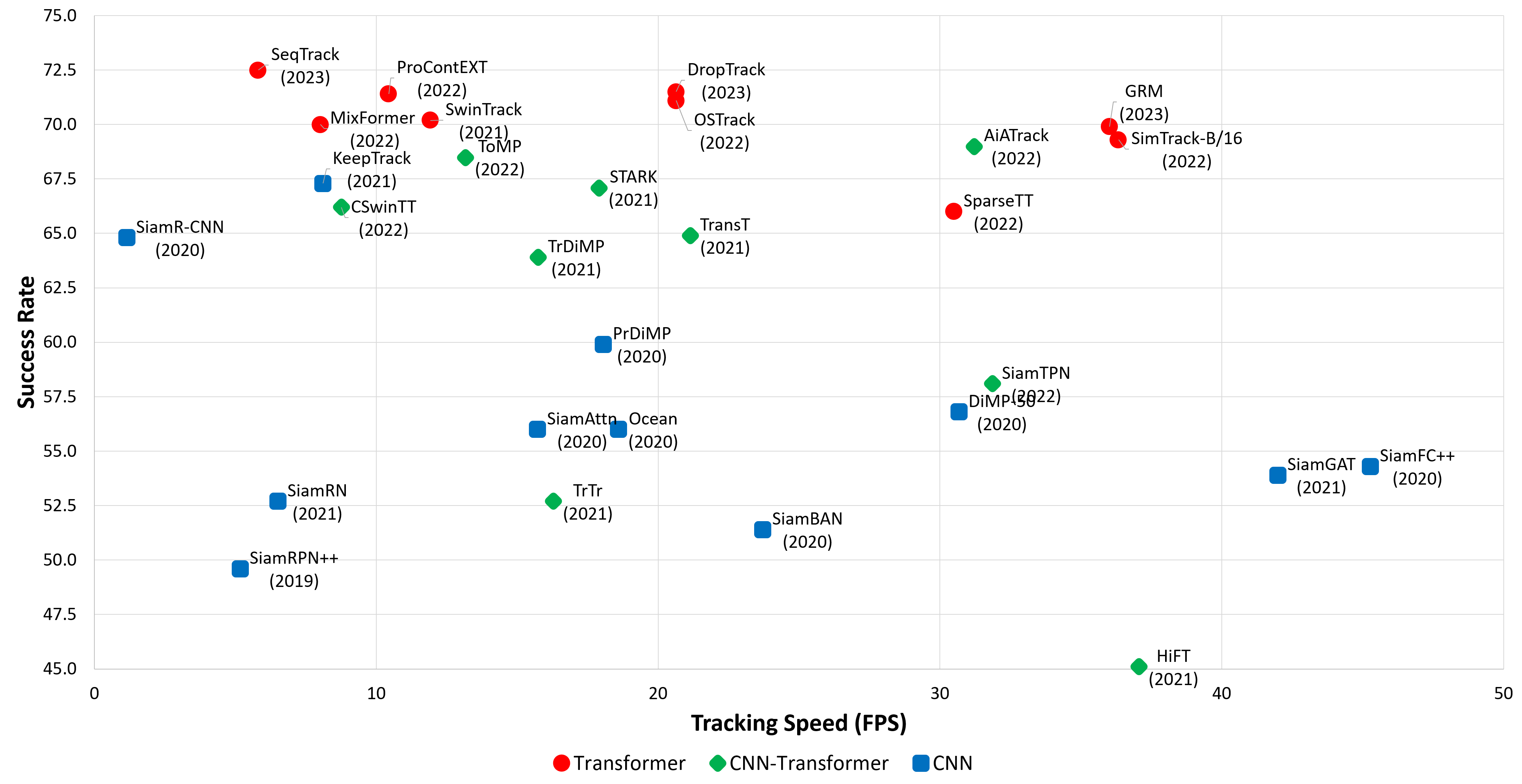}
	\caption{Performance comparison of state-of-the-art CNN-based, CNN-Transformer based, and fully-Transformer based trackers on LaSOT benchmark dataset. For some trackers, their reported tracking speed may be  different than the above graph since this  comparison study is conducted on a NVIDIA Quadro P4000 GPU with a 64GB of RAM. }
	\label{fig1:Comparison}
\end{figure*}

Single object tracking algorithms can be classified and analysed in several ways. Based on the features used in a tracking model, VOT approaches can be categorized as  hand-crafted and deep feature-based trackers.  Hand-crafted feature-based tracking approaches \cite{bib9,bib10,bib11, li2013survey, online2015} extract the features from images according to a certain manually predefined algorithm based on expert knowledge.  On the other hand, deep feature-based trackers \cite{zhu2021single,bib13,zhao2021deep, gate2017} capture the semantic cues from raw images by using the Convolutional Neural Networks (CNNs) \cite{bib15}.  Because of the hierarchical learning capability, deep features-based appearance trackers significantly outperform hand-crafted feature-based trackers \cite{bib40}. Based on how trackers differentiate  the target object from its surroundings, they can be discriminative and generative trackers. Discriminative trackers \cite{walid2021real,xu2021adaptive} treats the VOT as a binary classification task and separate the target object from the background.  On the other hand, most of the generative trackers \cite{bib19,bib20} treat VOT as a similarity matching problem by searching for the best candidate that closely matches the reference template in each frame. In recent trackers, a two branch CNN architecture, known as Siamese network \cite{bib21}, is utilized in similarity matching tracking. Over the past few years, a large number of Siamese-based trackers \cite{bib22,bib23,li2019siamrpn++,bib25,chen2020siamese,voigtlaender2020siam,guo2021graph,cheng2021learning} have been proposed in VOT since they showed excellent performance with high computational efficiency. However, discriminative capability of Siamese-based trackers are considerably poor \cite{bib29}, since they only  focus on learning a visual representation of the target object to match the similarity while ignoring the background and target specific information and hence showed low performance in occlusion and deformation scenes.  

Transformer \cite{bib30} was  introduced in the field of Natural Language Processing (NLP) to capture the long range dependencies between input and output sequences for machine translation tasks. Encoder and decoder stacks are the two primary components of a Transformer architecture and they are used to learn contextual information of the inputs and to generate the output sequences, respectively. Based on great success of the Transformers in NLP applications, researchers modified the Transformer architecture \cite{bib33} to solve the computer vision tasks. In the last three years, several Transformer architectures are proposed and  showed state-of-the-art performances in various vision tasks such as object detection\cite{bib35},  image classification\cite{bib33, chen2021crossvit}, semantic segmentation \cite{strudel2021segmenter},   and point cloud learning\cite{guo2021pct}. In addition, a few review studies \cite{han2022survey, khan2022transformers,islam2022recent} have also been conducted to analyze the performance of Transformers in vision tasks and to compare their performance with state-of-the-art CNN architectures.

Although Transformers and CNNs are two types of artificial neural network models, they differ from each other in several ways, including their architecture, design principles, and performances. The CNNs and Transformers consume images differently, with CNNs consuming images as arrays of pixel values and Transformers consuming images as sequences of patches, respectively. Based on the studies \cite{han2022survey}, Transformers are good at capturing global information of an image than the state-of-the-art CNN models. Furthermore, Transformers are better equipped to capture long-range dependencies in an image without sacrificing computational efficiency. On the other hand, increasing the size of convolutional kernels in CNNs can hinder their learning capability. Researchers \cite{bib33} also found that  Transformers are difficult to optimize and hence needs more training data than the CNN models. Additionally, Transformers have a larger number of parameters, which can lead to overfitting if not enough training data is available.  However, once trained, Transformers can produce testing outputs with fewer computational resources than the corresponding CNN models, as they can process the data in a parallel manner. In terms of transfer learning, Transformers shown promising results and has been gaining popularity in recent years while CNNs are generally better suited for small and medium-sized datasets.  Overall, Transformer models are replacing the CNNs in computer vision tasks because of their attention mechanism and  global feature capturing capabilities. Although accuracy of Transformer is modest in small and medium-sized datasets,  they are expected to replace CNNs in all tasks in the coming years.

For the last three years, due to the great success of Transformers in other computer vision tasks, a set of Transformer based VOT approaches \cite{bib66,bib37,bib36,zhao2021trtr,bib74,bib76,bib75,xing2022siamese,ma2022unified,zhong2022correlation,bib67,bib73, xie2021learning,bib38,bib77, bib80,bib79,bib78,lan2022procontext,yang2023bandt,wu2023dropmae,xie2023videotrack,gao2023generalized,wei2023autoregressive,chen2023seqtrack} have been proposed and showed excellent performance on publicly available  single object tracking benchmark datasets. They have been proposed as  Siamese-based template matching trackers or discriminative approaches and showed outstanding performances than the CNN-based trackers. In some early approaches \cite{bib66,bib37,bib36,zhao2021trtr,bib74,bib76}, a hybrid-type tracking model is proposed by combining the CNN and Transformer architectures together. As shown in Fig.~\ref{fig1:Comparison}, these hybrid-type trackers, called as CNN-Transformer based trackers, showed better tracking robustness while maintaining considerable tracking speed than the CNN-based trackers since they combined the  attention mechanism of Transformers with the hierarchical learning capability of CNNs. From last year, researchers have initiated a new era in VOT by proposing a set of trackers that rely solely on Transformer architecture. These trackers are referred to as fully-Transformer based trackers, and they have demonstrated exceptional tracking robustness compared to CNN-based and CNN-Transformer based trackers because they successfully handle the information flow between target template and search region features. Overall, the introduction of Transformers in VOT opens up several new possibilities and perspectives. Therefore, there is a need to analyze the performance of Transformers in single object tracking to identify future research directions.

VOT research has been conducted for more than two decades and several approaches have been proposed. Over these years, several review studies \cite{bib29,bib39,bib40,bib41,bib53,bib54,bib56,bib57,bib58,bib59,smeulders2013visual,cannons2008review, chen2022visual,javed2022visual} have been conducted to analyze and compare the performance of VOT approaches. Most of the early review studies \cite{bib39, bib53, bib56, bib59, smeulders2013visual,cannons2008review} were conducted to evaluate the performance of hand-crafted feature-based trackers, while recent studies \cite{bib29, bib40, bib41, bib54, bib57, bib58,javed2022visual}  focus solely on deep feature-based trackers. Furthermore, while most studies \cite{bib29,bib39,bib53,bib54,bib56,bib57,bib58,cannons2008review} only summarize the literature or compare the reported results of tracking approaches, a few studies \cite{bib40, bib59, chen2022visual, bib41} perform an analysis of tracker performance through experiments conducted on the same computing platform. To the best of our knowledge, no survey studies have been conducted that include  Transformer-based trackers.

In recent years, several challenging benchmark datasets \cite{bib42, bib62, bib72, mueller2016benchmark} have been introduced with various challenging attributes and performance measures. To identify the future direction of VOT, it is necessary to experimentally evaluate and compare all recently proposed trackers on the same computing platform in order to avoid bias. Additionally, even though the reported results of Transformer-based trackers are better than those of CNN-based trackers, there is a need to evaluate their performance through detailed experiments in order to  identify research gaps and propose future tracking models. With the aim of achieving these objectives, we have conducted an experimental survey on recently proposed single-object trackers. This study specifically focuses on trackers that utilize the Transformer in their tracking pipeline. We have considered all the Transformer trackers that were published in indexed journals and prestigious conferences.

This study makes the following notable contributions:
\begin{enumerate}
	\item We have conducted a comprehensive review of the literature on Transformer-based tracking approaches. Non-Transformer based trackers were not included in our literature review as they have already been covered by previous studies.  
	
	\item We have experimentally evaluated and then compared the tracking robustness of Transformers in VOT on five challenging benchmark datasets by categorizing and analyzing the state-of-the-art trackers as fully-Transformer based trackers, CNN-Transformer based trackers, and CNN-based trackers.
	\item The computational efficiencies of the state-of-the-art trackers were evaluated on a common computing platform using the source code and models provided by the authors for fair comparison. 
	\item We have conducted the tracking attribute-wise evaluation on three benchmark datasets to identify the most challenging scenarios for recent trackers. 
	\item Based on the experimental findings, we provide the suggestions for future directions of Transformer based single object tracking.    
\end{enumerate}
The rest of the manuscript is structured as follows: The previously conducted review studies are discussed and the rationale behind this survey study is established in Section \ref{sec2}.  Section \ref{sec3} provides an in-depth overview of the Transformer architecture. The detailed literature review of Transformer-based tracking approaches is presented in Section \ref{sec4}. The experimental  results of tracking robustness and computational efficiencies are presented in Section \ref{sec5}. The summary of the findings and the proposed future directions for Transformer-based tracking are discussed in Section \ref{sec6}. Finally, the paper is concluded in Section \ref{sec7}.

\section{Related Work}\label{sec2}

In this section, we justify the need for this study, despite the several experimental surveys and review studies conducted in the past two decades for single object tracking. We have summarized the previous studies in Table~\ref{ReviewSummary}, categorizing them by their review type and feature type.

\begin{table*}
	\caption{Summary of  the review and experimental survey studies in single object tracking.} \label{ReviewSummary}
	\centering\footnotesize
	\begin{center}
		\begin{tabular}{ccccccc@{\extracolsep{0.20cm}}cc}\hline
			\multirow{3}{*}{Study} &  \multirow{3}{*}{ Year} & \multicolumn{5}{c} {Review Type} &  \multicolumn{2}{c} {Feature   Type} \\ \cmidrule{3-7} \cmidrule{8-9}
			
			&  & Literature & Reported result & Experimental & Attribute & Efficiency & Hand  & \multirow{2}{*}{Deep} \\ 
			&  & Survey & based Analysis & Survey & Analysis & Analysis & crafted &  \\
			\midrule
			
			Yilmaz \emph{et al.} \cite{bib53} &	2006 & \checkmark & - &	- & - & - & \checkmark & -\\
			Cannons \emph{et al.} \cite{cannons2008review} & 2008  & \checkmark & - &	- & - &  - & \checkmark & -\\
			Li \emph{et al.} \cite{bib39} &	2013    & \checkmark & - &	- & \checkmark & - & \checkmark & -\\
			Deori \emph{et al.} \cite{bib56} &	2014    & \checkmark & - &	- & - &  - & \checkmark & -\\
			Wu \emph{et al.} \cite{bib59} & 2015 & \checkmark & - & \checkmark & \checkmark &  \checkmark & \checkmark & - \\
			Pflugfelder \emph{et al.} \cite{bib54} &	2017 & \checkmark & \checkmark & - & - & - &- &\checkmark   \\
			Li \emph{et al.} \cite{bib41} &	2018 &\checkmark	& -	& \checkmark & \checkmark & \checkmark&- & \checkmark\\
			Soleimanitaleb \emph{et al.} \cite{bib57} &	2019 & \checkmark & - & - & - & - &\checkmark &\checkmark \\
			Ondrašovič \emph{et al.} \cite{bib29}&	2021&\checkmark&\checkmark&	- & - & - & - & \checkmark \\
			Marvasti-Zadeh \emph{et al.} \cite{bib40} &	2021 & \checkmark & - & \checkmark & \checkmark & \checkmark & - &\checkmark\\
			Zhang \emph{et al.} \cite{bib58} &	2021 & \checkmark & \checkmark & - & - & - &  \checkmark & \checkmark \\
			Chen \emph{et al.} \cite{chen2022visual} & 2022 & 	\checkmark & - & \checkmark & \checkmark & \checkmark & \checkmark & \checkmark\\
			Javed \emph{et al.} \cite{javed2022visual} & 2022 & 	\checkmark & \checkmark & - & - & - & - & \checkmark\\

		\end{tabular}
	\end{center}
\end{table*}

During the period from 2006 to 2014, several review studies were conducted on hand-crafted feature-based VOT approaches. However, since no benchmark datasets were developed during that period, the evaluation of trackers was performed on various video sequences using different performance metrics. Therefore, comparing the performance of trackers is a challenging task, and as a result, the review studies focused only on conducting literature surveys. Yilmaz \emph{et al.} \cite{bib53} conducted a literature survey on object tracking by classifying the trackers into different categories based on the target representation, feature type, model learning algorithm, and searching technique.  Similar to that work, Cannons \emph{et al.} \cite{cannons2008review} developed a detailed technical report on visual tracking and conducted the literature survey by classifying the trackers based on their target representation as point, contour, and region trackers. Li \emph{et al.} \cite{bib39} performed a  study on VOT by providing a detailed literature survey. Also, they have compared a few trackers based on the qualitative results of some tracking attributes such as occlusion, illumination, and deformation.  Deori \emph{et al.} \cite{bib56} conducted a literature survey by categorising the trackers based on their searching algorithms. Since deep learning-based trackers outperform hand-crafted trackers by a large margin, these survey studies are no longer useful in the present era.

The first experimental survey on  single object tracking was conducted by Wu \emph{et al.} \cite{bib59} based on the evaluation of thirty-one appearance based trackers.  They have developed a benchmark dataset, known as Object Tracking Benchmark (OTB), and then evaluated the tracking performances and efficiencies of the trackers based on their overall accuracies and tracking speed on a same computing platform. In addition, trackers are evaluated and compared in eleven challenging tracking attributes (scenarios). Although this study only considers the hand-crafted feature based trackers, it is still followed by many researchers due to the important of the OTB dataset.  

In the last ten years, deep learning-based single object trackers have shown excellent tracking performance compared to hand-crafted feature-based trackers. Currently, only a few studies have been conducted to review the literature on deep trackers. Soleimanitaleb \emph{et al.} \cite{bib57} analyzed the literature of a few number of hand-crafted and deep trackers. Pflugfelder \emph{et al.} \cite{bib54} only considered the Siamese based deep trackers and compared their performances based on their reported results. Similar to their work, Ondrašovič \emph{et al.} \cite{bib29} considered a detailed literature survey on  Siamese based deep trackers and then compared their tracking performances based on their reported results on the well-known OTB100 \cite{bib59}, VOT2015 \cite{bib61}, VOT2016 \cite{VOT2016}, VOT2017 \cite{VOT2017}, VOT2018 \cite{kristan2018sixth} and GOT-10k \cite{bib62} benchmark datasets. Recently,  Zhang \emph{et al.} \cite{bib58} conducted a survey by classifying the recent trackers as deep trackers and Discriminative Correlation Filter (DCF) based trackers. Also, based on the reported results of corresponding trackers, they have compared the tracking performance and efficiency. Another similar survey study is conducted by Javed \emph{et al.} \cite{javed2022visual} by categorizing the tracking approaches as Siamese based trackers and DCF trackers. These review studies have analyzed the literature of the trackers by classifying them into different categories and provided future directions for VOT. However, since they did not conduct experimental evaluations, their reported results and comparisons are not considered acceptable. Moreover, these survey studies did not consider the efficiency comparison of trackers, even though it is an important aspect in evaluation as the tracking speed is directly related to the complexity of the tracker. Although some survey studies compared tracking speeds based on reported results, this comparison is not considered acceptable because the speed of a tracker is mainly dependent on the efficiency of the GPU on which it was tested.

The first experimental survey on deep-learning-based trackers is conducted by Li \emph{et al.} \cite{bib41} by evaluating the performance and efficiency of twenty-two trackers on OTB100  and VOT2015 benchmarks. Also, they have compared the individual performance of trackers based on eleven tracking attributes and then identified the future directions. Recently, Marvasti-Zadeh \emph{et al.} \cite{bib40}  conducted a comprehensive and experimental survey on deep learning-based VOTs based on the network architecture, tracking methods, long-term tracking, aerial-view tracking,  and online tracking ability of the trackers. In addition, they have compared tracking benchmarks based on their challenging attributes and conducted the tracking speed comparison of trackers. Similar to their work, Chen \emph{et al.} \cite{chen2022visual} conducted an experimental survey on deep and hand-crafted trackers on recently developed benchmarks: OTB, VOT, LaSOT \cite{bib42}, GOT-10k, and TrackingNet \cite{bib72}. Although these experimental evaluation studies reviewed the literature and compared the tracking performances of recently proposed deep trackers in several aspects, they did not give much attention to the comparison based on tracking efficiency.  

In the last three years, similar to other computer vision tasks, language models have played a crucial role in single object tracking. Transformer-based language models have opened up new possibilities and have shown excellent tracking performances and efficiencies compared to CNN-based deep trackers. Since previous review studies have not included Transformer-based trackers, as they were recently proposed, there is a need to categorize and review their model architecture and to identify their strengths and weaknesses for the future directions of single object tracking. To achieve this objective, we conducted a detailed literature review on Transformer-based single object trackers and conducted an experimental survey, which clearly showed that they significantly outperformed CNN-based trackers by a large margin. In addition, unlike many other review studies, we have also experimentally compared the efficiency of recent trackers based on their tracking speed, number of floating-point operations (FLOPs), and the number of parameters in the tracking model. Furthermore, we have discussed the future directions of Transformer-based tracking by identifying research gaps and providing suggestions.

\section{Transformer}\label{sec3}

In this section, we have covered the Transformer\cite{bib30} and Vision Transformer (ViT) \cite{bib33} architectures, as they form the foundation for Transformer-based single object trackers.

\begin{figure}[t]%
	\centering
	\includegraphics[width=0.5\textwidth]{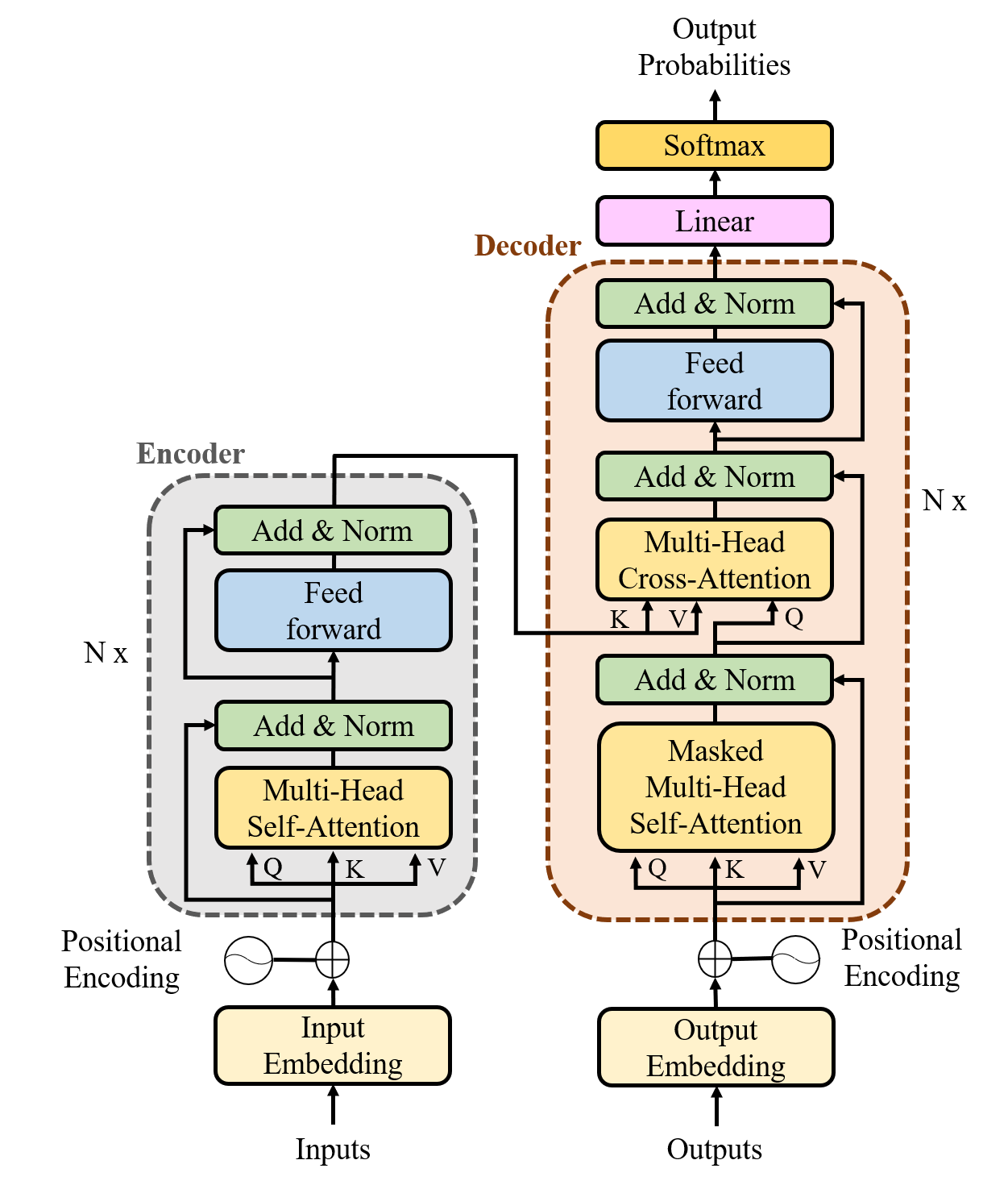}
	\caption{Architecture of the Transformer \cite{bib30} model used in machine translation task. }
	\label{fig2:Transformer}
\end{figure}

Transformer \cite{bib30} was initially introduced in  machine translation tasks and then based on its great success and efficiency it was used in other NLP tasks such as document summarisation and generation. The Transformer architecture is based on an attention mechanism that enables the model to weigh the importance of different parts of the input sequence during processing, leading to improved information flow and the capturing of long-range dependencies. As shown in Fig. ~\ref{fig2:Transformer}, the Transformer architecture consists of encoder and decoder components. The encoder component is made up of $N$ number of identical encoder layers by stacked them on top of each other and the  decoder component is also made up of   $N$ number of identical decoder layers, which are also stacked on each other. Encoder and decoder components of a Transformer architecture are illustrated in the left and right side of Fig.~\ref{fig2:Transformer}, respectively. In the Transformer architecture, all encoder layers have two sub-layers: a self-attention layer, and a  fully connected feed-forward layer.  In addition to those two sub-layers, all decoder layers have an encoder-decoder attention layer in the middle.

Transformer \cite{bib30} receive the input as a vector sequence and use a positional embedding algorithm to add information about the position of each token within that sequence to its representation. After the embedding, the input data is fed to the self-attention sub-layer of the encoder, as it helps to capture the contextual relationships. On the other side, in the decoder layer, an encoder-decoder attention sub-layer  is used to concentrate on relevant parts of the input data. The self-attention mechanism of the Transformer is described in detail in Sections \ref{selfatt} and \ref{multiself}. 

After the self-attention, a fully-connected feed forward layer is used to learn the complex representation of attention features.  It has a simple architecture with two linear transformations and a non-linear activation in between them. This layer can be described as two convolutions with the kernel size of 1. In both encoder and decoder layers, residual connection is included and then it is followed by a normalization layer.  Residual connections are used to preserve the cues from the original input data and to enable the model to learn more accurate representations of the input data.

After the decoder layer stacks, a liner layer is used to produce the output vectors. Finally, a Softmax layer is used to produce the probabilities of the outputs.

\subsection{Self-Attention}\label{selfatt}
Transformer architectures are designed based on an attention concept known as self-attention. It used  to process input sequences in parallel, rather than in a sequential manner like RNNs or LSTMs. In the Transformer architecture, self-attention mechanism captures contextual relationships between input tokens by calculating  the attention weights between each element in the input sequence and all other elements in the same sequence. Transformer used ``Query", ``Key", and ``Value" abstractions to calculate the attention of an input sequence. 

In the first step of the self-attention computation, three distinct vectors: a query vector $\mathbf q$, a key vector $\mathbf k$, and a value vector $\mathbf v$, are created by multiplying an input vector ($\mathbf x$) with three corresponding  matrices: $W_Q$, $W_K$, and $W_V$, that were trained in the training phase. Similarly, all input vectors are packed together to form an input matrix $X$ and then the corresponding ``Query", ``Key", and ``Value" matrices: $Q$, $K$, and $V$ are generated, respectively. 

As the second step of self-attention, a score matrix $S$ is computed for the input $X$ by taking the dot product of $Q$ and $K$ as follows:
\begin{equation}
	S = Q\cdot K^T,
\end{equation}

The score matrix $S$ provides how much attention should be given to other parts of the input sequence based on the input vector at a particular position. In the next step, score matrix $S$ is normalized to obtain more stable gradients as follows:  
\begin{equation}
	S_n = S/\sqrt{d_\mathbf{k}},
\end{equation}
where  $S_n$ is the normalized score matrix, and $d_\mathbf{k}$   is the dimension of the key vector. Then the SoftMax function is applied to the normalized score matrix to convert the scores into probabilities ($P$) as denoted in the following equation:
\begin{equation}
	P = \mathrm{SoftMax}(S_n),
\end{equation}
In the final step of self-attention, the obtained probabilities ($P$) are multiplied with the value matrix $V$ to find the self-attention output values ($Z$) as denoted in the following equation: 

\begin{equation}
	Z = P\cdot V,
\end{equation}
In summary, the entire self-attention mechanism can be unified into a single equation as follows:
\begin{equation}
	\mathrm{Attention}(Q,K,V)=\mathrm{SoftMax}\left(\frac{Q\cdot K^T}{\sqrt{d_\mathbf{k}}}\right)\cdot V
\end{equation}

The encoder-decoder attention sub-layer in the decoder layer is similar to the self-attention sub-layer of the encoder except the key matrix $K$, and the value matrix $V$ are obtained from the encoder block, and the query matrix $Q$ is obtained from its previous layer.

\subsection{Multi-Head Self-Attention}\label{multiself}
The self-attention mechanism is not much capable of focusing on a specific location of input without affecting the
attention on other equally vital locations at the same time. Therefore, the performance of self-attention mechanism is boosted by using multiple heads and this technique is known as multi-head self-attention. 

Multiple sets of weight matrices ($W_Q$, $W_K$, and $W_V$) are used in multi-head self-attention mechanism. They are randomly initialized and trained separately since they are used to project the same input data into a different subspace. Then, the attention function is computed concurrently on each of these projected versions of queries, keys, and values to produce the corresponding  output values. At the final stage of multi-head self-attention, outputs of all attention heads are concatenated and then multiplied with another trainable weight matrix $W_O$ to obtain the multi-head self-attention ($M_{\mathrm{atten}}$) as follows: 
\begin{equation}
	M_{\mathrm{atten}} = \mathrm{Concat}(\mathrm{head}_{1} , ...,\mathrm{head}_{n})\cdot W_O
\end{equation}

where $\mathrm{head}_{i}$ is the output of the attention head $i$. Multi-head self-attention layer is used to concentrate on different positions of input and to represent the input into different subspaces.  

\subsection{Vision Transformer (ViT)}\label{Vit}

\begin{figure}[t]%
	\centering
	\includegraphics[width=0.5\textwidth]{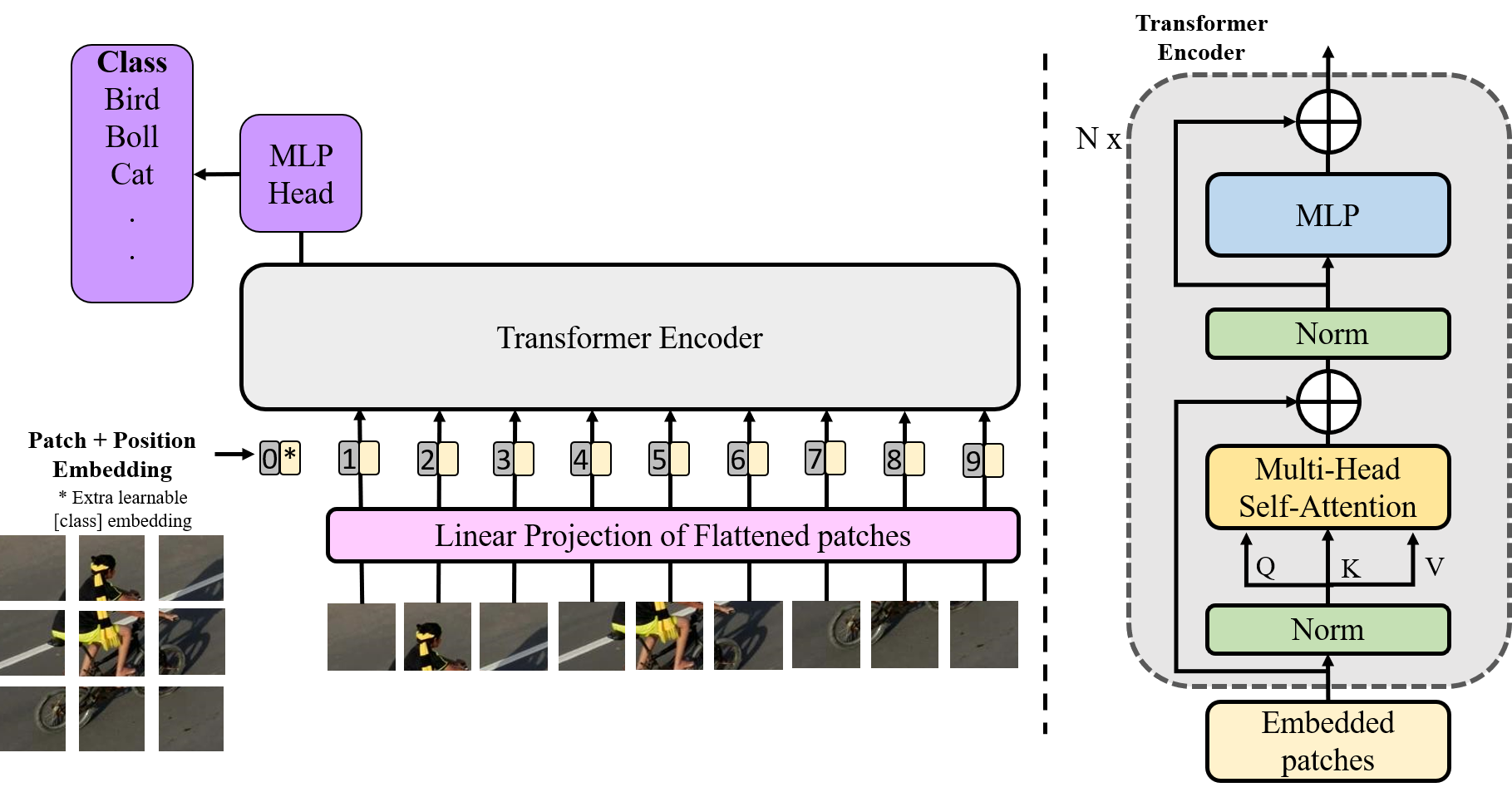}
	\caption{Architecture of the Vision Transformer (ViT) \cite{bib33}, proposed for image recognition task. }
	\label{ViT-Architecture}
\end{figure}
Based on the success of Transformers in NLP tasks, several researchers have attempted to apply them in computer vision tasks and have proposed various architectures. Among these models Vision Transformer (ViT) \cite{bib33} is more efficient than others with a simple architecture as demonstrated in Fig. \ref{ViT-Architecture}.

In the initial step of ViT, an input image $I \in \mathbb{R}^{H \times W \times C} $ is split as equal size patches with the size of $P \times P$. Here, $H$, $W$, and $C$ represent the height, width, and number of channels of the input image, respectively. Then the patches are flattened to form a 2D sequence with the size of $(N \times (P^2 \cdot C))$, where $N$ is the number of extracted patches from an input image. In the next stage of ViT architecture, patches are embedded with position and class information. Then the embedded patches are fed to a set of encoder layers in a sequence manner. The output of the encoder layers is then obtained and fed to a Multi-Layer Perceptron (MLP) network to produce the class-specific scores. The encoder layers of ViT are much similar to the Transformer \cite{bib30} architecture except GELU \cite{hendrycks2016gaussian} non-linear function is used instead of ReLU function.  

ViT, when trained on large datasets, has demonstrated superior results compared to state-of-the-art CNN models. Also, authors of ViT tested it in small and medium-sized datasets with fine-tuning and showed modest results. After the success of ViT in image recognition, several vision Transformer models have been proposed and it has been used in other computer vision tasks. To reduce the computational complexity of ViT, Swin Transformer \cite{bib34} performs self-attention locally within non-overlapping windows that partition an image and introduces a shifting window partitioning mechanism for cross-window connections. Unlike  fixed-size tokens in ViT, Swin Transformer constructs a hierarchical representation by starting from small-sized patches and then gradually merge the neighboring patches in deeper Transformer layers for multi-scale prediction to overcome the scaling problem.  Since pure Transformer models are poor to capture the local information,  CVT \cite{bib99} incorporates two convolution-based operations into the ViT architecture, namely convolutional token embedding, and convolutional projection.  The predefined positional embedding scheme of the ViT is replaced by a conditional position embedding in the CPVT \cite{bib97} Transformer architecture. The TNT \cite{bib98} Transformer further subdivides a $16 \times 16 $ image patch into $4 \times 4 $ sub-patches using a Transformer-in-Transformer framework. Inner Transformer blocks and outer Transformer blocks are used in TNT to capture the interaction between sub-patches and the relationship between patches, respectively. Since ViT is less efficient at encoding finer-level features, VOLO \cite{yuan2022volo} introduces light-weight attention mechanism called Outlooker to encode the token representations with finer-level information efficiently. Overall, ViT has ushered in a new era in computer vision tasks. 

\section{Transformer in Single Object Tracking}\label{sec4}

Based on the model architecture, feature extraction, and feature integration techniques, recent deep trackers can be classified as three categories: CNN-based trackers \cite{li2019siamrpn++,zhang2019deeper,bhat2019learning,yu2020deformable,chen2020siamese,xu2020siamfc++,voigtlaender2020siam,zhang2020ocean,danelljan2020probabilistic,cheng2021learning,guo2021graph,mayer2021learning}, CNN-Transformer based trackers \cite{bib66,bib37,bib36,zhao2021trtr,bib74,bib76,bib75,xing2022siamese,ma2022unified,zhong2022correlation,bib67,bib73} and fully-Transformer based trackers \cite{xie2021learning,bib38,bib77, bib80,bib79,bib78,lan2022procontext}. CNN-based trackers rely solely on a CNN architecture for feature extraction and target detection, while CNN-Transformer based trackers and fully-Transformer based trackers partially and fully rely on a Transformer architecture, respectively. The literature on CNN-Transformer based trackers and fully-Transformer based trackers is reviewed in this section.

\begin{table*}
	\begin{center}
		\scriptsize
		\caption{Summary of CNN-Transformer based and fully-Transformer based trackers.  }\label{TransformerTrackerSummery}%
		\begin{tabular}{p{1.8cm} p{0.4cm} p{1.9cm} p{1.8cm}  p{0.8cm} p{3.8cm} p{3.8cm}}
			\toprule
			\multirow{2}{*}{Tracker} & \multirow{2}{*}{Year} & \multirow{2}{*}{Type} & \multirow{2}{*}{Backbone}  & Template Update & \multirow{2}{*}{Training Dataset} & \multirow{2}{*}{Training Scheme} \\
			\midrule
			
			HiFT \cite{bib76} & 2021 & CNN-Transformer  &	AlexNet  & N & COCO, ImageNet VID, GOT-10k, Youtube-BB & cross-entropy loss, binary
			cross-entropy loss, IoU loss, SGD \\	
			STARK \cite{bib36} & 2021 & CNN-Transformer  &	ResNet-101  & Y & COCO, LaSOT, GOT-10k, TrackingNet & binary cross-entropy loss, L1 loss, GIoU loss, AdamW \\
			TransT \cite{bib37} & 2021 & CNN-Transformer  &	ResNet-50 & N & COCO, LaSOT, GOT-10k, TrackingNet &  binary cross-entropy loss, L1 loss, GIoU loss, AdamW \\
			DTT \cite{bib74}& 2021 & CNN-Transformer  &	ResNet-50  & Y & COCO, LAaSOT, GOT-10k, TrackingNet & cross-entropy loss, IoU loss \\
			TrDiMP \cite{bib66} & 2021 & CNN-Transformer  &	ResNet-50  & Y & COCO, LaSOT, GOT-10k, TrackingNet & L2 norm loss, AdamW \\
			TrTr \cite{zhao2021trtr} & 2021 & CNN-Transformer & ResNet-50 & N & Youtube-BB, ImageNet VID, GOT-10k, LaSOT, COCO & focal loss, L1 loss ,AdamW \\
			DualTFR \cite{xie2021learning} & 2021 & fully-Transformer & Custom & N & ImageNet-1K, COCO, LaSOT, GOT-10k, TrackingNet & cross-entropy loss, GIoU, L1 loss, AdamW \\
			SwinTrack \cite{bib38} & 2021 & fully-Transformer  &	Swin Transformer  & N & COCO, LaSOT, GOT-10k, TrackingNet & varifocal loss,  GIoU loss, AdamW \\
			ProContEXT \cite{lan2022procontext} & 2022 & fully-Transformer & MAE-ViT & Y & COCO, LaSOT, GOT-10k, TrackingNet & focal loss, IoU loss, L1 Loss, AdamW \\ 
			OSTrack \cite{bib78} & 2022 & fully-Transformer  &	MAE-ViT  & N &  COCO, LaSOT, GOT-10k, TrackingNet & L1 loss, GIoU loss, Gaussian weighted focal loss, AdamW \\
			MixFormer \cite{bib80} & 2022 & fully-Transformer  &	CVT  & Y & COCO, LaSOT, GOT-10k, TrackingNet & L1 loss, GIoU loss, cross-entropy, AdamW \\
			SimTrack \cite{bib79} & 2022 & fully-Transformer  &	ViT  & N & COCO, LaSOT, GOT-10k, TrackingNet & L1 loss, GIoU loss, AdamW \\
			SparseTT \cite{bib77} & 2022 & fully-Transformer  &	Swin Transformer & N & COCO, LaSOT, GOT-10k, TrackingNet, ILSVRC VID, ILSVRC DET &  focal loss, IoU loss, AdamW \\
			ToMP \cite{bib75} & 2022 & CNN-Transformer  &	ResNet-101  & Y & COCO, LaSOT, GOT-10k, TrackingNet & GIoU loss, AdamW \\
			CSWinTT \cite{bib67}& 2022 & CNN-Transformer  &	ResNet-50  & Y & LaSOT, GOT-10k, TrackingNet & L1 loss, GIoU loss, cross-entropy, AdamW \\
			AiATrack \cite{bib73} & 2022 & CNN-Transformer & ResNet-50  & Y & COCO, LaSOT, GOT-10k, TrackingNet &  L1 loss, GIoU loss, AdamW \\
			CTT \cite{zhong2022correlation} & 2022 & CNN-Transformer & ResNet-50 & N & COCO, LaSOT, GOT-10k, TrackingNet & cross-entropy loss, GIoU loss, L1 loss,AdamW\\
			UTT \cite{ma2022unified} & 2022 & CNN-Transformer & ResNet & N & COCO, LaSOT, GOT-10k, TrackingNet & GIoU loss, L1 loss, AdamW\\			
			SiamTPN \cite{xing2022siamese} & 2022 & CNN-Transformer & ShuffleNetV2 & N & COCO, LaSOT, GOT-10k, TrackingNet &  cross-entropy loss, GIoU loss, L1 loss, AdamW \\
			BANDT \cite{yang2023bandt} & 2023 & CNN-Transformer & ResNet-50 & Y & COCO, LaSOT, GOT-10k, TrackingNet & cross-entropy loss, GIoU loss, Focal loss, smooth L1 loss, AdamW \\
			GRM \cite{gao2023generalized} & 2023 & fully-Transformer & MAE-ViT & N & COCO, LaSOT, GOT-10k, TrackingNet & GIoU loss, Focal loss, L1 loss, AdamW\\
			DropTrack \cite{wu2023dropmae} & 2023 & fully-Transformer & DropMAE-ViT & N & COCO, LaSOT, GOT-10k, TrackingNet & L1 loss, GIoU loss, Gaussian weighted focal loss, AdamW\\
			MATTrack \cite{Zhao2023CVPR} & 2023 & fully-Transformer & MAT-ViT & N & COCO, LaSOT, GOT-10k, TrackingNet, VID & L1 loss, GIoU loss, AdamW\\
			VideoTrack \cite{xie2023videotrack} & 2023 & fully-Transformer & MAE-ViT & Y & COCO, LaSOT, GOT-10k, TrackingNet & - \\
			SeqTrack \cite{chen2023seqtrack} & 2023 & fully-Transformer & MAE-ViT & Y & COCO, LaSOT, GOT-10k, TrackingNet & cross-entropy loss AdamW \\
			ARTrack \cite{wei2023autoregressive} & 2023 & fully-Transformer & ViT & N & LaSOT, GOT-10k, TrackingNet & cross-entropy loss, SIoU loss, AdamW \\

			\bottomrule
		\end{tabular}
	\end{center}
\end{table*}

Generally, Transformer architectures require a massive number of training samples \cite{bib33} to train their models. Since the target is given in the first frame of a tracking sequence, obtaining a large number of samples is not possible in VOT and hence all of the fully-Transformer based and CNN-Transformer based trackers use a pre-trained network and considered it as the backbone model. In addition, some of these trackers update the target template to capture the temporal cues during tracking, while others do not. Moreover, they  trained their models on various benchmark datasets such as COCO \cite{lin2014microsoft}, LaSOT, GOT-10k, TrackingNet, and Youtube-BB \cite{real2017youtube}. We have summarized these details in Table \ref{TransformerTrackerSummery} for all Transformer-based and CNN-Transformer based approaches, providing information on their backbone network, template update details, training dataset, and training scheme details.

\begin{figure}
	\centering
	\includegraphics[width=0.47\textwidth]{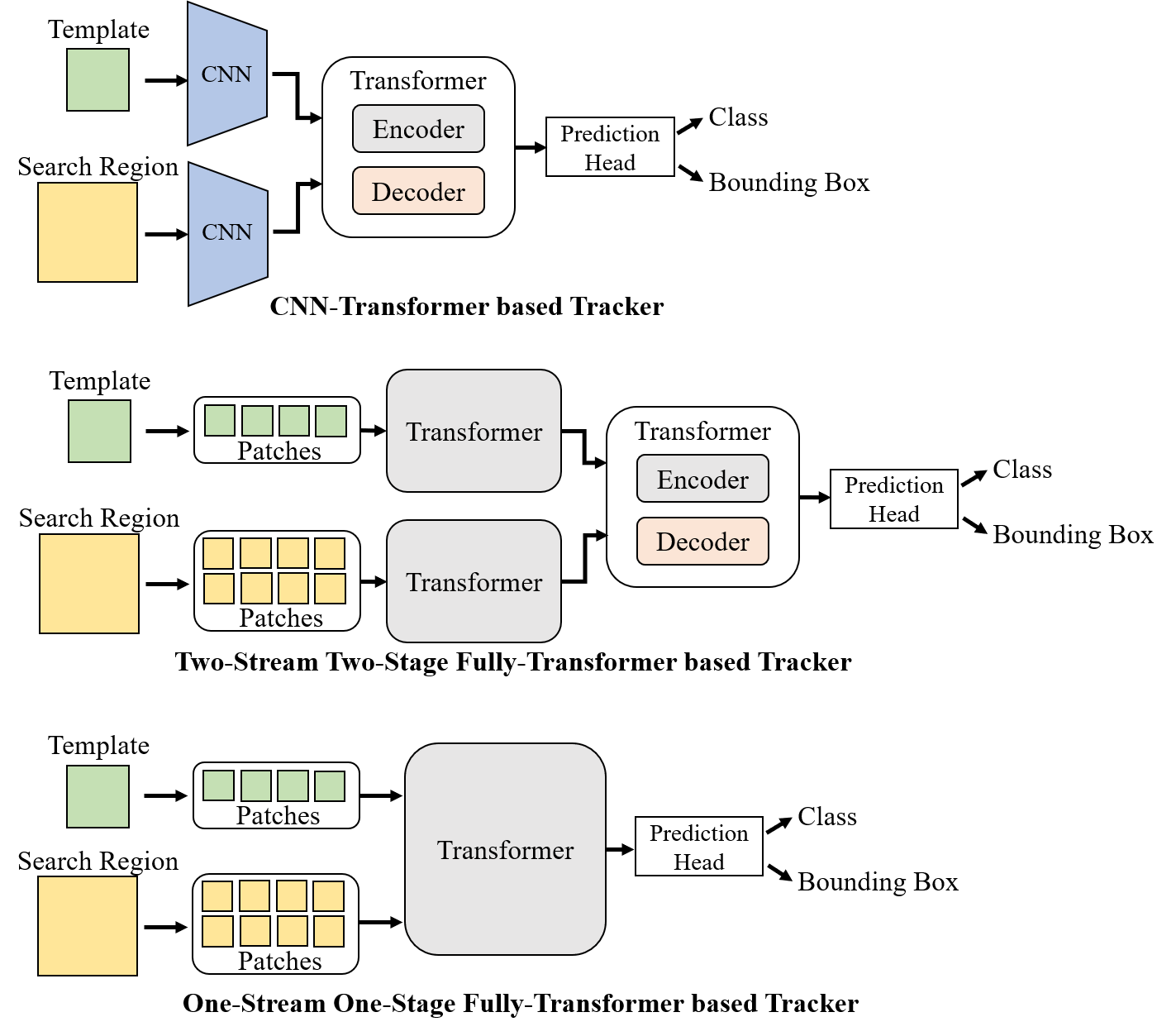}
	\caption{General architectures of CNN-Transformer, One-stream One-stage fully-Transformer, and Two-stream Two-stage fully-Transformer based trackers. Although ARTrack \cite{wei2023autoregressive} and SeqTrack \cite{chen2023seqtrack} belong to the One-stream One-stage category, they are capable of predicting the target without the need for a prediction head.}
	\label{Transformerimage}
\end{figure}

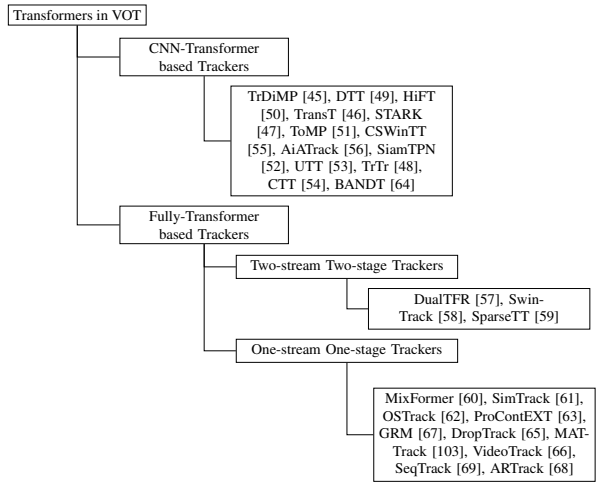
\begin{figure} [ht]
	\begin{adjustbox}{width=0.47\textwidth}
		\begin{tikzpicture}
			[
			level 1/.style = {text width=3cm},
			level 2/.style = {text width=4cm},
			level 3/.style = {text width=4cm},
			every node/.append style = {draw, anchor = west, align = center},
			grow via three points={one child at (0.4,-0.8) and two children at (0.5,-0.8) and (0.5,-1.6)},
			edge from parent path={(\tikzparentnode\tikzparentanchor)  |- (\tikzchildnode\tikzchildanchor)};
			]
			\footnotesize
			\node {Transformers in VOT}
			child {node  {CNN-Transformer based Trackers}
				child [missing] {}
				child {node {TrDiMP \cite{bib66}, DTT \cite{bib74}, HiFT \cite{bib76}, TransT \cite{bib37}, STARK \cite{bib36}, ToMP \cite{bib75}, CSWinTT \cite{bib67}, AiATrack \cite{bib73}, SiamTPN \cite{xing2022siamese}, UTT \cite{ma2022unified}, TrTr \cite{zhao2021trtr}, CTT \cite{zhong2022correlation}, BANDT \cite{yang2023bandt}}}
			} 
			child [missing] {}
			child [missing] {}
			child [missing] {}
			child {node {Fully-Transformer based Trackers}
				child {node {Two-stream Two-stage Trackers}
					child {node {DualTFR \cite{xie2021learning}, SwinTrack \cite{bib38}, SparseTT \cite{bib77}}}}
				child [missing] {}
				child {node {One-stream One-stage Trackers}
					child [missing] {}
					child {node {MixFormer \cite{bib80}, SimTrack \cite{bib79}, OSTrack \cite{bib78}, ProContEXT \cite{lan2022procontext}, GRM \cite{gao2023generalized}, 								DropTrack \cite{wu2023dropmae},  MATTrack \cite{Zhao2023CVPR}, VideoTrack \cite{xie2023videotrack}, SeqTrack \cite{chen2023seqtrack}, ARTrack \cite{wei2023autoregressive}}}}
			};

		\end{tikzpicture}
	\end{adjustbox}
	\caption{Classification of CNN-Transformer and fully-Transformer trackers based on their model architecture and tracking pipeline. Multiple pipeline of networks are used in `Two-stream Two-stage'  trackers while single pipeline of networks are used in 'One-stream' trackers. Also, feature extraction and feature fusion are done through two distinguishable stages in `Two-stage' trackers while they are done together in `One-Stage' trackers.}
	\label{fig4:Classification}
\end{figure}

We have conducted a literature review of CNN-Transformer based and fully-Transformer based trackers, focusing on their model architecture.  As shown in Fig.~\ref{Transformerimage},  CNN-Transformer based trackers used a CNN backbone for feature extraction and then used a Transformer architecture for feature integration. Fully-Transformer based trackers can be classified further as "Two-stream Two-stage" trackers, and "One-stream One-stage" trackers.  In Two-stream Two-stage trackers, two identical pipeline of network branches (two-stream) are used to extract the features of the target image and search image. Also, in this category of trackers,  feature extraction and feature fusion of target template and search region are done in two distinguishable stages (two-stage). On the other side, in One-stream One-stage trackers, a single pipeline of networks are used and the feature extraction and feature fusion are done together through a single stage.  Fig.~\ref{fig4:Classification} shows the classification of CNN-Transformer based, and fully-Transformer based trackers. 

\subsection{CNN-Transformer based Trackers}
Most of the recent CNN-based trackers \cite{li2019siamrpn++,zhang2019deeper,bhat2019learning,yu2020deformable,chen2020siamese,xu2020siamfc++,voigtlaender2020siam,zhang2020ocean,danelljan2020probabilistic,cheng2021learning,guo2021graph,mayer2021learning} followed a Siamese network architecture by using two identical pipelines of CNNs. In these trackers, features of the target template and search region are extracted by using two identical CNN branches. Then the target localization is done by finding the similarity of target's features in search region's features using a correlation function. Although the correlation operation is simple and fast for feature similarity matching process, it is not good enough to capture the non-linear interaction (occlusion, deformation, and rotation) between the target template and search region and hence the tracker's performance is limited. To successfully address this issue, researchers have started incorporating the Transformer for the feature fusion process, resulting in trackers known as CNN-Transformer based trackers.

Similar to most of the CNN-based trackers, CNN-Transformer based trackers also use two Siamese-like identical pipeline of networks. In the beginning of these pipelines, target template's and search region's features are extracted using a CNN backbone. Then the extracted deep features are flattened into vectors and thereafter fed to a Transformer to capture the similarity features of the target in the search region.

\begin{figure}[b]%
	\centering
	\includegraphics[width=0.47\textwidth]{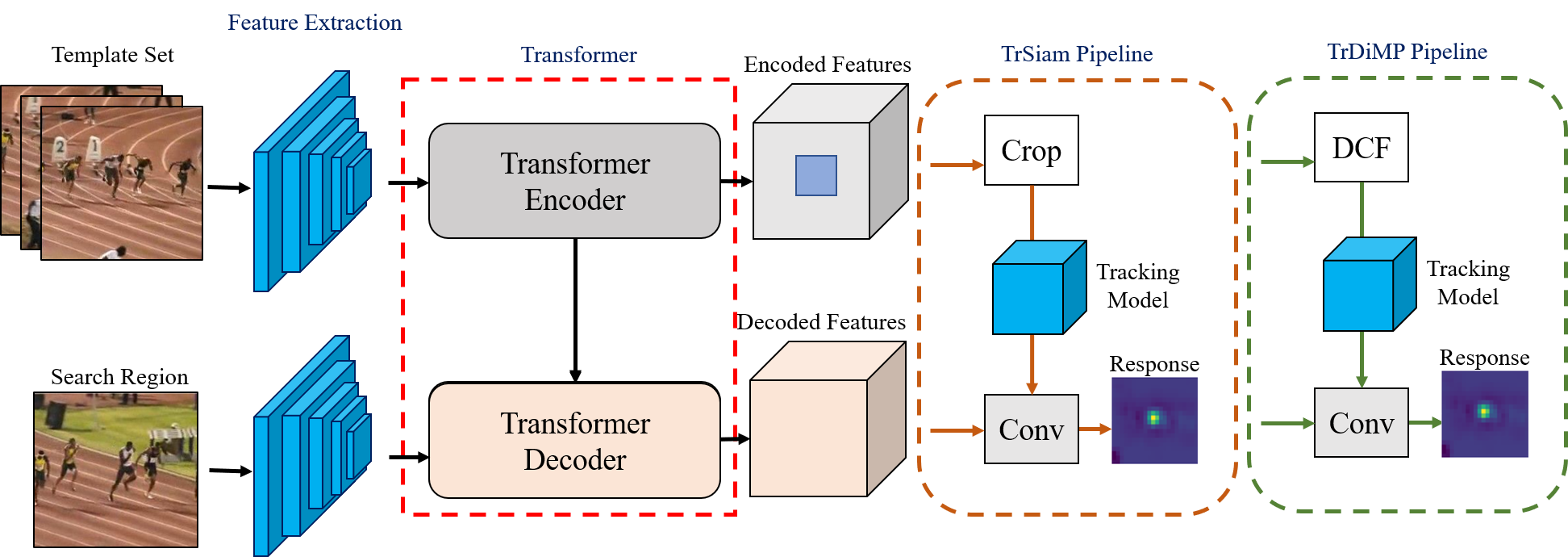}
	\caption{Architecture of the  first CNN-Transformer based tracker\cite{bib66} with TrSiam and TrDiMP branches. Target template set's features and search region features are fed to the Transformer encoder and decoder to locate the target.}
	\label{TrDiMP-Architecture}
\end{figure}

The first CNN-Transformer based tracker was proposed by Wang \emph{et al.} \cite{bib66} by introducing a  Transformer into both generative and discriminative tracking paradigm. In their Siamese-like tracking architecture, a set of template patches and search region are fed to a CNN backbone to extract the deep features. Then, as shown in Fig. \ref{TrDiMP-Architecture}, extracted template features are fed to the Transformer's encoder to capture the high-quality target's features using the attention mechanism. Similarly, search region features are fed to the Transformer decoder to produce the decoded features by aggregating the informative target cues from previous frames with the search region features. In the TrSiam pipeline, similar to the SiamFC\cite{bib19} tracker, the target feature is cropped from the encoded feature, and then cross-correlated with the decoded features to locate the target location. In the TrDiMP pipeline, end-to-end Discriminative Correlation Filters(DCF) \cite{bhat2019learning} are applied on the encoded features to generate a target model, which is then used to convolve with decoded features to locate the target in the search region. Since a collection of target template features is fed to the Transformer of this tracker, it is able to locate the target even with severe appearance changes.

Similar to the TrDiMP \cite{bib66} tracker, Yu \emph{et al.} \cite{bib74} introduced the encoder-decoder Transformer architecture \cite{bib30} in VOT and their tracker is referred to as DTT. They have also used a Siamese-like tracking pipeline and extracted the deep features using a backbone CNN architecture. In their tracking model, target templates are fed with background scenes and then the Transformer architecture captures the most discriminative cues of the target. Since their approach involves conducting the tracking without the need to train a separate discriminative model, it is simple and has demonstrated high tracking speed in benchmark datasets.

Recently, Yang \emph{et al.} \cite{yang2023bandt} found that the TrDiMP \cite{bib66} tracker is computationally expensive since it computes the attention in all possible spatial locations of the target template feature, and hence the model weights are increased with respect to the pixel counts of the feature maps. To address this issue, they proposed a border-aware tracking framework with deformable Transformers, referred to as the BANDT tracker. Instead of computing attention among all possible locations of the target template features, the BANDT tracker only calculates attention in a few specific locations around the target boundary. Based on the inspiration from the Deformable DETR Transformer \cite{zhudeformable}, the BANDT tracker extracts a small set of key sampling points around the border of the target object in the template image. Since the BANDT tracker enhances the border features of the target object, it shows better performance than the TrDiMP tracker with fewer computational costs. 

\begin{figure}[t]%
	\centering
	\includegraphics[width=0.47\textwidth]{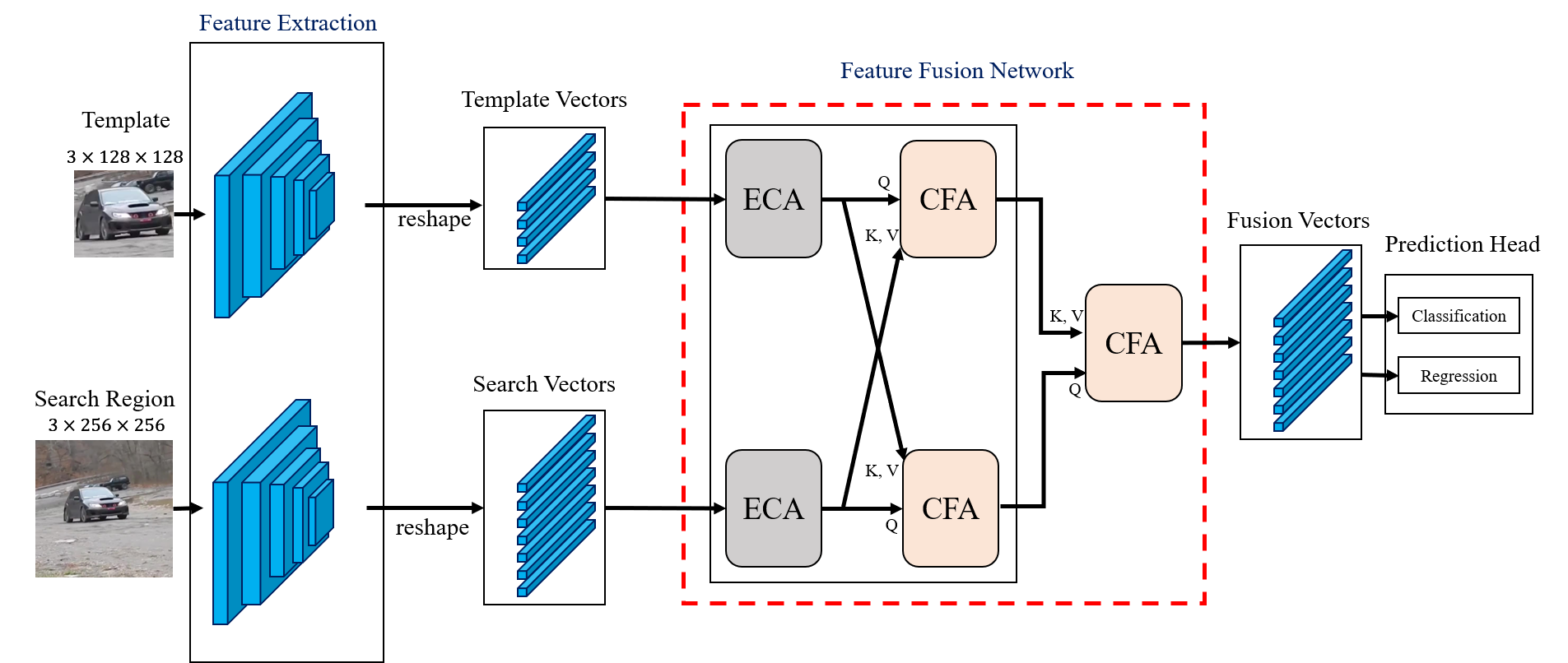}
	\caption{Architecture of the  TransT tracker \cite{bib37}. It has a CNN backbone network, a Transformer network, and a prediction network for  deep feature extraction, feature fusion, and  target localization, respectively.  }
	\label{TransT-Architecture}
\end{figure}

Another Siamese-like architecture is proposed in TransT \cite{bib37} tracker.  As demonstrated in Fig. \ref{TransT-Architecture}, TransT tracker has three modules:  a CNN backbone network, a Transformer based feature fusion network, and a prediction network. Similar to other CNN-Transformer based trackers, target template's and search region's features are extracted using the ResNet50 \cite{he2016deep} model. Then these features are reshaped using a $ 1 \times 1$ convolutional layer and fed to the  feature fusion network. The Transformer-based feature fusion network has a set of  layers, and each layer has an ego-context augment module (ECA) and a cross-feature augment (CFA) module to enhance the self-attention and cross-attention, respectively. Finally, fused features are fed to a prediction head, and it locates the target and find the coordinates using a simple classification and regression branches, respectively. TransT tracker showed excellent performance than the CNN-based trackers by utilizing the Transformer for feature fusion instead of the correlation matching of the previous approaches.  

Several CNN-Transformer trackers are proposed by training the Transformer to capture the relationship between the target template and search region features. The TrTr tracker  \cite{zhao2021trtr} trained a Transformer to capture the global information of  the target template and then used that cues to find the accurate correlation between target and search region. Another similar approach is proposed by Xiuhua \emph{et al.} \cite{hu2022transformer} with a template update mechanism. Recently, Zhong \emph{et al.} \cite{zhong2022correlation} proposed a tracker referred to as CTT, and improved the feature fusion between target and search region features by including a correlation module into the Transformer architecture. CTT tracker avoids the background interference in long-term tracking by using their correlation based Transformer architecture. 

\begin{figure}[t]%
	\centering
	\includegraphics[width=0.47\textwidth]{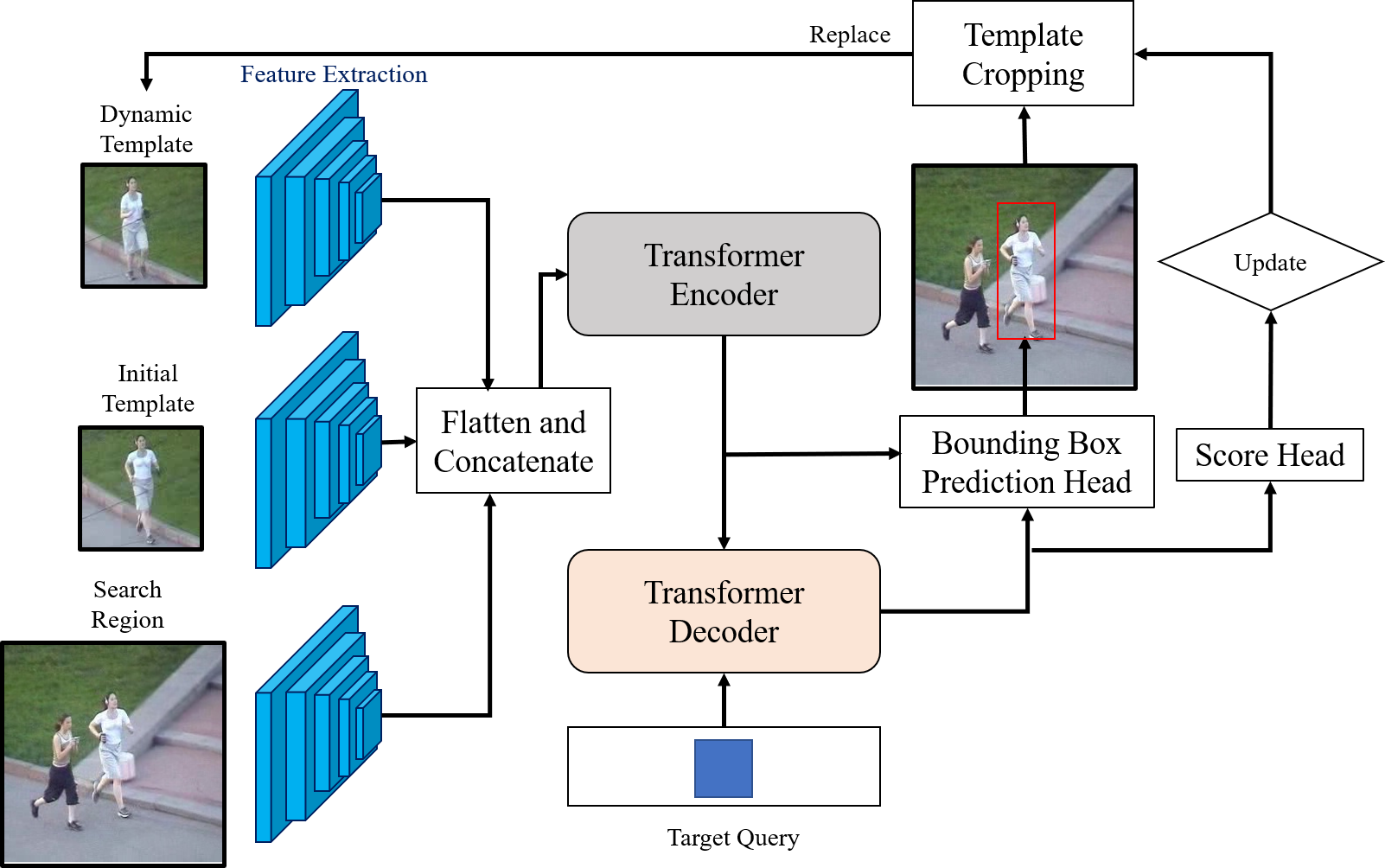}
	\caption{Architecture of the  STARK tracker \cite{bib36}. It captures the spatial and temporal information of the target using the Transformer. }
	\label{STARK-Architecture}
\end{figure}

In contrast to the previously mentioned CNN-Transformer based trackers, Yan \emph{et al.} \cite{bib36} proposed a new Transformer architecture for VOT based on the DETR \cite{bib35} object detection Transformer. Their tracker is referred to as STARK and they have trained the Transformer to capture the spatial and temporal cues of the target object. ResNet \cite{he2016deep} is used to extract the deep features of the initial target template, dynamic target template, and search regions. Then, as shown in Fig. \ref{STARK-Architecture}, these features are flattened, concatenated, and then fed to a Transformer with encoder and decoder architectures. The STARK tracker's Transformer consists of a set of encoder layers, with each layer containing a multi-head self-attention module and a feed-forward network module. The encoder captures the feature dependencies between every element in the tracking sequence, and reinforces the original features with global contextual information. This enables the model to learn discriminative features for object localization. The decoder of the tracker learns a query embedding to predict the spatial positions of the target object by using the Transformer-based detection approach of the DETR. To determine the target's bounding box in the current frame, STARK introduced a corner-based prediction head. Additionally, a score head is learned to control the updates of the dynamic template images. Since the architecture of the STARK is simple and capable of capturing the spatio-temporal information of the target, it has demonstrated better tracking robustness and good tracking speed compared to other CNN-based trackers.

Similar to the STARK tracker, Mayer \emph{et al.} \cite{bib75} proposed a tracker called ToMP, utilizing the DETR object detection Transformer. In this tracker, the target template and search region features from the CNN backbone are concatenated and jointly processed by the Transformer encoder and decoder. Additionally, similar to the DiMP \cite{bhat2019learning} tracker, the ToMP tracker incorporates a target model for localizing the target. However, unlike DiMP, the weights of the target model are obtained using the Transformer architecture instead of a discriminative correlation filter. According to the reported experimental results, the ToMP tracker demonstrates superior tracking performance and speed compared to the STARK tracker.

\begin{figure}[t]%
	\centering
	\includegraphics[width=0.47\textwidth]{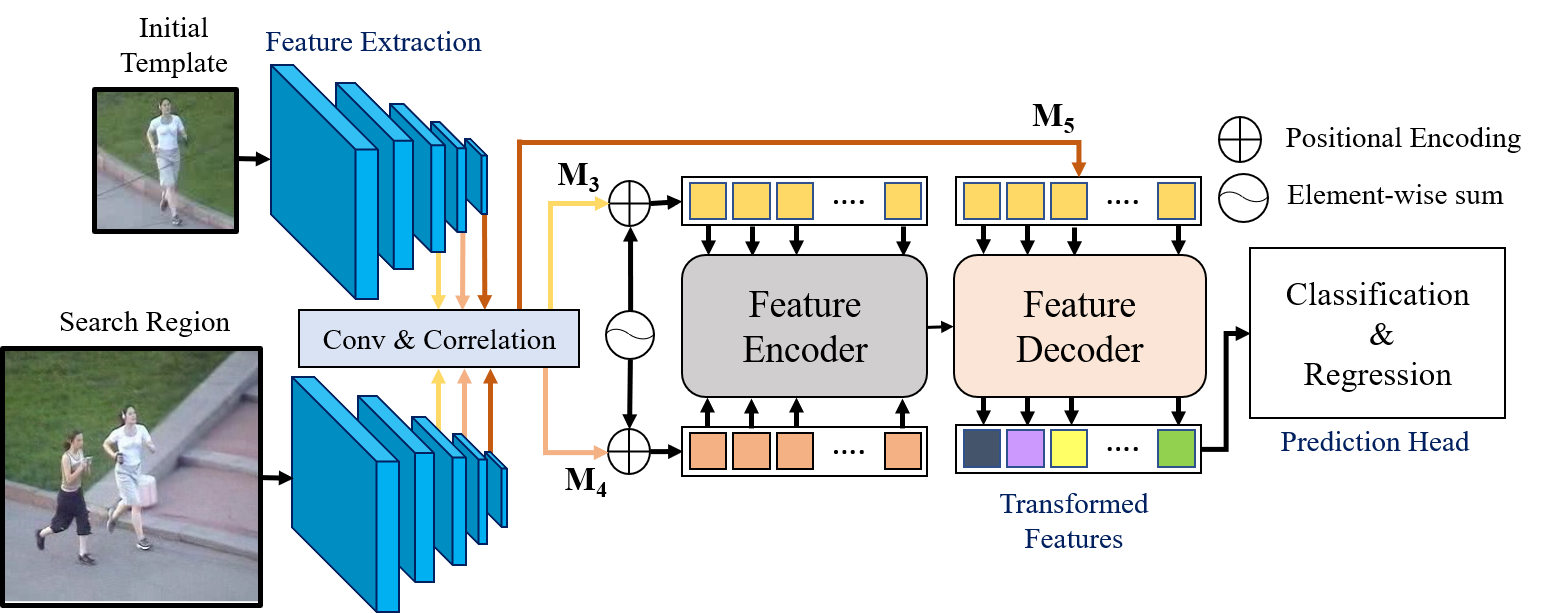}
	\caption{Architecture of the  HiFT tracker\cite{bib76}. It extracted the multi-level convolutional features of the target template and search region and then fed the high-resolution features to encoder and low-resolution features to decoder.}
	\label{HiFT-archritecture}
\end{figure}

Most of the CNN-Transformer based trackers \cite{bib36, bib75, bib37, bib74, bib66} utilized the deep features from the last convolutional layer of a backbone CNN. Different from these trackers,  as illustrated in Fig. \ref{HiFT-archritecture}, Cho \emph{et al.} \cite{bib76} utilized the multi-level convolutional features of the target template and search region from the last three layers (3 to 5) of the AlexNet \cite{bib15} model. Their tracker is referred to as HiFT and it was originally proposed for aerial tracking. In this tracker, high resolution cross-correlated features are fed to the Transformer's encoder to capture the spatial cues of the target with different scales. Also, low resolution cross-correlated features of the last convolutinal layers of the target template and search region are fed to the  Transformer's decoder to capture the semantic features. Finally, a prediction head is applied on the Transformer features to locate the target state. HiFT tracker showed better performance than the CNN-based trackers because of its strong discriminative capability.  

Similar to HiFT tracker, Xing \emph{et al.} \cite{xing2022siamese} proposed an aerial tracking approach, referred to as SiamTPN, by  utilizing the multi-level convolutional features of a backbone CNN. However, their Transformer architecture is different than the HiFT tracker since they used a Transformer Pyramid Network (TPN) to fuse the multi-level features of the target and search region. In addition, instead of the multi-head self attention layer, a pooling attention (PA) layer is used in the Transformer architecture of this tracker to reduce the computational complexity and memory load. Based on the reported results, SiamTPN tracker is able to track a target in real-time speed on a CPU. 

\begin{figure}[t]%
	\centering
	\includegraphics[width=0.47\textwidth]{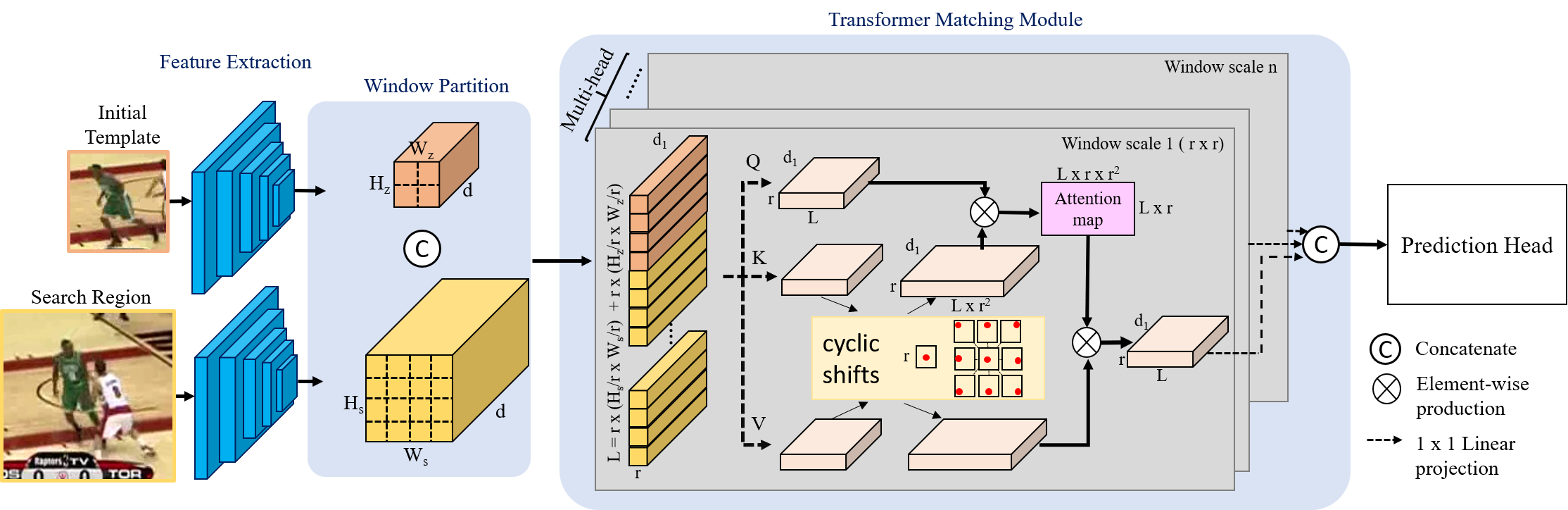}
	\caption{Architecture of the  CSWinTT tracker \cite{bib67}. Instead of pixel-level attention, this tracker considered the window-level attention by fed the flattened window feature pairs to the Transformer.  }
	\label{CSWinTT-Architecture}
\end{figure}

All  of the early CNN-Transformer based approaches \cite{bib36, bib75, bib37, bib74, bib66, bib76, xing2022siamese} extracted the deep features of the target template and search region using a CNN backbone and then flattened them to fed a Transformer network. Although the Transformer captures the pixel-wise attention from the flattened features, the valuable object-level cues from the relative pixel positions are lost. Recently, Song \emph{et al.} \cite{bib67} noticed this issue and proposed a multi-scale cyclic shifting window Transformer tracker (CSWinTT) to tackle this issue. Similar to other CNN-Transformer based trackers,  CSWinTT also extracted the deep features of the target template and search region using the ResNet-50.  Thereafter, as shown in Fig. \ref{CSWinTT-Architecture}, these features are split as small windows and then each windows are flattened and then fed to the Transformer to capture the window-wise attention. In addition, a multi-head multi-scale attention module is used in the Transformer of the CSWinTT to find the relevance between target template and search region windows at a particular scale. Moreover, a cyclic shifting technique is proposed in the window-wise attention module of the CSWinTT to fuse the outputs of each attention heads with different scales. Although the cyclic shifting and window-wise attention are computationally expensive,  CSWinTT approach showed excellent tracking performance  than the other CNN-Transformer based trackers because it considered the window-level attention instead of the pixel-level attention.  

The first CNN-Transformer based unified tracker is proposed by Ma \emph{et al.} \cite{ma2022unified} and it is referred to as UTT.  This unified tracker provided a single model architecture  for single object tracking and multi object tracking to track the single and multiple targets together. In this tracker, deep features of the target template and search region, obtained from a backbone CNN, are fed to a Transformer to locate a target. In this UTT approach, target-specific features are extracted by using a  decoder component of the Transformer. Similarly, a proposal decoder component is used to extract the search region features with respect to the target. Finally, these two features are fed to a Transformer to predict the coordinates of the target in the search region. Experimental results showed that UTT approach successfully handles the single and multi object tracking through a single tracking architecture. 

\begin{figure}[t]%
	\centering
	\includegraphics[width=0.47\textwidth]{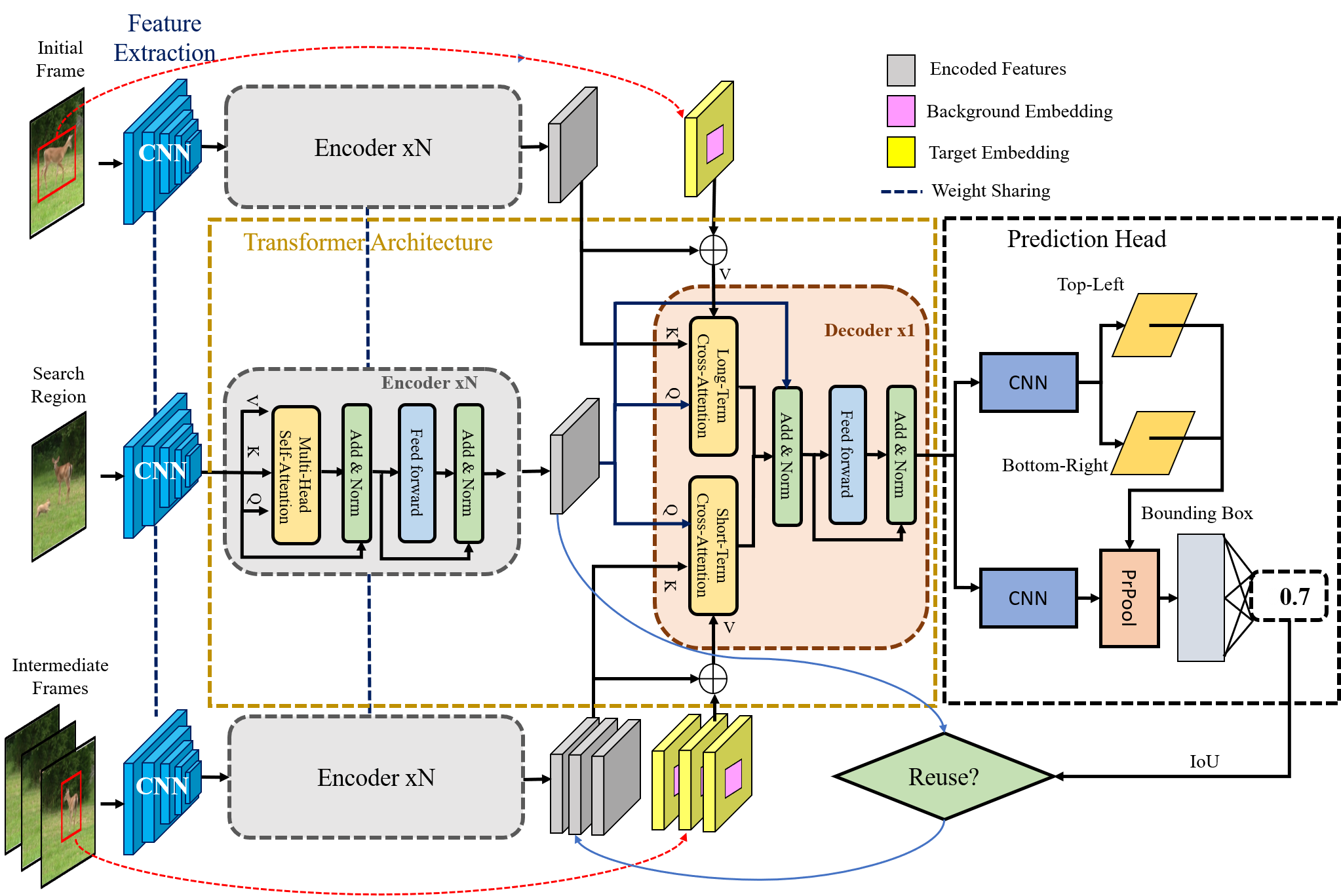}
	\caption{Architecture of the  AiATrack approach \cite{bib73}. An attention-in-attention (AiA) module is used in this tracker to enrich the cross-correlation attention by searching for consensus among all correlation vectors and hence ignores the noisy correlations.     }
	\label{AiATrack-Architecture}
\end{figure}

In most of the CNN-Transformer based trackers, CNN features of the target template and search regions are enriched using the self-attention mechanism. Then, the relationship between them is computed using the cross-attention modules. Based on the correlation between each query-key pair, the highest correlated pair is selected and then used to identify the target in the search region.   Gao \emph{et al.} \cite{bib73} found that if a key has high correlation with a query and the neighbors of that key has less correlation with that particular query then that correlation should be a noise. Based on this finding, as illustrated in Fig. \ref{AiATrack-Architecture}, they have proposed an attention-in-attention (AiA) module by including an inner attention module in Transformer architecture and their tracker is referred to as AiATrack. The proposed inner attention module of the AiATrack approach improves the attention by searching for consensus among all correlation vectors and ignores the noisy correlations. In addition, AiATrack approach introduces a learnable target-background embedding to differentiate the target from the background while maintaining the contextual information. Because of the enhanced attention mechanism, AiATrack approach successfully avoid the distractor objects and hence improves the tracking performance.  

In summary, CNN-Transformer based trackers extracted the deep features of the target template and search region using a  CNN backbone such as ResNet or AlexNet. Then the relationship between these features is computed using the attention mechanism of a Transformer. Finally, a prediction head is employed to locate the target using the features generated by the Transformer. CNN-Transformer based trackers are successfully outperforms the CNN-based trackers since they have used a learnable Transformer instead of the linear cross-correlation operation. Although, a few early trackers borrowed the Transformer architectures from the object detection task and used them without any modification, recent approaches identified the Transformer based tracking issues and then modified their architectures accordingly. In overall, CNN-Transformer based trackers showed excellent tracking performance than the CNN-based trackers.

\subsection{Fully-Transformer Based Trackers}

Although CNN-Transformer based trackers utilize the attention mechanism of Transformers for the feature integration of the target template and search region, they still rely on convolutional features as they use a backbone CNN for feature extraction. Since CNNs capture features through local convolutional kernels, CNN-Transformer based trackers struggle to capture global feature representations.

Due to the success of the fully-Transformer architectures in other computer vision tasks, researchers start using them in single object tracking. Based on the tracking network formulation, we classified the fully-Transformer trackers as Two-stream Two-stage trackers and One-stream One-stage trackers and reviewed their literature in the following subsections.

\subsubsection{Two-stream Two-stage Trackers}\label{subsubsec1}
Two-stream Two-stage trackers consist of two identical and individual Transformer-based tracking pipelines to extract the features of the target template and search region. Another Transformer network is then employed to find the relationship between these features. Finally, a prediction head is utilized to locate the target by using the attended features.

\begin{figure}[t]%
	\centering
	\includegraphics[width=0.47\textwidth]{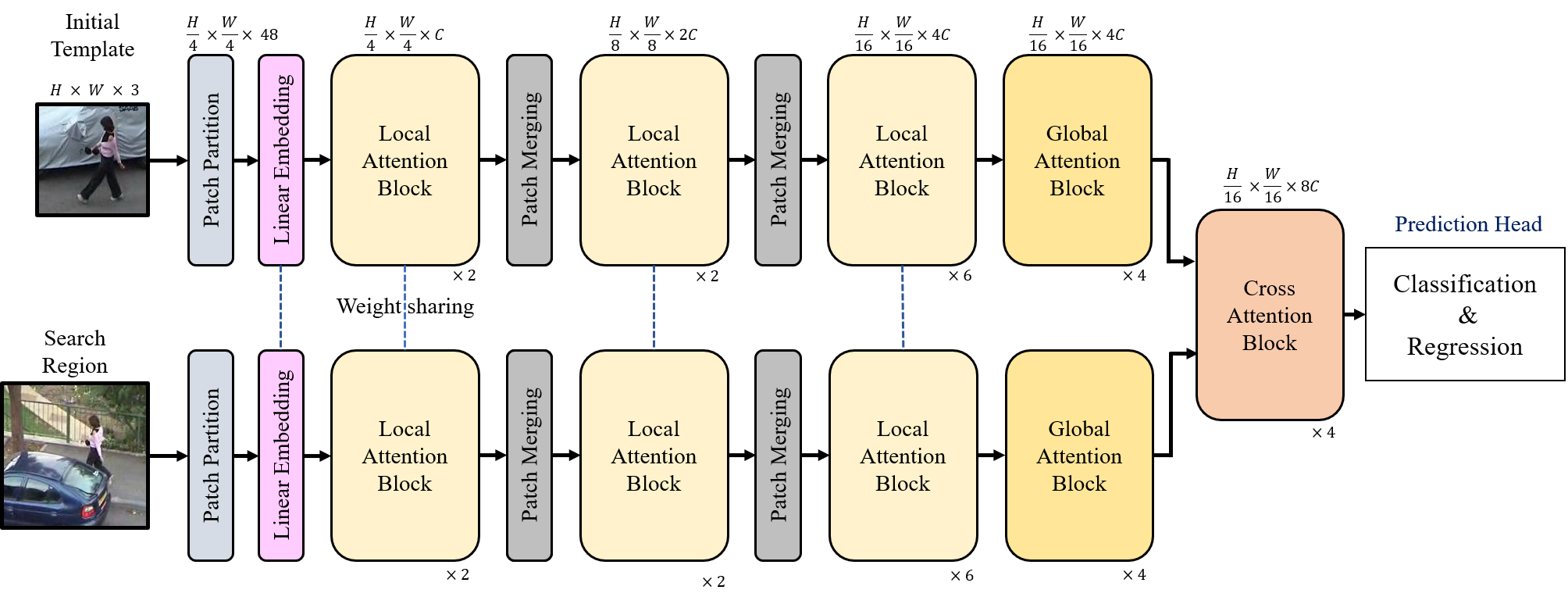}
	\caption{Architecture of the  DualTFR tracker \cite{xie2021learning}. It has a set of local attention blocks and a global attention block to capture the attention in  a small sized window and whole image, respectively. At the end of the network,  a cross-attention block is used to fuse the features of both branches.  }
	\label{DualTFR-Architecture}
\end{figure}

The first fully-Transformer based Two-stream Two-stage tracker is proposed by Xie  \emph{et al.} \cite{xie2021learning}, and it is referred to as DualTFR. In this tracker, template and search region images are split as tokens and then fed to the corresponding feature extraction branches.  DualTFR has a set of local attention blocks (LAB) in feature extraction branches to capture the attention in a small size window. Then the extracted features are fed to a global attention block (GAB) to capture the long range dependencies. Finally, as demonstrated in Fig. \ref{DualTFR-Architecture}, output features of both branches are fed to a cross-attention block to compute the attention between target template and search region. 
Since  the LAB computes attention within a small window of tokens on the high-resolution feature maps and the GAB computes the attention among all tokens of the same picture on the low-resolution feature maps, DualTFR tracker successfully achieved the high accuracy while maintaining above real-time speed. 

\begin{figure}[t]%
	\centering
	\includegraphics[width=0.47\textwidth]{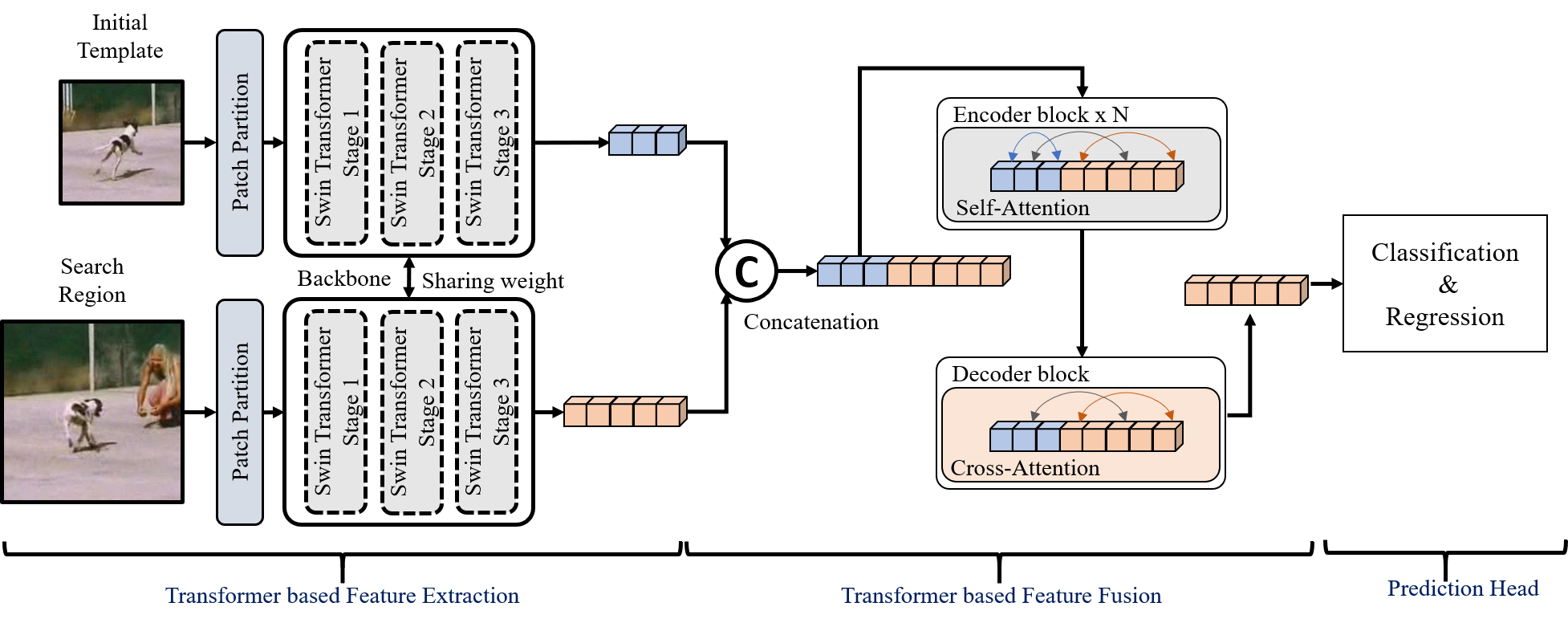}
	\caption{Architecture of the  SwinTrack tracker \cite{bib38}. It utilized the Swin Transformer \cite{bib34} for feature extraction. Then the features are concatenated with their positional encoding and then to fed to the encoder and decoder to enhance them using self-attention and cross-attention, respectively.    }
	\label{SwinTrack-Architecture}
\end{figure}

Lin  \emph{et al.} \cite{bib38} also used a two-branch fully-Transformer architecture for VOT and it is referred to as  SwinTrack. They have utilized the Swin Transformer \cite{bib34} for feature extraction since it showed better performance with fewer computational cost in object detection. In Swin Transformer, images and features are partitioned into non-overlapping windows and then the self-attention is computed within the window. Also,  windows are shifted in the next layers to maintain the connectivity. After the feature extraction using  the Swin Transformer,  features of the template and search region are concatenated as shown in Fig. \ref{SwinTrack-Architecture}. Then the concatenated features and their corresponding positional embedding are fed to the Transformer encoder to enrich them using the self-attention mechanism. Then a Transformer decoder block is used to find the relationship between the template and search region features using the cross-attention mechanism. Finally, the target is located by feeding the features to a prediction head, which has a binary classifier and a bounding box regression module. Based on the tracking accuracy, SwinTrack outperformed CNN-Transformer based trackers, and CNN-based trackers in several benchmark datasets. 

While the self-attention mechanism of the Transformer captures long-range dependencies, it may not adequately capture target-specific cues, making it susceptible to distraction by background clutter. Fu \emph{et al.} \cite{bib77} noticed this issue and then proposed a tracker, referred to as SparseTT, by using a sparse attention module to provide the target-specific attention in the search region features. In this tracker, the Swin Transformer is used to extract the target template and search region features and then fed them to a sparse Transformer. Encoder blocks of the sparse Transformer are similar to the other Transformer architectures.  Sparse multi-head self-attention modules are used in the decoder blocks to focus the foreground regions. Finally, enhanced features are fed to a double-head prediction network to detect the target with bounding box coordinates.  

In overall,   Two-stream Two-stage  fully-Transformer trackers used a backbone Transformer model to extract the features of the target template  and search region. Another Transformer is used to fuse and enhance the extracted futures. Finally, a small prediction network is used to locate the target. All of the   Two-stream Two-stage trackers have a simple and neat tracking architecture and showed better performance than the CNN-based trackers and CNN-Transformer based trackers since they fully utilized the attention mechanism of the Transformer for tracking a target. 

\subsubsection{One-stream One-stage Trackers}\label{subsubsec1}
\begin{figure}[t]%
	\centering
	\includegraphics[width=0.47\textwidth]{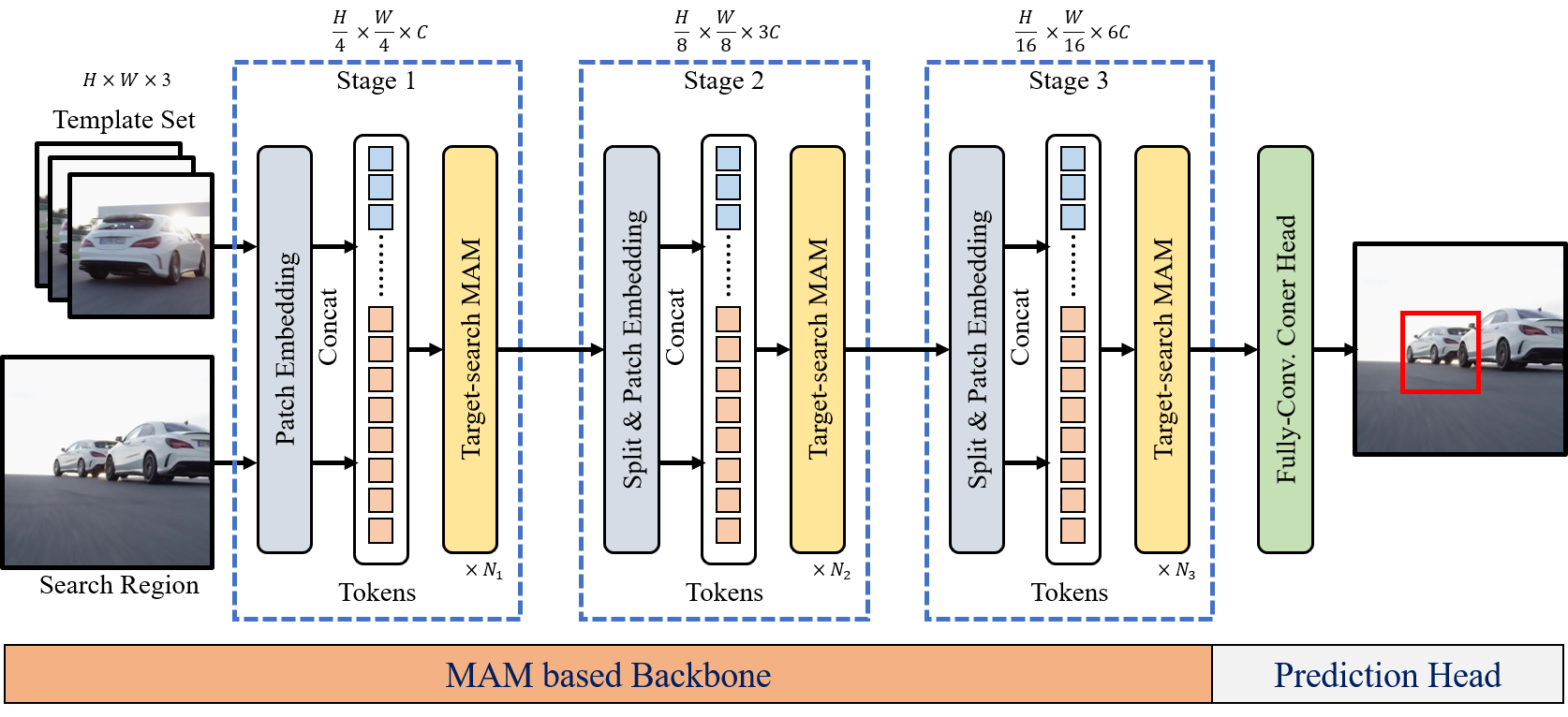}
	\caption{Architecture of the  MixFormer Tracker \cite{bib80}. It has a set of Mixed Attention Modules (MAM) to simultaneously extract and integrate the features of the target template and search region and to extract the target-specific discriminative cues.  }
	\label{MixFormer-Architecture}
\end{figure}

One-stream One-stage Trackers have a single pipeline of fully-Transformer based network architecture. Also, feature extraction and feature fusion process are done in a single stage in these approaches instead of two stages in the previously mentioned tracking methods. 

Recently, Cui \emph{et al.} \cite{bib80} found that combining the feature extraction and feature fusion processes is important for object tracking, as it enables extraction of more target-specific cues in the search region and improves correlation.
Based on this fact, they proposed a fully-Transformer based One-stream One-stage tracker referred to as MixFormer. As illustrated in Fig. \ref{MixFormer-Architecture}, a set of Mixed Attention Modules (MAM) are used in the MixFormer tracker to simultaneously extract and integrate the features of the target template and search region. MixFormer Tracker consumes multiple target templates and search regions are the inputs and locate the target using a simple fully-convolutional based prediction head network. Also, MixFormer  utilized the pre-trained CVT Transformer \cite{bib99} to design the MAM modules since CVT is excellent to capture the local and global feature dependencies in an image. Instead of the self-attention mechanism of the CVT, MAM  employs a dual attention mechanism on target template and search region tokens to capture the target-specific and search-specific cues, respectively. In addition, an asymmetric mixed attention technique is used in the MAM modules to reduce the computational cost by eliminating the unnecessary cross-attention between the tokens of the target and search region. Based on the reported results, MixFormer showed excellent tracking performance in benchmark datasets. However, MixFormer showed poor tracking speed since the MAM modules are computationally inefficient.  

\begin{figure}[t]%
	\centering
	\includegraphics[width=0.47\textwidth]{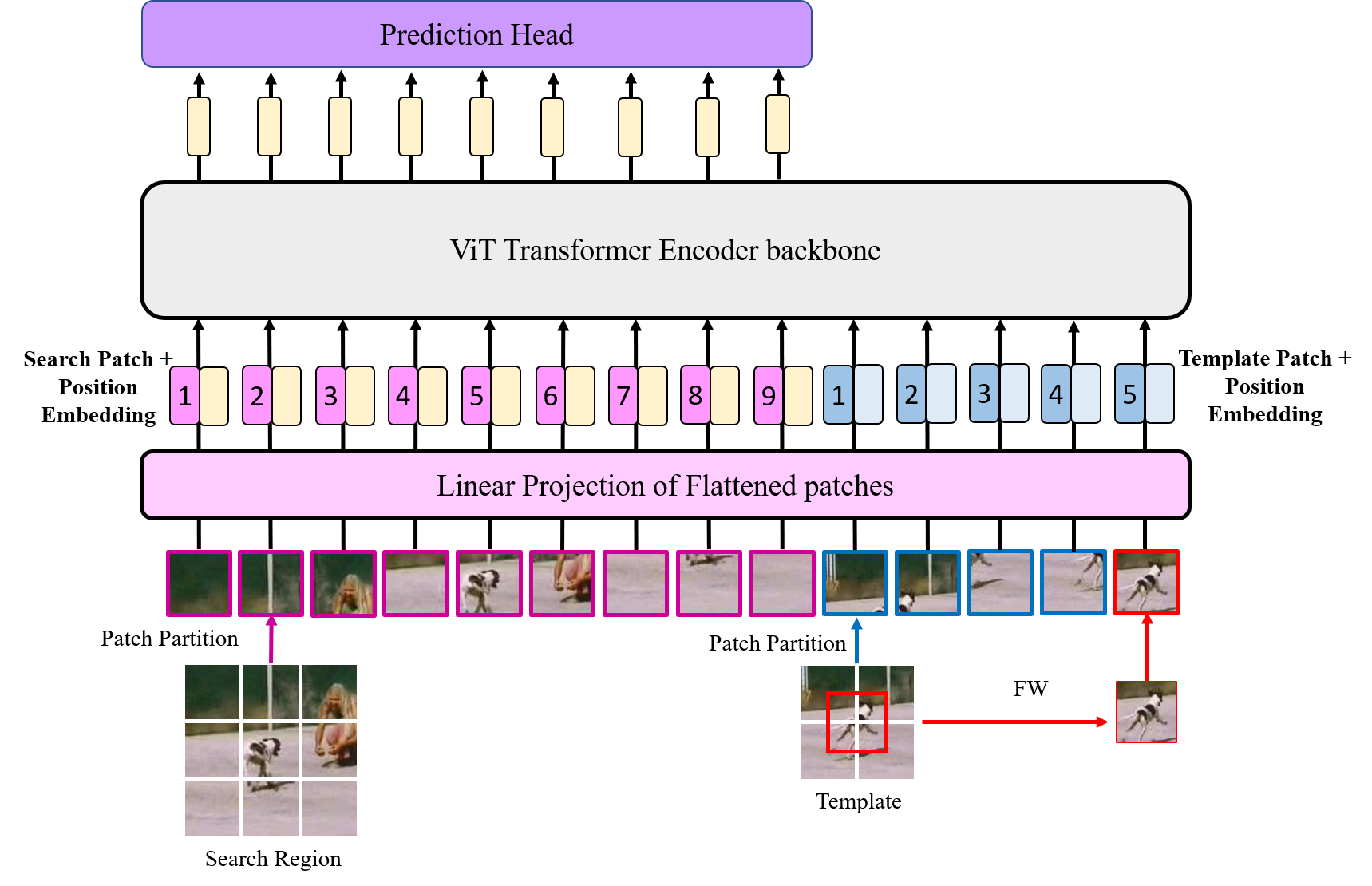}
	\caption{Architecture of the  SimTrack Approach \cite{bib79}. In this tracker, a target template, a search region, and an exact target region from the template are split as tokens and then fed to the   ViT \cite{bib33} backbone to locate the target.   }
	\label{SimTrack-Architecture}
\end{figure}

Another One-stream One-stage tracker is proposed by Chen \emph{et al.} \cite{bib79} and it is referred to as SimTrack. In this tracker, as shown in Fig. \ref{SimTrack-Architecture}, the pre-trained ViT \cite{bib33} model is utilized as the backbone Transformer to combine feature extraction and fusion. In SimTrack approach, target template and search region are split as a set of tokens, concatenated, and then fed to the backbone Transformer with their positional embedding. Since the target template tokens contain some background regions due to the splitting process,  SimTrack followed a foveal windowing technique to accurately capture the target-specific cues. In the foveal windowing technique, a smaller region of the template image is cropped with the target in middle and then serialized into image tokens.  In addition to the tokens of target template and search region, foveal sequence also fed to the Transformer to capture more target-specific features.

\begin{figure}[t]%
	\centering
	\includegraphics[width=0.47\textwidth]{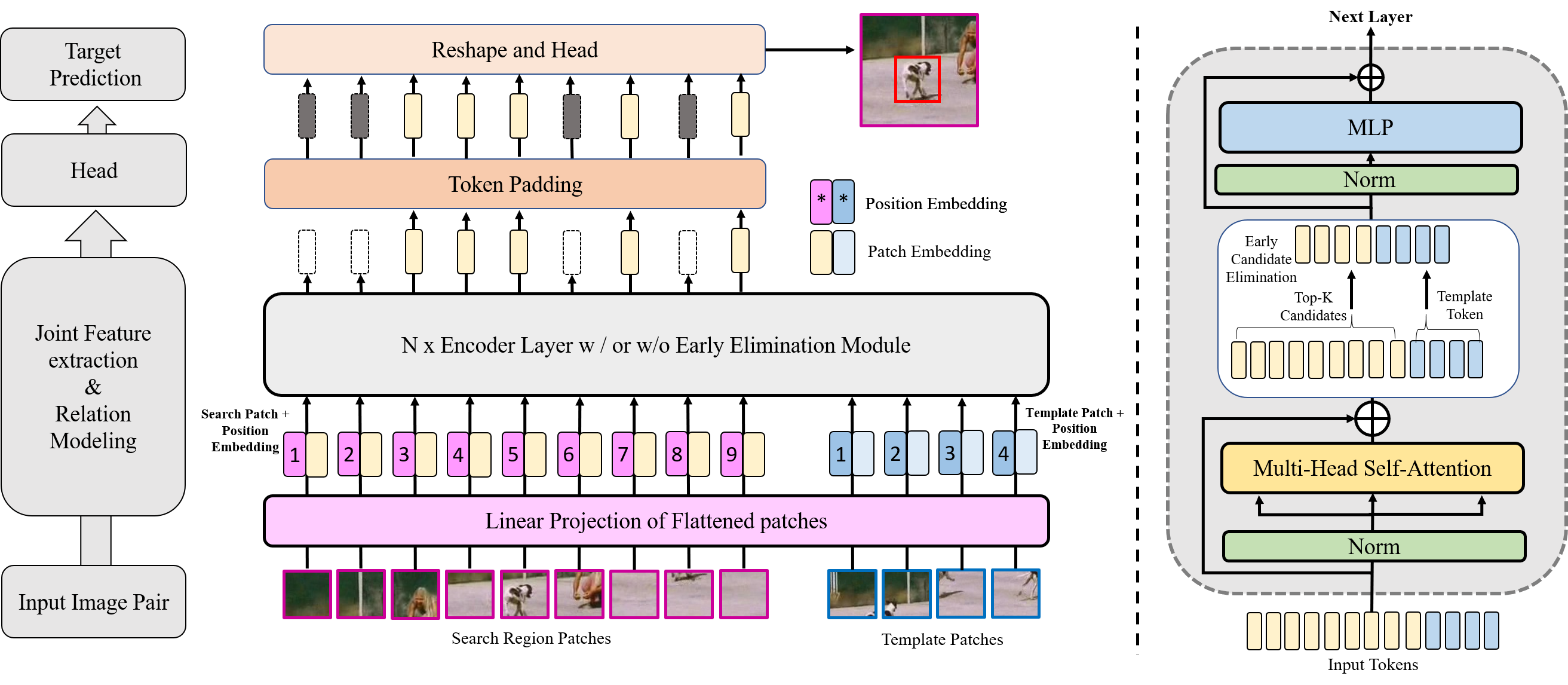}
	\caption{Architecture of the  OSTrack  Approach \cite{bib78}. It used an early elimination module in encoder layers  to identify and remove the features of the background tokens from the search image and hence the performance of the tracker is increased.  }
	\label{OSTrack-Architecture}
\end{figure}

OSTrack \cite{bib78} is another One-stream One-stage approach that combines the feature learning and feature fusion processes using the ViT backbone, as demonstrated in Fig. \ref{OSTrack-Architecture}. OSTrack utilizes a self-supervised learning-based Masked Auto Encoder (MAE) \cite{he2022masked} pre-trained model to initialize the ViT backbone. The authors of OSTrack have found that some of the tokens from the search image contain background information, and hence including these tokens in the tracking process is unnecessary. Based on this fact, OSTrack includes an early candidate elimination module in some of the encoder layers to remove tokens containing background information. Due to the candidate elimination module, the tracking speed and accuracy of OSTrack are increased. Since OSTrack efficiently uses the information flow between the features of the target template and search region, target-specific discriminative cues are extracted, and unnecessary background features are eliminated. Therefore, it has shown excellent tracking performance with high tracking speed in several benchmark datasets.

\begin{figure}[b]%
	\centering
	\includegraphics[width=0.47\textwidth]{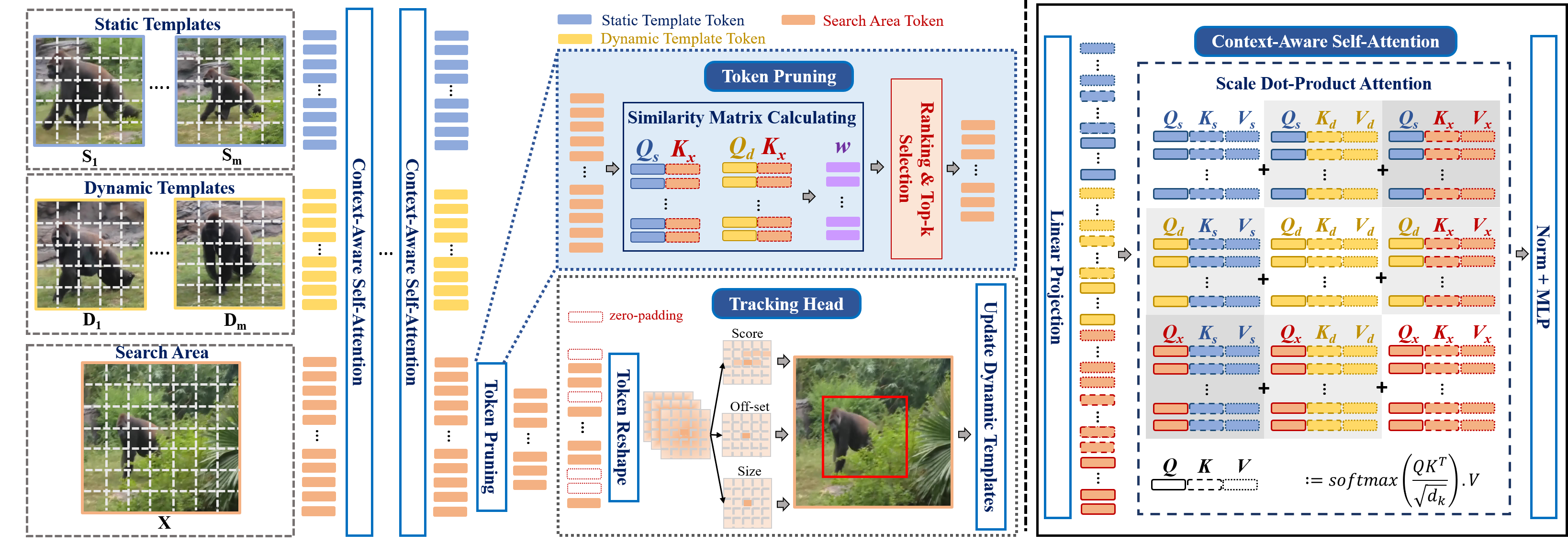}
	\caption{Architecture of the  ProContEXT Tracker \cite{lan2022procontext}. Spatial and temporal cues are captured using a context-aware self-attention module by feeding a set of static and dynamic target templates. Also, an upgraded candidate token elimination module is used in this tracker. }
	\label{ProContEXT}
\end{figure}

Due to the great success of OSTrack in the tracking community, several recent follow-up approaches \cite{lan2022procontext, xie2023videotrack, gao2023generalized, wu2023dropmae} have been proposed, utilizing self-supervised learning based MAE pre-trained model to initialize the backbone network. Lan \emph{et al.} \cite{lan2022procontext} upgraded the performance of OSTrack by using a candidate token elimination module and including a set of dynamic templates to adopt the temporal features of the target. Their tracker, referred to as ProContEXT, is shown in Fig. \ref{ProContEXT}. The ProContEXT tracker focuses on capturing the spatial and temporal cues of the target templates by utilizing a context-aware self-attention module. In this tracker, static target templates, dynamic target templates with spatial and temporal cues, and the search region are split and then fed to the attention module. Additionally, the candidate token elimination module of ProContEXT performs better than OSTrack as it incorporates temporal cues in the removal of background tokens. Based on the reported results, the ProContEXT approach outperformed OSTrack and demonstrated state-of-the-art performance in tracking benchmarks.

The success of the One-stream One-Stage trackers lies in the information flow between the features of target template patches and search region patches.  Although the attention operation is used to identify target-specific features in search regions, it is not necessary to calculate the attention between the template patches and all of the search patches
since patches containing background and distractor features can degrade the efficiency and accuracy of the tracker.  Although the OSTrack \cite{bib78} and ProContEXT \cite{lan2022procontext} trackers incorporate modules to eliminate background patches in some encoder layers, there is a need for a more robust and generalized mechanism. Recently, Gao \emph{et al.} \cite{gao2023generalized} have proposed a Generalized Relation Modeling (GRM) mechanism to address this limitation in One-stream One-stage trackers. In the first stage of GRM tracker, a simple MLP network is utilized to determine the probabilities of target template features in search region patches. Subsequently, the search region patches are divided into two groups, with only the high probability search patches allowed to interact with the template patches in the encoder layers. This adaptive selection of search region patches enhances the flexibility of relation modeling and improves discrimination capabilities. The experimental results demonstrate that the GRM tracker outperforms OSTrack with an approximately 2\% higher success rate and improved tracking speed. 

The state-of-the-art One-stream One-stage trackers \cite{bib78, bib79, bib80} do not fully utilize the temporal features, making them less robust for appearance changes. Recently, Xie \emph{et al.} \cite{xie2023videotrack} identified this limitation and proposed an end-to-end Transformer tracker called VideoTrack, which aims to exploit the temporal context. VideoTrack captures the spatio-temporal features of the tracking sequence by feeding patches from a set of adjacent frames to the Transformer. The VideoTrack approach encodes inter-frame and intra-frame patches using distinct encoding schemes and then feeds them to a Transformer encoder to capture spatio-temporal features. Due to the spatio-temporal feature learning, VideoTrack outperforms OSTrack\cite{bib78} and MixFormer \cite{bib80} approaches with a considerable margin.

\begin{figure}[t]%
	\centering
	\includegraphics[width=0.47\textwidth]{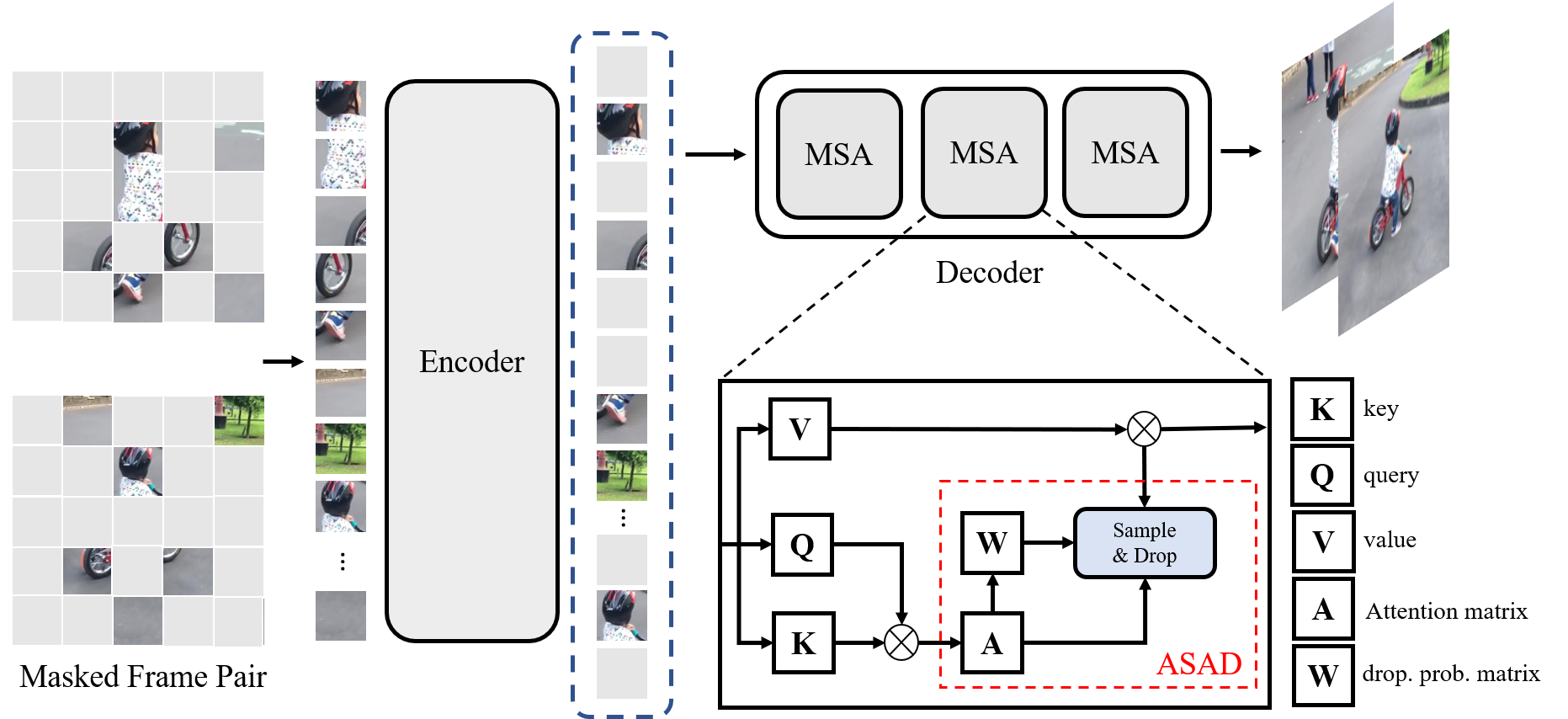}
	\caption{Architecture of the  DropMAE \cite{wu2023dropmae} masked auto encoder. It is used to capture the spatial cues within an individual image and to capture the correlated spatial cues between target and search images. The OSTrack \cite{bib78} approach showed better performance while fine-tuning the DropMAE as the backbone compared to the MAE \cite{he2022masked} backbone. }
	\label{DropMAE-Architecture}
\end{figure}

Most single-object tracking approaches traditionally rely on pre-trained models initialized through supervised learning. However, recent fully Transformer-based trackers, such as OSTrack \cite{bib78}, ProContEXT \cite{lan2022procontext}, GRM \cite{gao2023generalized}, and VideoTrack \cite{xie2023videotrack}, have demonstrated that initializing the backbone with a self-supervised learning based Masked Autoencoder (MAE) \cite{he2022masked} pre-trained model can achieve higher tracking accuracy compared to models based on supervised learning. This improvement can be attributed to the MAE's ability to capture fine-grained local structures within an image, which are essential for accurate target localization.  Recently, Wu \emph{et al.} \cite{wu2023dropmae} discovered that the MAE architecture exhibits a lack of robustness when applied to feature matching tasks between two images. As a result, a modified variant of the MAE architecture named DropMAE was introduced, as depicted in Figure \ref{DropMAE-Architecture}, which is specifically designed to improve feature matching for tracking and segmentation tasks. DropMAE captures spatial cues within individual images and also captures the correlated spatial cues between two frames by randomly masking the input frames and processing them through the encoder-decoder architecture. Additionally, DropMAE follows an attention dropout mechanism that restricts the interaction between tokens within the same frame. This restriction leads to increased interaction between the tokens of search and target image pairs, resulting in more reliable capture of temporal cues.  The experimental results demonstrated that the OSTrack \cite{bib78} approach achieved superior performance when utilizing DropMAE to initialize the backbone compared to initializing with MAE backbone. 

Closely related to DropMAE, another masked encoder-based pre-trained model has been specifically designed and trained for the tracking task, referred to as MAT \cite{Zhao2023CVPR}. Similar to DropMAE, MAT randomly masks patches of template and search image pairs, which are then jointly processed by the encoder to capture their visual representations.  After encoding, in contrast to the single decoder of other models \cite{he2022masked, wu2023dropmae}, MAT employs two identical decoders to separately reconstruct the search image and the target  region in the search image. The encoded features of the target template are used to  reconstruct the target  region in the search image that enables the capturing of tracking-specific representations in the search image.  Finally, a simple and lightweight tracker (MATTrack) is designed to evaluate the representation of the MAT masked encoder.

All present CNN-based and fully-Transformer based approaches treat object tracking as a template matching problem between the target image and the search region. Although some approaches follow a template update mechanism, the temporal dependencies between the frames are neglected or not fully investigated. Recently, Wei \emph{et al.} \cite{wei2023autoregressive} proposed a tracker called ARTrack, which treats object tracking as a coordinate sequence interpretation problem. Taking inspiration from language modeling, ARTrack interprets target trajectories by learning from previous states in a sequence of frames. The tracking pipeline of ARTrack differs from other One-stream One-stage Transformer trackers as it employs an encoder-decoder architecture and lacks a prediction head. By treating tracking as a coordinate sequence interpretation task and adopting a general encoder-decoder architecture, ARTrack simplifies the tracking pipeline and achieves superior performance compared to existing trackers that rely on prediction heads. 

\begin{figure}[t]%
	\centering
	\includegraphics[width=0.47\textwidth]{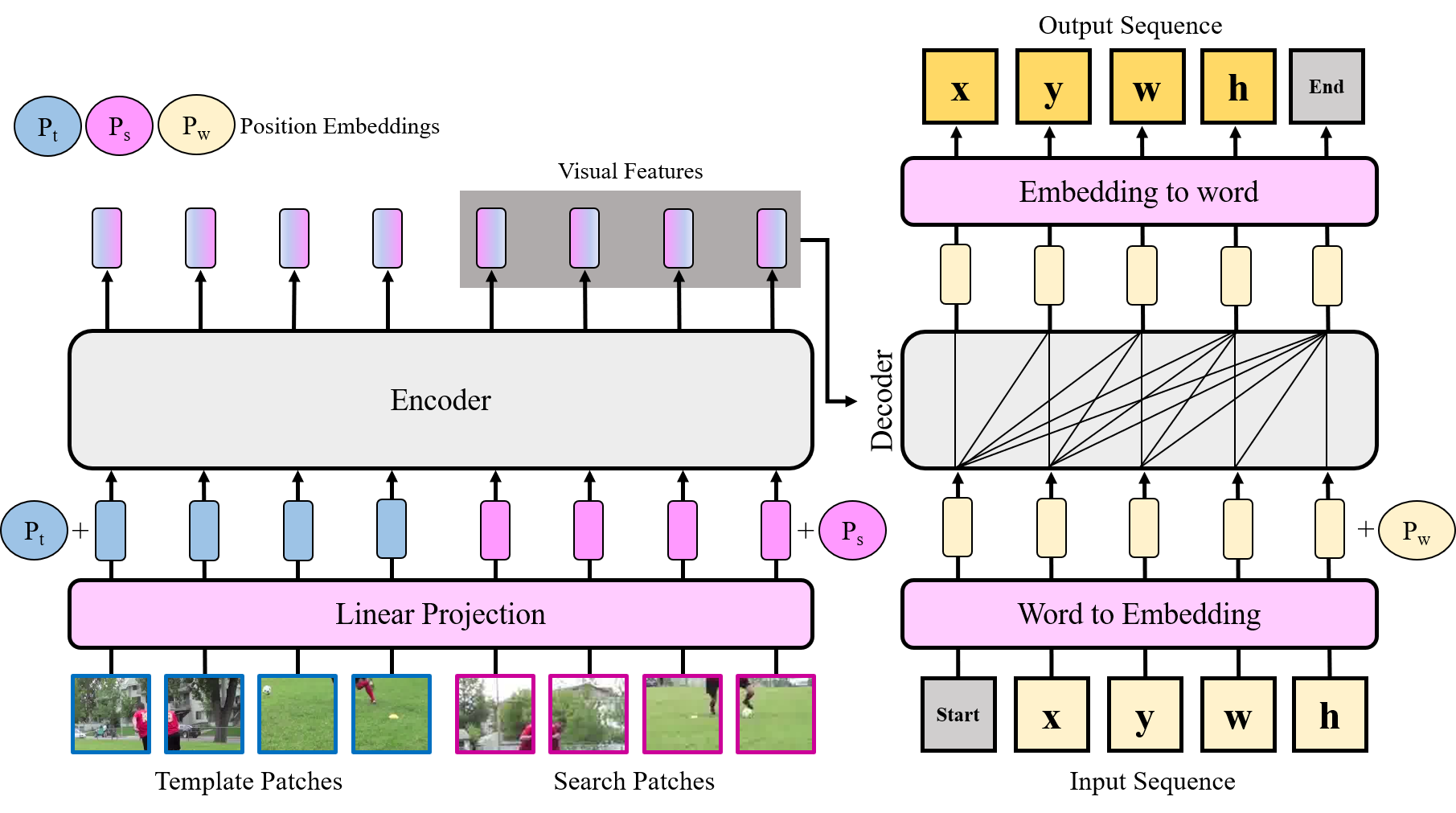}
	\caption{Architecture of the  SeqTeack \cite{chen2023seqtrack} approach. It treats the tracking as a sequence learning  problem instead of template matching.  Without a 	prediction head, it is able to locate a target using an end-to-end Transformer architecture. }
	\label{SeqTrack}
\end{figure}

Closely related to the ARTrack\cite{wei2023autoregressive}, the recently proposed SeqTrack \cite{chen2023seqtrack} tracker also considers tracking as a sequence learning problem. This tracker utilizes an encoder-decoder Transformer architecture to extract visual features and autoregressively generate the bounding box coordinates of the target, respectively. Different from ARTrack, in SeqTrack, the predicted bounding box values rely solely on the target stage in the previous frame, achieved through the use of an attention masking technique in the decoder. Experimental results shows that SeqTrack better performance than the template matching based trackers.

In summary, One-stream One-stage trackers combined the feature extraction and feature fusion process using a fully-Transformer architecture. In these trackers, target template and search region images are split as tokens and concatenated with their positional embedding, and then fed to a Transformer. Since these trackers extract features with a single Transformer network, the features of the template  and search regions are efficiently integrated, leading to more discriminative features being identified and unnecessary features being eliminated. Based on these facts, fully-Transformer based One-stream One-stage trackers have shown outstanding performance compared to other types of trackers on all benchmark datasets.

\section{Experimental Analysis}  \label{sec5}
In the last two decades, numerous approaches have been proposed for single object tracking. Since their performance is evaluated on various benchmark datasets using different evaluation metrics, it is crucial to experimentally evaluate these approaches to identify future directions, particularly after the introduction of Transformers in VOT. This study focuses on evaluating the tracking robustness and computational efficiency performances of CNN-Transformer based trackers and fully-Transformer based trackers. Additionally, we have included 12 recently proposed CNN-based trackers in this experimental evaluation study. Five benchmark datasets are utilized, and their details and evaluation metrics are described in Section \ref{datasets}. The results of tracking robustness and efficiency are presented in Section \ref{performance} and Section \ref{efficiency}, respectively.

\subsection{Benchmark Datasets and Evaluation Metrics}\label{datasets}
Several benchmark datasets are constructed and publicly available for VOT. Each dataset is different from the others based on the target object class, number of sequences, annotation method, length of the tracking sequence, attributes, and complexity. Additionally, these datasets follow various performance metrics to measure the trackers' performance. To analyze the performance of recent trackers, we have selected five benchmark datasets: OTB100, UAV123, LaSOT, TrackingNet, and GOT-10k. The details of these datasets are summarized in Table ~\ref{re:4}. 

\begin{table*}
	\caption{Details of the datasets used in experimental analysis study. The abbreviations are denoted as SV for scale variation, ARC for aspect ratio change, FM for fast motion, LR for low resolution, OV for out-of-view, IV for illumination variation, CM for camera motion, MB for motion blur, BC for background clutter, SOB for Similar Object, DEF for deformation, ROT for rotation, IPR for in-plane rotation, OPR for out-of-plane rotation, OCC for occlusion, FOC for full occlusion, POC for partial occlusion, and VC for viewpoint change.}\label{re:4}%
	\begin{minipage}{\textwidth}			
		\centering		\footnotesize	
		\begin{tabular}{l l l l l l p{4cm} p{3cm}}
			\toprule
			Dataset	&	Year	&	No of	&	No of	&	Average 	&	Absent	&	Attributes	&	Evaluation \\
			& & Sequences  & Classes & Frames & Label & & Metrics\\
			\midrule
			OTB100 \cite{bib59}   & 2015   & 100  & 16 & 598 & No & SV, FM, LR, OV, IV, MB, &  Precision, Success \\
			& & & & & & BC, DEF, IPR, OPR, OCC  & \\
			UAV123 \cite{mueller2016benchmark}   & 2016   & 123  & 9 & 915 & No & SV, FM, LR, OV, IV, CM, &  Precision, Success \\
			& & & & & &  BC, SOB, OCC, POC, FOC & \\
			TrackingNet \cite{bib72}   & 2018   & 30643  & 27 & 471 & No & SV, ARC, FM, LR, OV, IV,  & Precision, Success,  \\
			& & & & & & CM, MB, BC, SOB, DEF, IPR,  & Normalized Precision\\
			& & & & & & OPR, POC, FOC & \\
			LaSOT  \cite{bib42}  & 2019 & 1400 & 70 & 2506 & Yes & SV, ARC, FM, LR, OV, IV, & Precision, Success,\\
			& & & & & & CM, MB, BC, DEF, ROT, FOC,  & Normalized Precision \\
			& & & & & & POC, VC & \\
			GOT-10k \cite{bib62} & 2019 & 10000 & 563 & 149 & Yes & SV, ARC, FM, LR, IV, OCC & Average Overlap, \\
			& & & & & & & Success Rate\\
			\bottomrule
		\end{tabular}
	\end{minipage}
\end{table*}

OTB was the first widely used benchmark dataset in VOT. It gained widespread acceptance among researchers due to its simplicity and better evaluation scheme.  It is initially proposed with 51 short tracking video sequences and they are referred to as OTB2013 \cite{bib71}. Few years later, another 49 sequences are included into the benchmark and then the entire sequences are referred to as OTB100 \cite{bib59}. We have selected OTB100 to evaluate the performance of the trackers with various tracking attributes (scenarios). Similar to other researchers, precision and success plots are used to measure the performance of a tracker in OTB100. Precision in OTB100 is computed by calculating the distance between the midpoints of the tracked target bounding box and the corresponding ground truth bounding box. The precision plot is used to show the proportion of frames in which the tracking locations are within a threshold distance. In this study, threshold of 20 pixel distance is used to rank the trackers based on their precision scores. 
Success of a tracker is measured in OTB100 by measuring the average overlap scores and it is computed as follows:

\begin{equation}
	\mathrm{Overlap\_Score} = \frac{\mid B_t \cap B_g \mid}{\mid B_t \cup B_g \mid},
\end{equation}
where $B_t$ and $B_g$ are the bounding boxes of tracked target and ground-truth, respectively. Also, $\cap$ and $\cup$ are denoting intersection and union operations, respectively. The area-under-curve (AUC) scores are used to rank the trackers in success plots, representing the average number of successful frames when the overlapping scores change from 0 to 1.

UAV123 \cite{mueller2016benchmark} dataset contains a total of 123 video sequences from an aerial viewpoint. All the video sequences of UAV123 are captured by using unmanned aerial vehicles (UAV). Compare to OTB100 and other benchmark datasets, target object is small in UAV123 and the tracking sequences have several distractor objects and long occlusions. Since both the OTB100 and UAV123 datasets have a small number of sequences with fewer frames, they are not sufficient to train deep learning-based tracking architectures. We have used the UAV123 dataset to evaluate the aerial tracking performances of  trackers and to measure their occlusion handling capability.  UAV123 dataset follows the same evaluation scheme of OTB100 to compare the performance of trackers. 

TrackingNet \cite{bib72} is the first large scale benchmark dataset and contains more than 30K video sequences. Tracking video sequences of TrackingNet are sampled from real-world YouTube videos and 30,132 and 511  numbers of sequences are allocated for training and testing, respectively. The TrackingNet dataset contains a rich distribution of target object classes, with almost 14 million bounding boxes. The average length of tracking sequences is less than 500 frames, and hence the TrackingNet dataset is not suitable to evaluate a tracker's long-term tracking capability. However, we have used the TrackingNet dataset to evaluate the performance of trackers in real-world scenarios across a wide range of classes. Similar to the OTB100, TrackingNet dataset also used the precision and success plots to evaluate the performance of trackers. In addition to these metrics, normalized precision is used since the precision is depending on the sizes of the images and bounding boxes. The normalized precision ($P_n$) metric is computed as:

\begin{equation}
	P_n = \lVert W (B_t - B_g) \rVert _2
\end{equation}
where $W$ denotes the size of the bounding box. 

Recently, the large-scale single object tracking (LaSOT) benchmark \cite{bib42} has been developed, and it is one of the largest tracking datasets to date. On average, more than 2500 frames are in a tracking sequence in LaSOT, and the total duration of video sequences are more than 32 hours. In addition, target objects are annotated with detail information such as absent label, in which out-of-view and fully occluded target are labeled. Furthermore, the LaSOT dataset maintained the class balance in video sequences by containing equal number of video sequences for each target class. Similar to the TrackingNet dataset, LaSOT dataset used the precision, success, and normalized precision metrics to measure the performance of the trackers. In this experimental analysis study, LaSOT benchmark is mainly used to evaluate the long-term tracking performances of the trackers.

GOT-10k \cite{bib62} is another large scale tracking benchmark and it has 10,000 video sequences which were captured from real-world scenarios.  Compare to other benchmark datasets, GOT-10k has a huge number of target object classes and almost 1.5 million bounding boxes are used in annotation. Furthermore, the annotations include motion and absent labels, allowing for the efficient evaluation of trackers that can handle motion and occlusions. The average overlap (AO) and success rate are used to evaluate the trackers in GOT-10k and they are calculated by measuring the average overlaps between $B_t$ and $B_g$, and the percentage of correctly tracked frames, respectively.  We have evaluated the trackers on GOT-10k to measure 
the one-shot tracking performances of the trackers since the training object classes are not overlapped with testing object classes.

\subsection{Tracking Performance Analysis}\label{performance}
We have presented the qualitative results of the trackers in this section. To conduct an unbiased evaluation and compare their tracking performances, we reproduced the success and precision scores of the trackers and their attribute-wise tracking results using their source codes. Additionally, we considered the reported results of a few recent trackers since their source codes are unavailable.
\begin{table*}
	\footnotesize
	\begin{center}
		\caption{Tracking robustness comparison of the recently proposed state-of-the-art approaches on OTB100, UAV123,  TrackingNet, LaSOT, and GOT-10k benchmarks. The abbreviations are denoted as $\mathrm{SR}$: Success Rate, $\mathrm{P}$:Precision, $\mathrm{P_n}$: Normalized Precision, $\mathrm{AO}$: Average Overlap, $\mathrm{SR_{0.5}}$: Success Rate at $0.5$ threshold, and  $\mathrm{SR_{0.75}}$: Success Rate at $0.75$ threshold. All the results are reported in percentage. The evaluation considered the top-performing models of each tracker. The red, blue, and green colors are used to mark the top three results, respectively. The symbols \dag and \ddag are denoting One-stream One-stage trackers, and Two-stream Two-stage trackers, respectively. The $\ast$ symbol is used to indicate tracking models that are only trained on the training set of GOT-10k.  }\label{re:3}%
		\begin{tabular}{lll@{\extracolsep{0.40cm}}l@{\extracolsep{0.8cm}}l@{\extracolsep{0.30cm}}l@{\extracolsep{0.8cm}}l@{\extracolsep{0.30cm}}l@{\extracolsep{0.30cm}}l@{\extracolsep{0.8cm}}l@{\extracolsep{0.30cm}}l@{\extracolsep{0.30cm}}l@{\extracolsep{0.8cm}}l@{\extracolsep{0.30cm}}l@{\extracolsep{0.30cm}}l}
			\midrule
			&\multirow{2}{*}{\textbf{Trackers}} &  \multicolumn{2}{c}{\textbf{OTB100}} & \multicolumn{2}{c}{\textbf{UAV123}} & \multicolumn{3}{c}{\textbf{LaSOT}} &
			\multicolumn{3}{c}{\textbf{TrackingNet}} & \multicolumn{3}{c}{\textbf{GOT-10k  $\ast$}}\\ 
			\cmidrule{3-4} \cmidrule{5-6} \cmidrule{7-9} \cmidrule{10-12} \cmidrule{13-15}
			& &SR  & P & SR & P & SR & $\mathrm{P_{n}}$ & P &  SR & $\mathrm{P_{n}}$ & P & AO & $\mathrm{SR_{0.5}}$ & $\mathrm{SR_{0.75}}$ \\
			\midrule
			\multirow{11}{*}{\rotatebox[origin=c]{90}{\parbox[c]{3.5cm}{\centering Fully-Transformer based Trackers}}}
			& SeqTrack \cite{chen2023seqtrack}\dag & 68.3 & 89.1 & 68.5 & 89.1 & \color{red}72.5 & \color{red}81.5 & \color{red}79.2	& \color{red}85.5	& \color{red}89.8 &	\color{red}85.8 &	\color{blue}74.8 &	81.9 &	\color{blue}72.2
			\\
			& DropTrack \cite{wu2023dropmae}\dag & 69.5 & 91.0  &  \color{red}70.9 &  \color{red}92.4 & \color{blue}71.5 & \color{red}81.5 & \color{blue}77.9	&\color{green} 84.1	& \color{green}88.9 &	- &	\color{red}75.9 &	\color{red}86.8 &	\color{green}72.0
			\\
			& GRM \cite{gao2023generalized}\dag & 68.9 & 90.0 & 69.0 & 89.8 & 69.9 & 79.3 & 75.8	& 84.0	& 88.7 & \color{green}83.3 &	73.4 &	82.9 &	70.4
			\\
			& VideoTrack \cite{xie2023videotrack}\dag & - & - &	69.7 & 89.9 & 70.2 & - & 76.4 & 83.8	& 88.7 &	83.1 &	72.9 &	81.9 &	69.8
			\\
			& MATTrack \cite{Zhao2023CVPR}\dag & - & -  &  68.0 &  - & 67.8 & 77.3 &  -	& 81.9 & 86.8  & - & 67.7	 & 78.4	 & -	\\				
			& ProContEXT \cite{lan2022procontext} \dag & 66.9 & 91.1 & 69.2 & 90.0 & \color{green}71.4 & \color{blue}81.2 & \color{blue}77.9 & \color{blue}84.6 & \color{blue}89.2 & \color{blue}83.8 & \color{green}74.6 & \color{blue}84.7 & \color{red}72.9			
			\\
			& OSTrack \cite{bib78}\dag & 68.1 & 88.7 &	\color{blue}70.7 & \color{blue}92.3 & 71.1 & \color{green}81.1 & \color{green}77.6	& 83.9	& 88.5 &	83.2 &	73.7 &	\color{green}83.2 &	70.8
			\\
			& MixFormer \cite{bib80}\dag & 70.4 & 92.2 &	69.5 &	\color{green}90.9 &	70.0 &	79.9 &	76.3 &	83.9 &	\color{green}88.9 &	83.1 & 70.7 &	80.0 & 67.8		
			\\
			& SimTrack-B/16 \cite{bib79}\dag & 66.1 &	85.7 &	69.8 &	89.6 &	69.3 &	78.5 &	74.0 &	82.3 &	86.5 &	80.2 &	68.6 &	78.9 &	62.4
			\\
			
			& SparseTT \cite{bib77}\ddag & 69.8 & 90.5 &	68.8 &	88.0 &	66.0 &	74.8 &	70.1 &	81.7 &	86.6 &	79.5 &	69.3 &	79.1 &	63.8						
			\\
			
			& SwinTrack\cite{bib38}\ddag & 69.1 &	90.2 &	69.8 &	89.1 &	70.2 &	78.4 &	75.3 & 84.0 &	88.2 &	83.2 &	69.4 &	78.0 &	64.3				
			\\
			& DualTFR \cite{xie2021learning}\ddag & - & - & 68.2 & - & 63.5 & 72.0 & 66.5 & 80.1 & 84.9 & - & - & - & - 
			\\
			\midrule
			\multirow{13}{*}{\rotatebox[origin=c]{90}{\parbox[c]{3.5cm}{\centering CNN-Transformer based Trackers}}} & BANDT \cite{yang2023bandt} & 70.6 & 91.1 & 69.6 & 90.0 & 64.4 & 73.0 & 67.0 & 78.5 & 82.7 & 74.5 & 64.5 & 73.8 & 54.2 \\
			
			& AiATrack \cite{bib73} & 	69.6 &	91.7 &	69.3 &	90.7 &	69.0 &	79.4 &	73.8 &	82.7 &	87.8 & 80.4 & 69.6 &	80.0 &	63.2								
			\\
			& ToMP \cite{bib75} & 70.0 &	90.6 &	65.9 &	85.2 &	68.5 &	79.2 &	73.5 &	81.5 &	86.4 &	78.9 &	- &	- &	- 		
			\\
			& CSWinTT \cite{bib67} & 67.1 &	87.2 &	\color{green}70.5 &	90.3 &	66.2 &	75.2 &	70.9 &	81.9 &	86.7 &	79.5 &	69.4 &	78.9 &	65.4 
			\\
			& SiamTPN \cite{xing2022siamese} & 70.2 & 90.2 & 63.6 & 82.3 & 58.1 & 68.3 & 57.8 & 70.8 & 77.1 & 65.1 & 57.6 & 68.6 & 44.1 \\
			& UTT \cite{ma2022unified} & - & - & - & - & 64.6 & - & 67.2 & 79.7 & - & 77.0 & - & - & -
			\\
			& CTT \cite{zhong2022correlation} & - & - & - & - & 65.7 & 75.0 & 69.8 & 81.4 & 86.4 & - & - & - & -
			\\
			& DTT \cite{bib74} & - & - & - & - & 60.1 & - & - & 79.6 & 85.0 & 78.9 & 68.9 & 79.8 & 62.2
			\\
			& TrTr \cite{zhao2021trtr} & \color{red}71.2 & \color{red}93.1 & 63.3 & 83.9 & 52.7 & 61.9 & 54.4 & 70.7 & 79.5 & 67.4 & - & - & -
			\\
			& STARK \cite{bib36} & 68.0 &	88.4 &	68.5 &	89.5 &	67.1 &	76.9 &	72.2 &	82.0 &	86.9 &	79.1 &	68.8 &	78.1 &	64.1
			\\			
			& TransT \cite{bib37} & 69.5	& 89.9	& 68.1 & 87.6 &	64.9 &	73.8 &	69.0 &	81.4 &	86.7 &	80.3 &	67.1 &	76.8 &	60.9 
			\\
			& TrDiMP \cite{bib66} & 70.8 &	\color{green}92.5 &	67.0 &	87.6 &	63.9 &	73.0 &	66.2 &	78.4 &	83.3 &	73.1 &	67.1 &	77.7 &	58.3
			\\
			& HiFT \cite{bib76} & 61.4 & 81.4 & 58.9 & 78.7 & 45.1 & 52.7 & 42.1 & 66.7 & 73.8 & 60.9 & - & - & - \\
			\midrule
			\multirow{12}{*}{\rotatebox[origin=c]{90}{\parbox[c]{3.0cm}{\centering CNN-based Trackers}}} & KeepTrack \cite{mayer2021learning} & \color{green}70.9 &	92.2 &	69.1 &	\color{green}90.9 &	67.3 &	77.4 &	70.4 &	78.1 &	83.5 &	73.8 &	68.3 & 79.3 & 61.0		
			\\
			& SiamGAT \cite{guo2021graph} & \color{blue}71.0 &	91.6 &	64.6 &	84.3 &	53.9 &	63.3 &	53.0 &	75.3 &	80.7 &	69.3 &	62.7 &	74.3 &	48.8
			\\
			& SiamRN \cite{cheng2021learning} & 70.1 &	\color{red}93.1 &	64.3 &	86.1 &	52.7 &	63.5 &	53.1 &	- &	- &	- &	- &	- &	-			
			\\
			& SiamR-CNN \cite{voigtlaender2020siam}  & 70.1 &	89.1 &	64.9 &	83.4 &	64.8 &	72.2 &	68.4 &	81.2 &	85.4 &	80.0 &	64.9 &	72.8 &	59.7						
			\\
			& PrDiMP-50 \cite{danelljan2020probabilistic} & 69.7	& 89.7 &	66.9 &	87.8 &	59.9 &	69.0 &	60.8 &	75.8 &	81.6 &	70.4 &	63.4 &	73.8 &	54.3			
			\\
			& Ocean \cite{zhang2020ocean} & 68.4 &	92.0 &	62.1 &	82.3 &	56.0 &	65.1 &	56.6 &	69.2 &	79.4 &	68.7 &	61.1 &	72.1 &	47.3						
			\\
			& SiamFC++ \cite{xu2020siamfc++} & 68.3 &	89.6 &	62.3 &	81 &	54.3 &	62.3 &	54.7 &	75.4 &	80.0 &	70.5 &	59.5 &	69.5 &	47.3			
			\\			
			& SiamBAN \cite{chen2020siamese} & 69.6 &	91.0 &	63.1 &	83.3 &	51.4 &	59.8 &	52.1 &	- &	- &	- &	- &	- &	-						
			\\
			& SiamAttn \cite{yu2020deformable} & \color{red}71.2 &	\color{blue}92.6 &	65.0 &	84.5 &	56.0 &	64.8 &	- &	75.2 &	81.7 &	- &	- &	- &	-								
			\\
			& DiMP-50 \cite{bhat2019learning} & 68.0 &	88.8 &	64.8 &	85.8 &	56.8 &	64.8 &	56.4 &	74.0 &	80.1 &	68.7 &	61.1 &	71.7 &	49.2						
			\\			
			& SiamRPN++ \cite{li2019siamrpn++} & 69.6 &	91.5 &	64.2 &	84.0 &	49.6 &	56.9 &	49.1 &	73.3 &	80.0 &	69.4 & - &	- &	-			
			\\
			& SiamDW \cite{zhang2019deeper} & 67.0 &	89.2 &	53.6 &	77.6 &	38.5 &	48.0 &	38.9 &	61.1 &	71.3 &	56.3 &	42.9 &	48.3 &	14.7
			\\			
			\bottomrule
		\end{tabular}
	\end{center}
	
\end{table*}

We have selected 37 single-object trackers for this experimental evaluation study. They were published in the last four years in reputed conferences and indexed journals. Although we have included the ARTrack\cite{wei2023autoregressive} in the literature review, it is not included in the experimental evaluation because the raw results or tracking model of this tracker are still not publicly available.  In this evaluation, these selected trackers are categorized as CNN-based trackers, CNN-Transformer based trackers,  and fully-Transformer based trackers and then their category-wise performances are discussed. The overall performances of these trackers are summarized in the Table~\ref{re:3} and their attribute-wise results are detailed in the Table~\ref{re:attribute}. Tracking performances of these approaches for each benchmark datasets are discussed in the following subsections.

\begin{table*}
	\begin{center}
		\footnotesize
		\caption{Tracking attribute-based comparison on OTB100, UAV123, and LaSOT datasets. In each tracking attribute, percentage of success rate scores (SR) of best three trackers and their types are listed.  The symbols \dag, \ddag, and $\Diamond$ are denoting One-stream One-stage trackers, Two-stream Two-stage trackers, and CNN-Transformer based trackers, respectively.    }\label{re:attribute}%
		\begin{tabular}{cl@{\extracolsep{0.40cm}}l@{\extracolsep{1.0cm}}l@{\extracolsep{0.30cm}}l@{\extracolsep{1.0cm}}l@{\extracolsep{0.30cm}}l@{\extracolsep{0.30cm}}}
			\midrule
			\multirow{2}{*}{\textbf{Tracking Attribute}} &  \multicolumn{2}{c}{\textbf{OTB100}} & \multicolumn{2}{c}{\textbf{UAV123}} & \multicolumn{2}{c}{\textbf{LaSOT}} \\ 
			\cmidrule{2-3} \cmidrule{4-5} \cmidrule{6-7} 
			& SR  & Tracker & SR & Tracker & SR & Tracker  \\
			\midrule
			\multirow{3}{*}{ ARC} & - & - & 70.6 &  CSWinTT \cite{bib67} $\Diamond$& 70.9 & SeqTrack \cite{chen2023seqtrack} \dag\\
			& - & - & 70.2 & SimTrack-B/16 \cite{bib79} \dag&70.1 & DropTrack \cite{wu2023dropmae}\dag\\
			& - & - & 70.1 & OSTrack \cite{bib78} \dag& 69.8 & ProContEXT \cite{lan2022procontext}\dag\\
			\midrule
			\multirow{3}{*}{ BC } & 71.5 & TrTr \cite{zhao2021trtr}$\Diamond$ & 57.6 & OSTrack \cite{bib78}\dag & 65.9 & SeqTrack \cite{chen2023seqtrack}\dag\\
			& 70.4 & SiamAttn \cite{yu2020deformable} & 56.1 &  DropTrack \cite{wu2023dropmae}\dag& 64.4 & ProContEXT \cite{lan2022procontext}\dag\\
			& 70.3 & TrDiMP \cite{bib66} $\Diamond$&  55.4 &  AiATrack \cite{bib73} $\Diamond$& 63.7 & DropTrack \cite{wu2023dropmae}\dag\\
			\midrule
			\multirow{3}{*}{ CM } & - & - & 72.7 & CSWinTT \cite{bib67} $\Diamond$& 75.8 & ProContEXT \cite{lan2022procontext}\dag \\
			& - & - & 72.4 & MixFormer \cite{bib80} \dag& 74.6 &  SeqTrack \cite{chen2023seqtrack}\dag\\
			& - & -  & 72.4 &  DropTrack \cite{wu2023dropmae}\dag & 74.0 &  AiATrack \cite{bib73} $\Diamond$\\
			\midrule
			\multirow{3}{*}{ DEF } & 69.1 & SiamGAT \cite{guo2021graph} & - & - & 74.9 & SeqTrack \cite{chen2023seqtrack}\dag\\
			& 69.0 & TrTr \cite{zhao2021trtr} $\Diamond$& - & - & 72.9 & ProContEXT \cite{lan2022procontext} \dag\\
			& 68.6 & SiamTPN \cite{xing2022siamese} $\Diamond$ & - & - &  72.3 & GRM \cite{gao2023generalized} \dag\\
			\midrule
			\multirow{3}{*}{ FM } & 72.9 & SparseTT \cite{bib77} \ddag& 68.3 & AiATrack \cite{bib73} $\Diamond$& 60.1 & DropTrack \cite{wu2023dropmae}\dag\\
			& 72.6 & SwinTrack \cite{bib38} \ddag& 67.9 & CSWinTT \cite{bib67} $\Diamond$ & 59.1 & OSTrack \cite{bib78}\dag\\
			& 72.5 &  MixFormer \cite{bib80} \dag & 67.6 & DropTrack \cite{wu2023dropmae}\dag &  58.6 & ProContEXT \cite{lan2022procontext}\dag\\
			\midrule
			\multirow{3}{*}{ FOC } & - & - & 56.5 & SimTrack-B/16 \cite{bib79}\dag & 64.5 &  DropTrack \cite{wu2023dropmae} \dag\\
			& - & - & 56.1 & OSTrack \cite{bib78}\dag & 63.1 & ProContEXT \cite{lan2022procontext} \dag\\
			& - & -  & 54.4 & DropTrack \cite{wu2023dropmae} \dag& 63.0 & SeqTrack \cite{chen2023seqtrack}\dag\\
			\midrule
			\multirow{3}{*}{ IV } & 74.2 & TrTr \cite{zhao2021trtr} $\Diamond$& 67.0 & AiATrack \cite{bib73}  $\Diamond$&  71.7 & ProContEXT \cite{lan2022procontext} \dag\\
			& 73.8 & SiamAttn \cite{yu2020deformable} & 65.9 & CSWinTT \cite{bib67} $\Diamond$& 71.4 & OSTrack \cite{bib78}\dag\\
			& 72.8 & KeepTrack \cite{mayer2021learning} & 65.5 & OSTrack \cite{bib78} \dag& 71.3 & DropTrack \cite{wu2023dropmae}\dag\\
			\midrule
			\multirow{3}{*}{ IPR } & 72.1  & TrDiMP \cite{bib66}$\Diamond$ & - & - & - & - \\
			& 72.0 & SiamAttn \cite{yu2020deformable} & - & - & - & -\\
			& 71.7 & SiamBAN \cite{chen2020siamese}  & - &  - & - & -\\
			\midrule
			\multirow{3}{*}{ LR } & 74.5  & SiamAttn \cite{yu2020deformable} & 58.6 &  DropTrack \cite{wu2023dropmae} \dag & 65.6 &  DropTrack \cite{wu2023dropmae} \dag\\
			& 74.2 & SparseTT \cite{bib77} \ddag&  58.4 & SwinTrack \cite{bib38} \ddag & 65.2 & SeqTrack \cite{chen2023seqtrack}\dag\\
			& 73.3 & SiamGAT \cite{guo2021graph}  & 57.4 & OSTrack \cite{bib78}\dag & 65.0 & OSTrack \cite{bib78} \dag\\
			\midrule
			\multirow{3}{*}{ MB } & 75.5 & MixFormer \cite{bib80}\dag & - & - & 71.0 & ProContEXT \cite{lan2022procontext} \dag\\
			& 75.1 & SparseTT \cite{bib77} \ddag& - & - & 70.5 & SeqTrack \cite{chen2023seqtrack}\dag\\
			& 74.5 & AiATrack \cite{bib73} $\Diamond$ & - & - & 69.7 & DropTrack \cite{wu2023dropmae} \dag\\
			\midrule
			\multirow{3}{*}{ OCC/POC } & 68.5  & TrTr \cite{zhao2021trtr} $\Diamond$&  66.0 & OSTrack \cite{bib78}\dag & 69.7 & SeqTrack \cite{chen2023seqtrack}\dag\\
			& 68.3 & SiamAttn \cite{yu2020deformable} & 65.8 & CSWinTT \cite{bib67} $\Diamond$& 69.5 & DropTrack \cite{wu2023dropmae} \dag\\
			& 68.2 & KeepTrack \cite{mayer2021learning} & 65.8 & DropTrack \cite{wu2023dropmae} \dag & 68.9 & ProContEXT \cite{lan2022procontext} \dag\\
			\midrule
			\multirow{3}{*}{ OPR } & 70.9  & SiamAttn \cite{yu2020deformable} & - & - & - & -\\
			& 70.7 & SiamGAT \cite{guo2021graph} & - & - & - & -\\
			& 70.4 & TrTr \cite{zhao2021trtr} $\Diamond$ & - & - & - & - \\
			\midrule
			\multirow{3}{*}{ OV } & 70.3 & MixFormer \cite{bib80}\dag & 68.7 & ProContEXT \cite{lan2022procontext} \dag & 67.4 &   DropTrack \cite{wu2023dropmae} \dag\\
			& 69.8 & ProContEXT \cite{lan2022procontext} \dag & 68.5 & DropTrack \cite{wu2023dropmae} \dag & 67.2 &  SeqTrack \cite{chen2023seqtrack}\dag\\
			& 69.6 & SeqTrack \cite{chen2023seqtrack}\dag & 68.0 & SimTrack-B/16 \cite{bib79} \dag & 66.4 &  OSTrack \cite{bib78}\dag\\
			
			\midrule  
			\multirow{3}{*}{ ROT } 
			& - & - & - & - & 72.4 & SeqTrack \cite{chen2023seqtrack} \dag\\
			& - & - & - & - & 71.1 & ProContEXT \cite{lan2022procontext} \dag \\
			& - & -  & - & -  &  70.8 & DropTrack \cite{wu2023dropmae} \dag\\
			\midrule
			\multirow{3}{*}{ SV } & 73.4 & ProContEXT \cite{lan2022procontext}\dag & 70.0 & DropTrack \cite{wu2023dropmae} \dag & 72.4 & SeqTrack \cite{chen2023seqtrack}\dag\\
			& 72.8 & KeepTrack \cite{mayer2021learning} & 69.7 & OSTrack \cite{bib78}\dag & 71.3 & DropTrack \cite{wu2023dropmae}\dag \\
			& 72.8 & TransT \cite{bib37} $\Diamond$ & 69.4 & CSWinTT \cite{bib67} $\Diamond$ & 71.2 & ProContEXT \cite{lan2022procontext} \dag\\
			\midrule
			\multirow{3}{*}{ SOB } & - & - & 69.7 & SimTrack-B/16 \cite{bib79} \dag& - & - \\
			& - & - & 69.1 & OSTrack \cite{bib78} \dag& - & - \\
			& - & -  & 68.6 & CSWinTT \cite{bib67} $\Diamond$ & - & -\\
			\midrule
			\multirow{3}{*}{ VC } & - & - & 74.0 & SimTrack-B/16 \cite{bib79} \dag& 74.9 & SeqTrack \cite{chen2023seqtrack} \dag \\
			& - & - & 73.9 & DropTrack \cite{wu2023dropmae} \dag& 73.4 & ProContEXT \cite{lan2022procontext} \dag\\
			& - & -  &  73.4 & OSTrack \cite{bib78} \dag &72.6 & DropTrack \cite{wu2023dropmae} \dag\\
			
			\bottomrule
		\end{tabular}
	\end{center}	
\end{table*}

\subsubsection{Analysis on OTB100 Dataset}
\begin{figure*}
	\centering
	\begin{minipage}[b]{0.245\linewidth}
		\centering
		\centerline{\includegraphics[width=\textwidth]{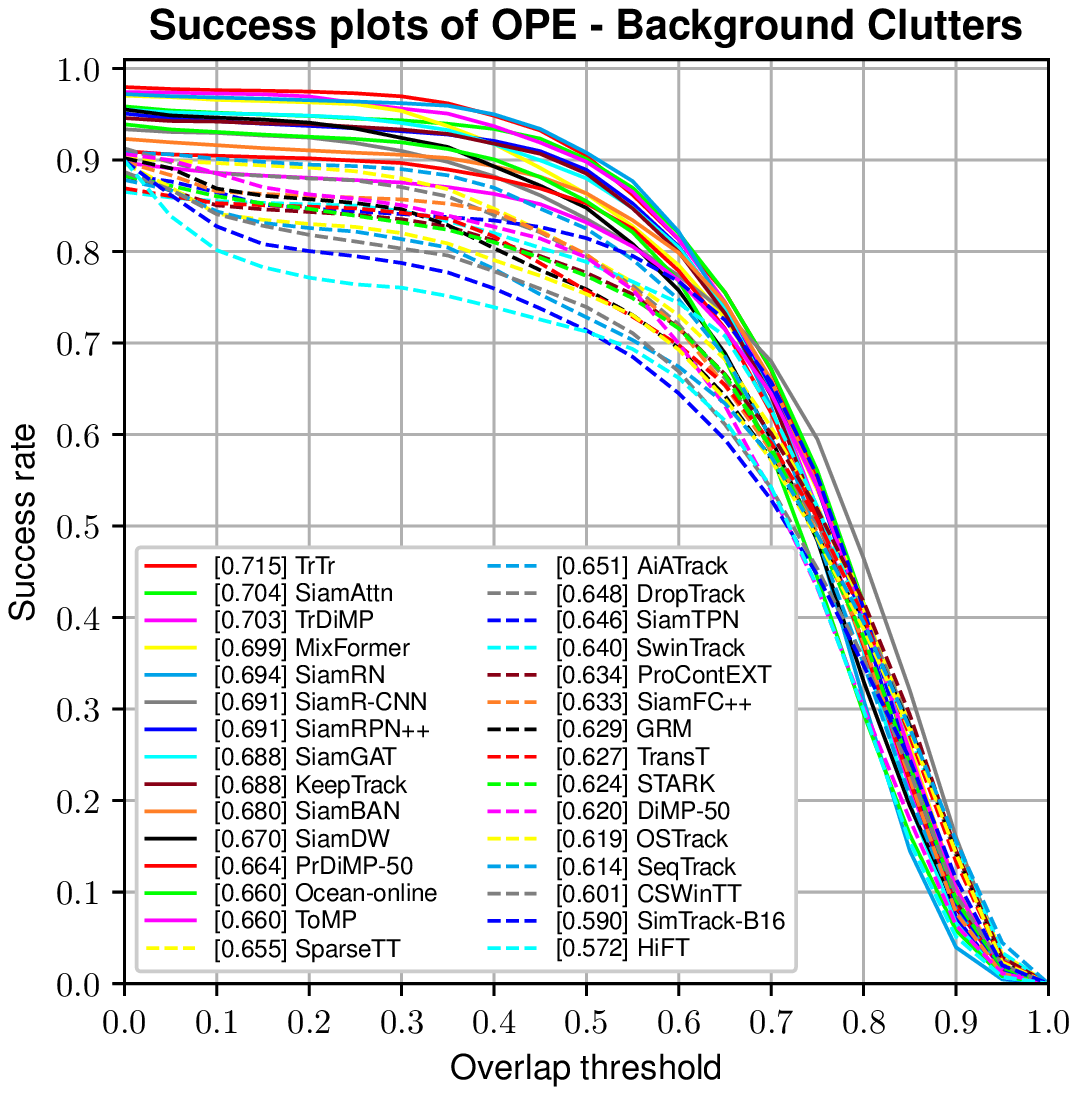}}
	\end{minipage}
	\begin{minipage}[b]{0.245\linewidth}
		\centering
		\centerline{\includegraphics[width=\textwidth]{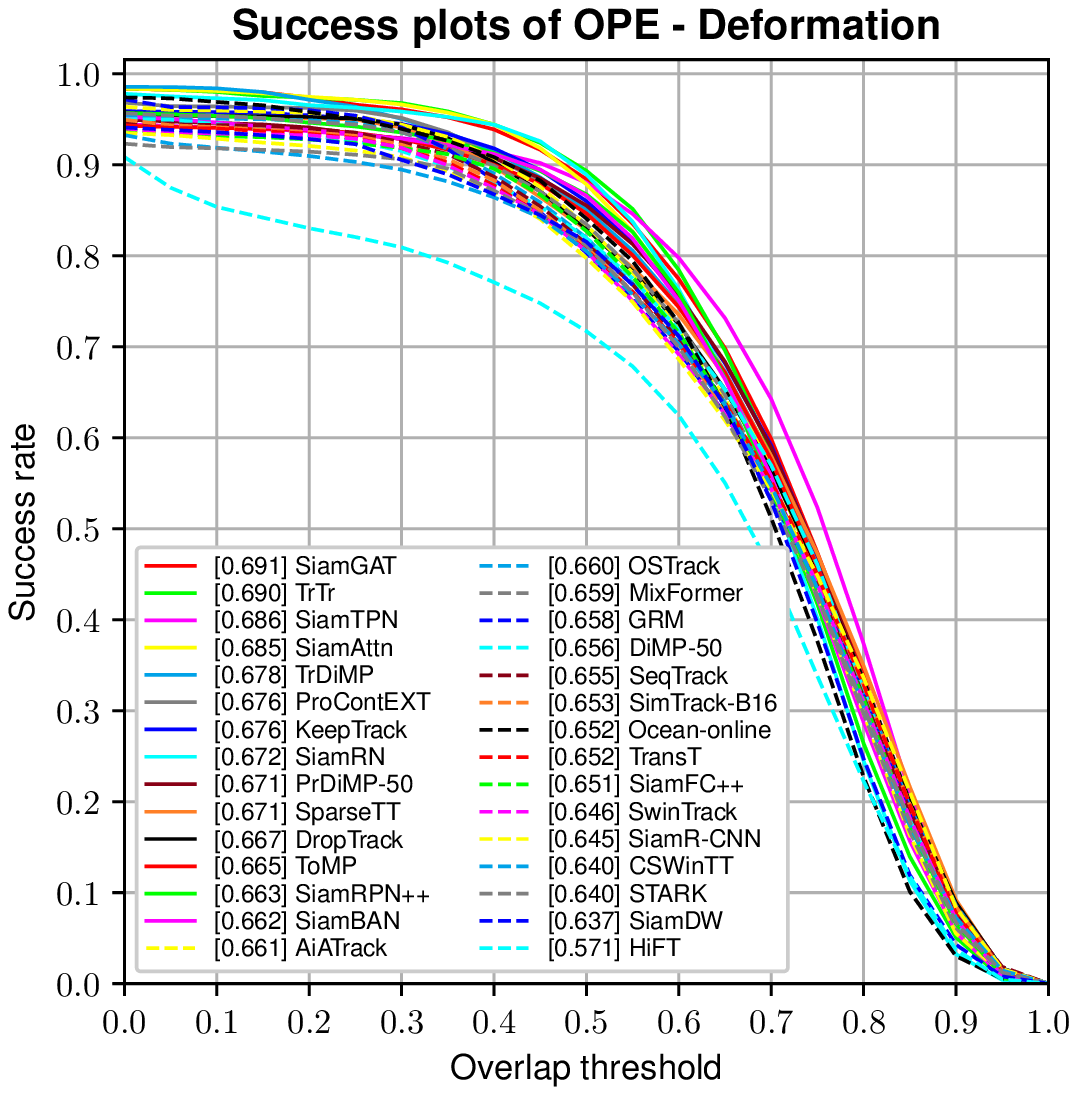}}
	\end{minipage}
	\begin{minipage}[b]{0.245\linewidth}
		\centering
		\centerline{\includegraphics[width=\textwidth]{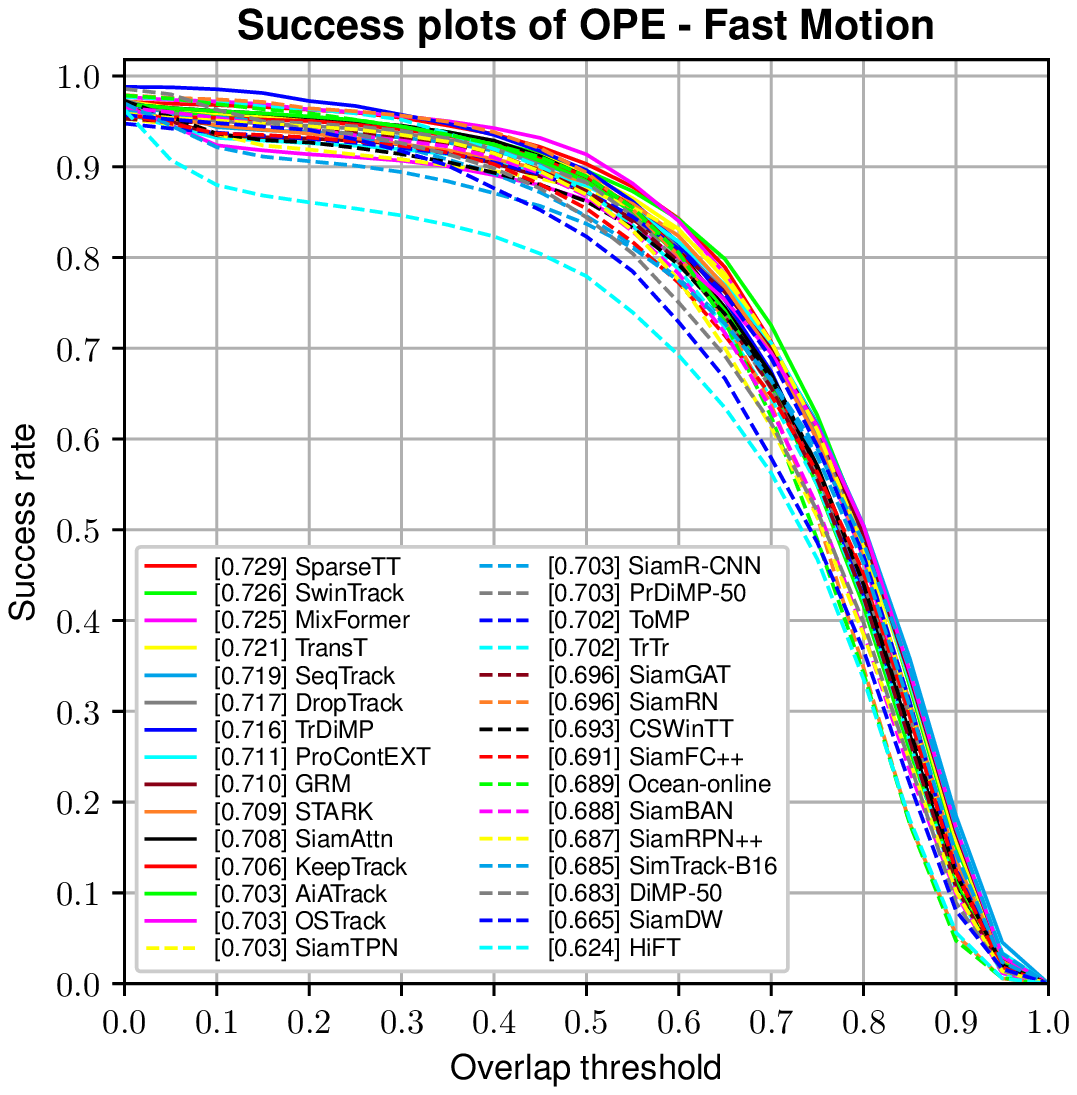}}
	\end{minipage}
	\begin{minipage}[b]{0.245\linewidth}
		\centering
		\centerline{\includegraphics[width=\textwidth]{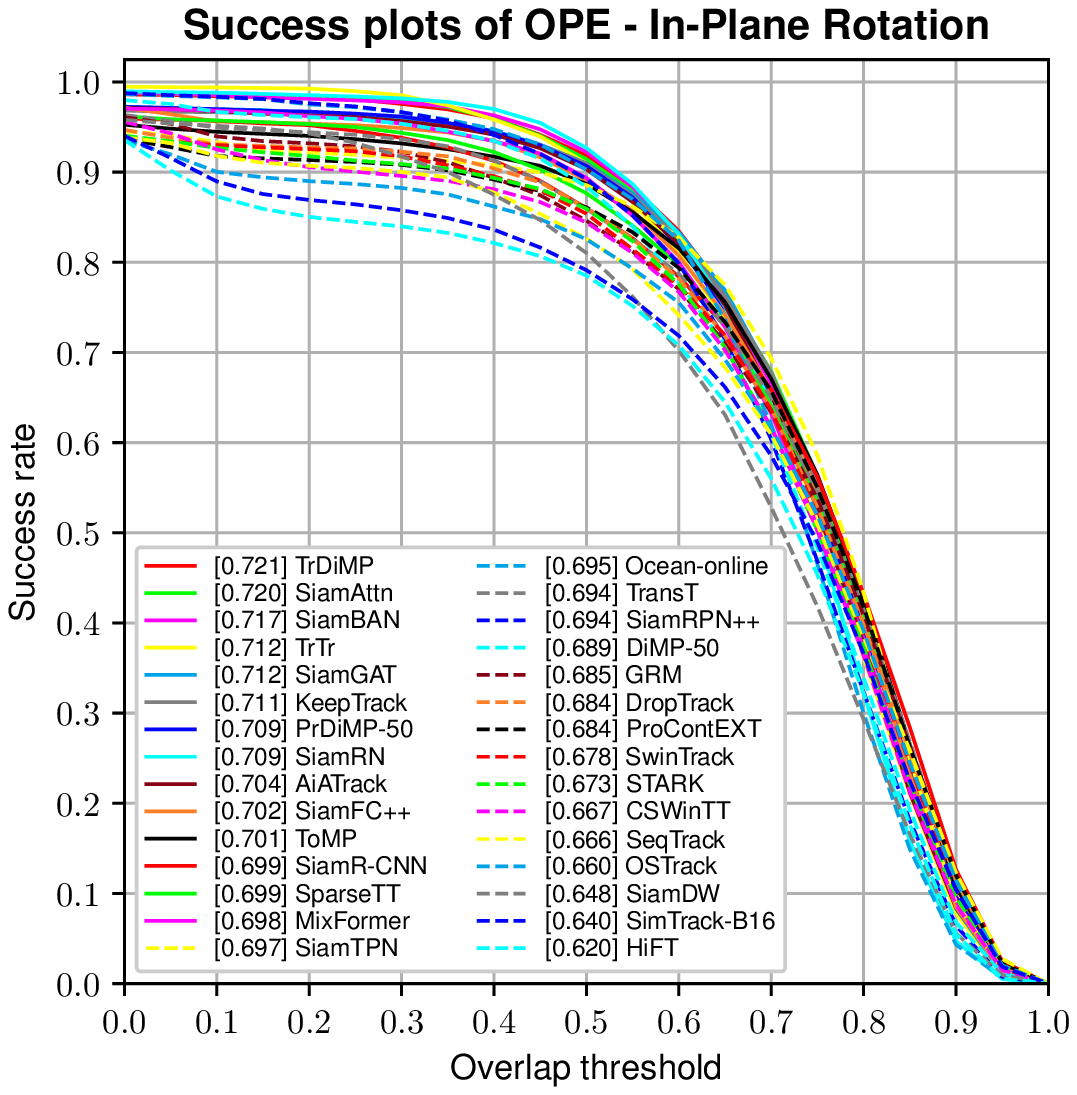}}
	\end{minipage}
	\begin{minipage}[b]{0.245\linewidth}
		\centering
		\centerline{\includegraphics[width=\textwidth]{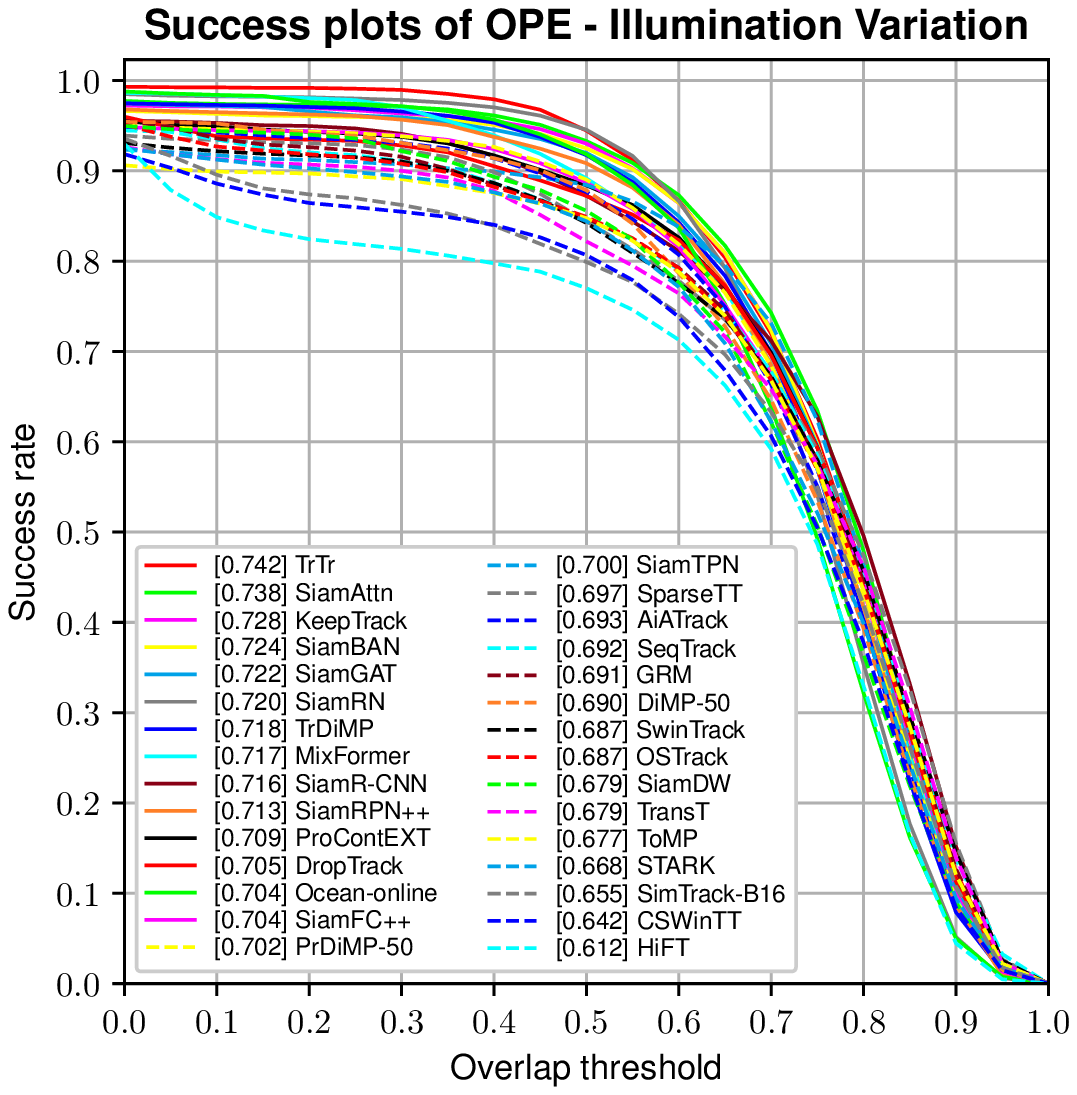}}
	\end{minipage}
	\begin{minipage}[b]{0.245\linewidth}
		\centering
		\centerline{\includegraphics[width=\textwidth]{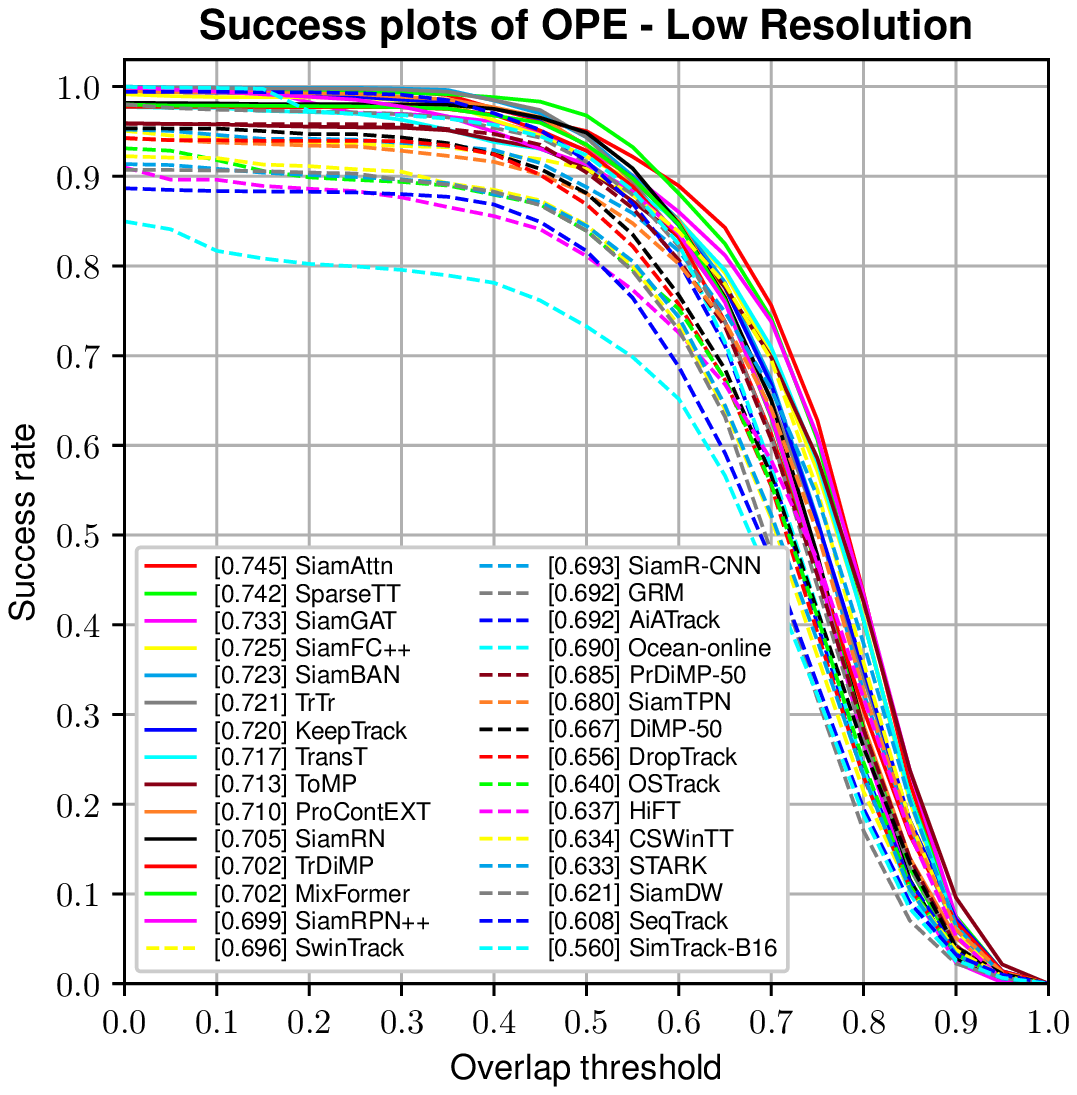}}
	\end{minipage}
	\begin{minipage}[b]{0.245\linewidth}
		\centering
		\centerline{\includegraphics[width=\textwidth]{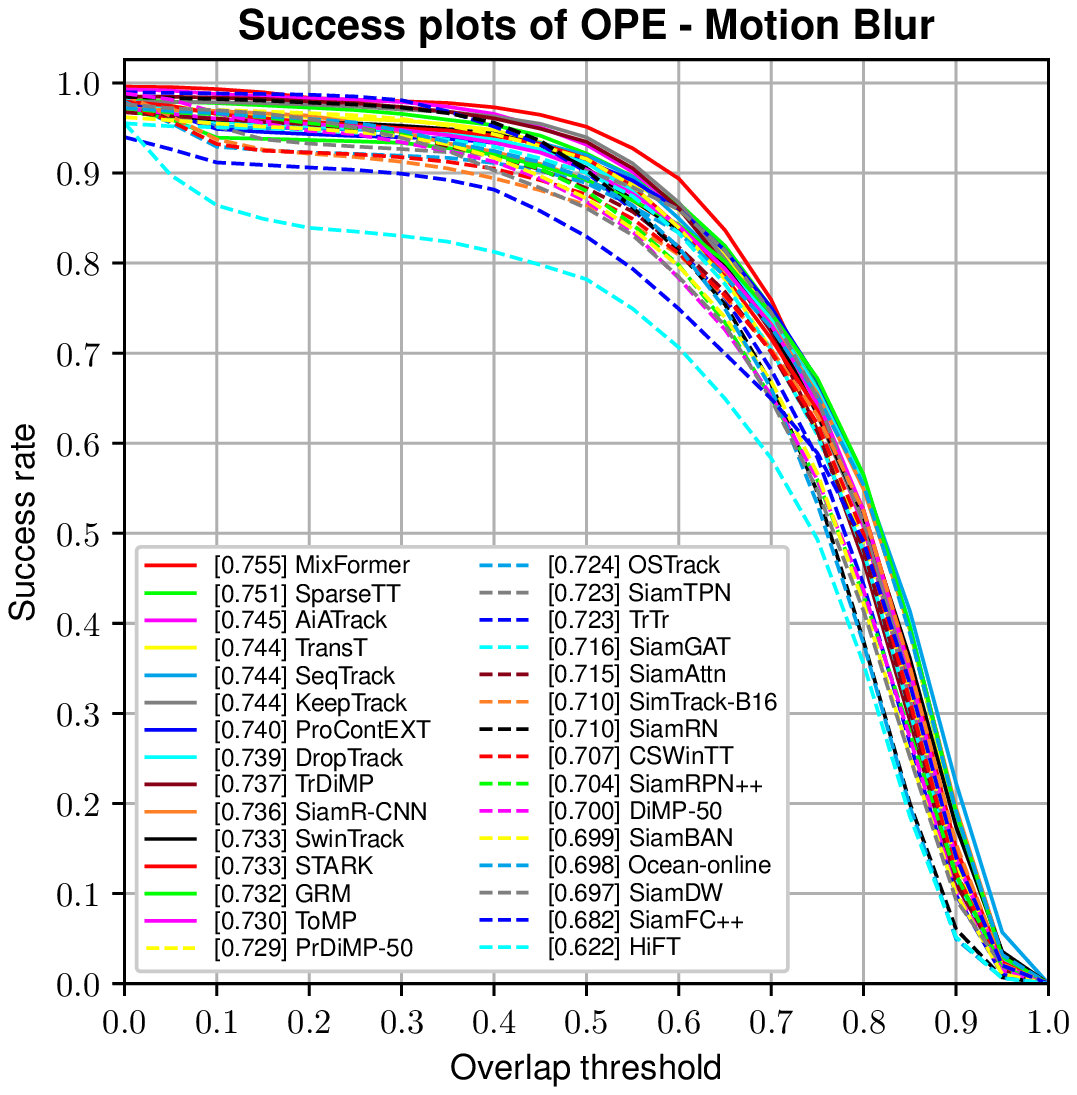}}
	\end{minipage}
	\begin{minipage}[b]{0.245\linewidth}
		\centering
		\centerline{\includegraphics[width=\textwidth]{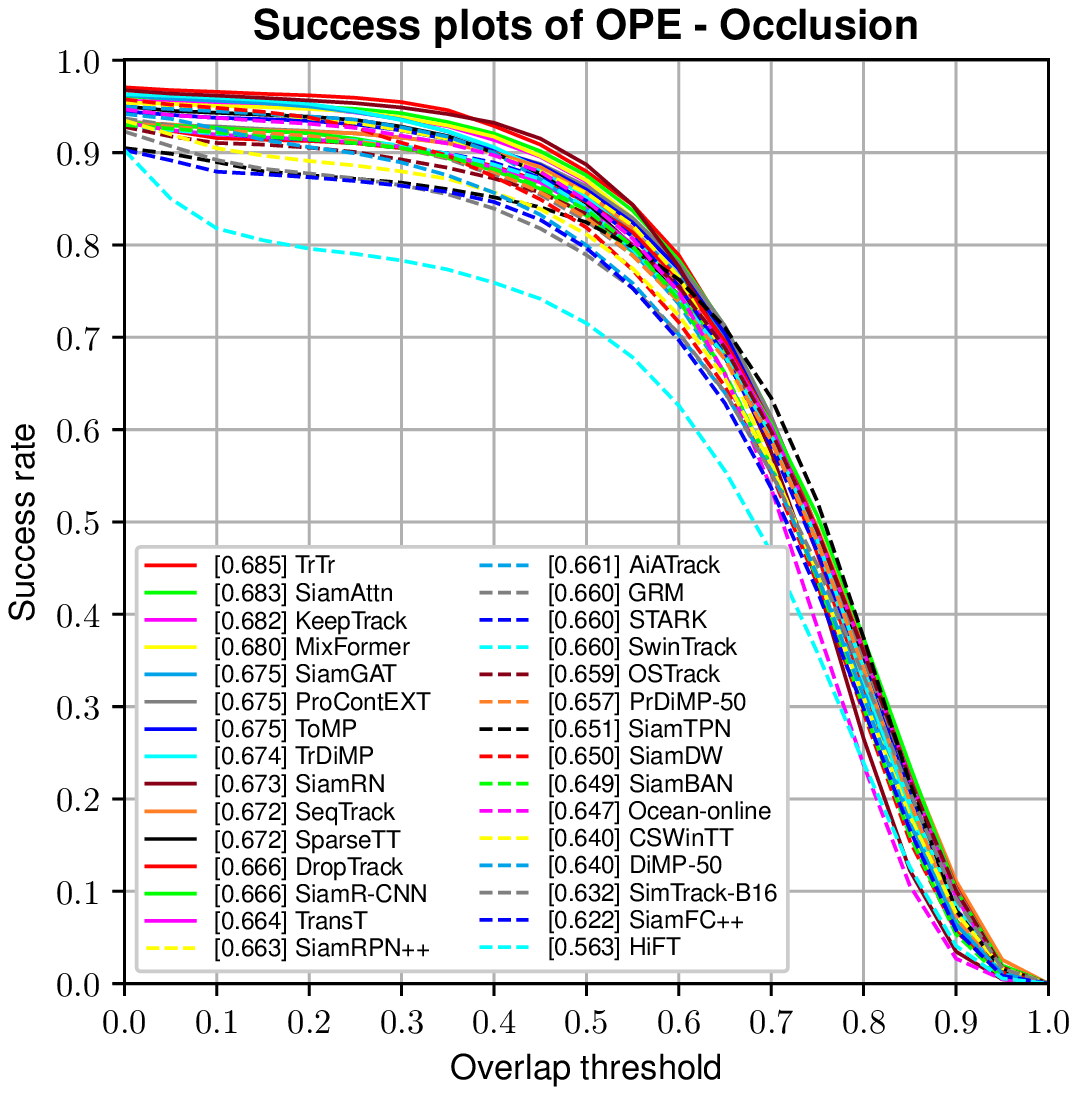}}
	\end{minipage}
	\begin{minipage}[b]{0.245\linewidth}
		\centering
		\centerline{\includegraphics[width=\textwidth]{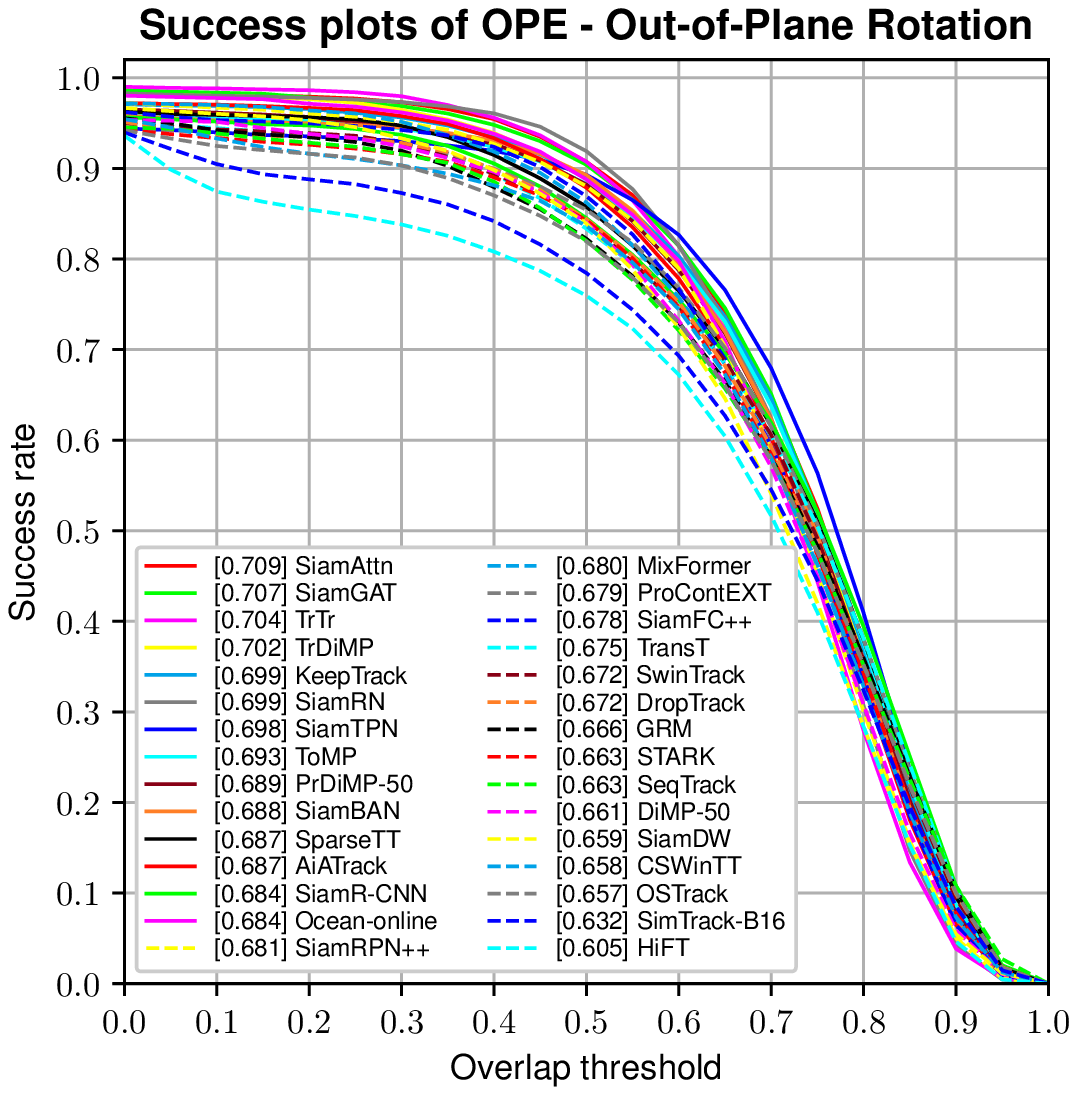}}
	\end{minipage}
	\begin{minipage}[b]{0.245\linewidth}
		\centering
		\centerline{\includegraphics[width=\textwidth]{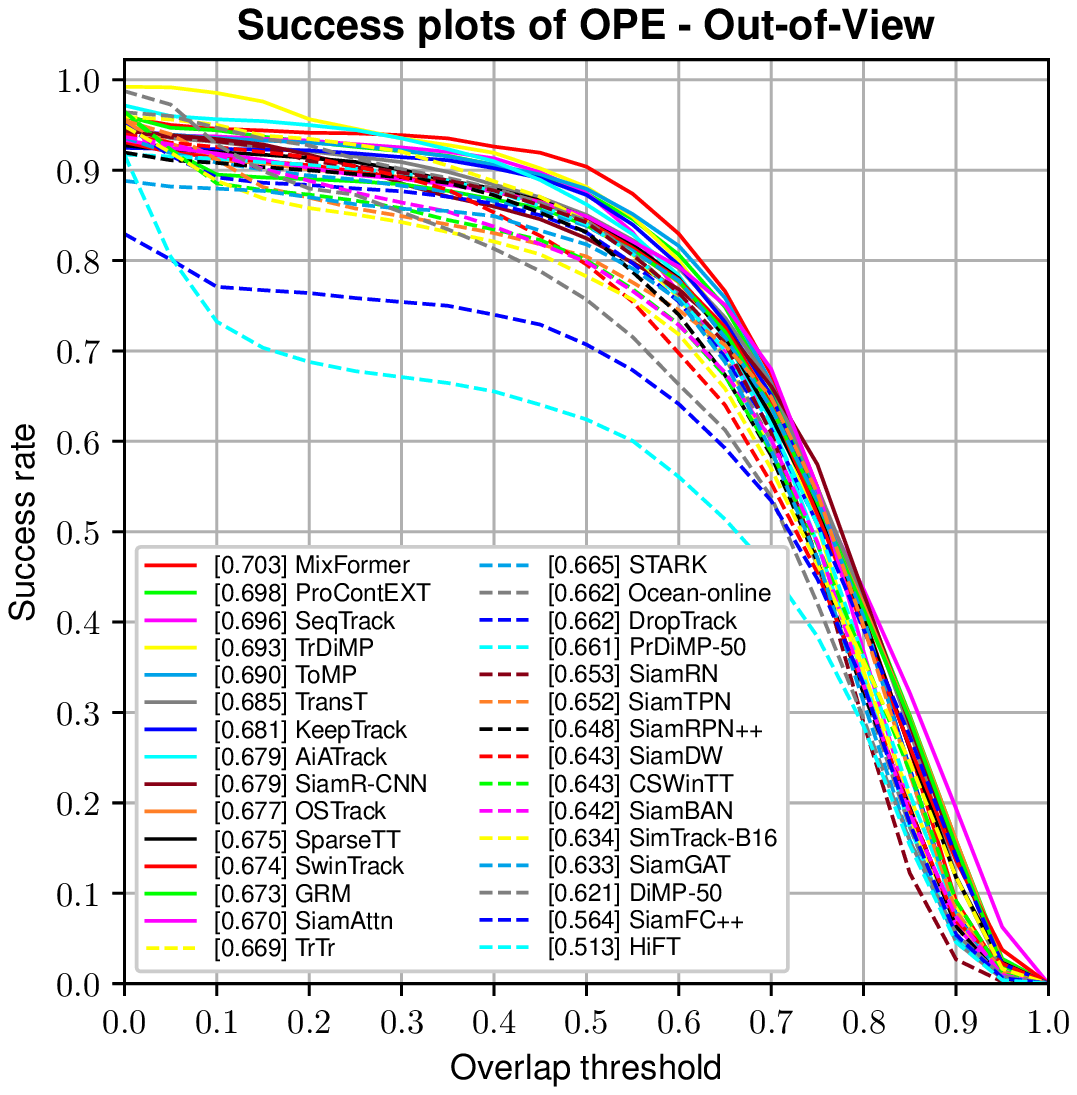}}
	\end{minipage}
	\begin{minipage}[b]{0.245\linewidth}
		\centering
		\centerline{\includegraphics[width=\textwidth]{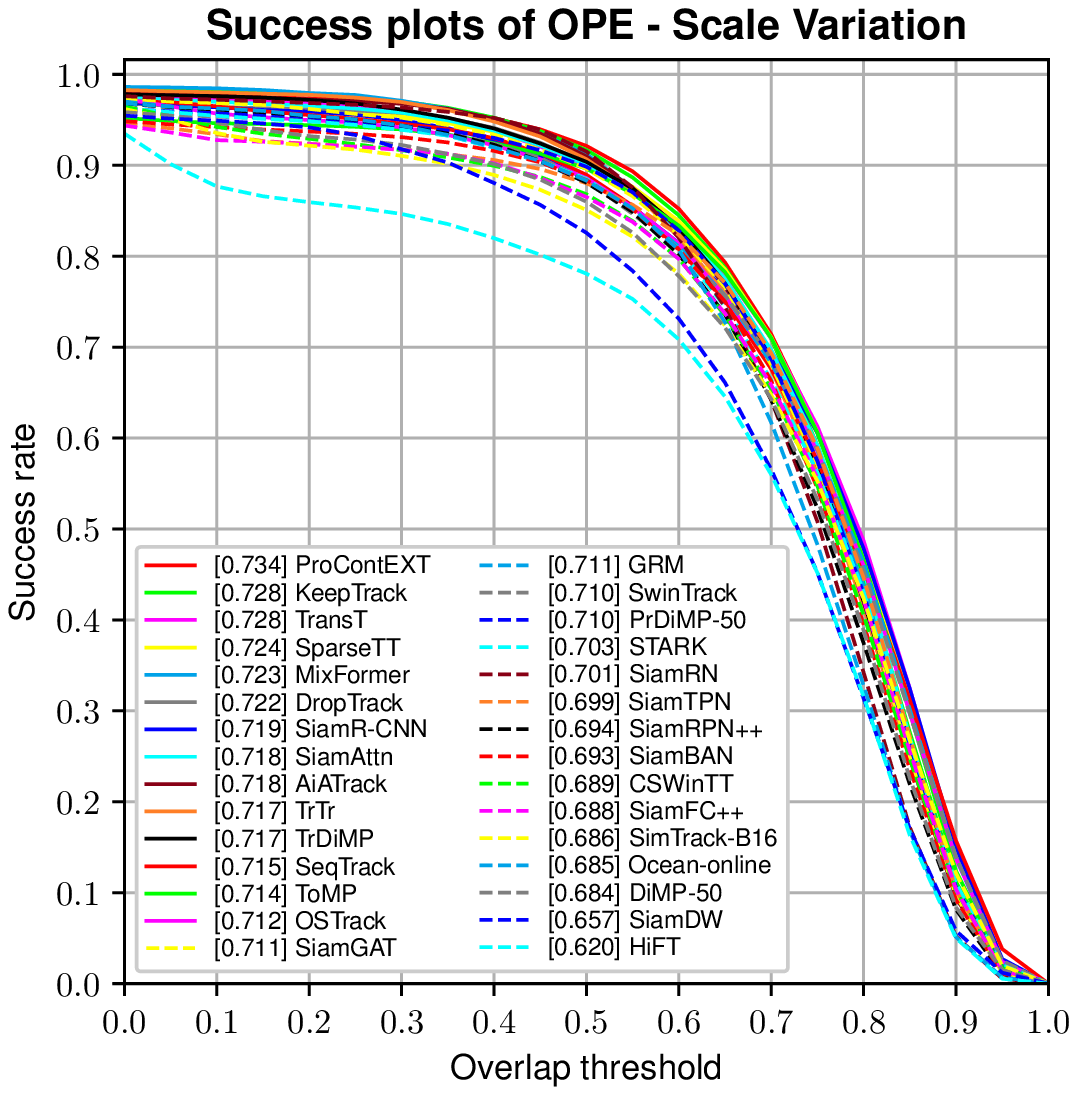}}
	\end{minipage}
	\caption[The success plots of the trackers  for 11 challenging attributes on OTB100.]{The success plots of trackers  for 11 challenging attributes on OTB100.}
	\label{fig2:Attribute_analysis_OTB}
\end{figure*}

We used the official toolkit of OTB100 to evaluate the tracking performance of the trackers. Success plots were ranked based on the area-under-curve (AUC) scores, while the precision (P) scores at a threshold of 20 pixels were used to evaluate the trackers. In addition to the overall evaluation, the performances of the trackers were measured based on eleven tracking attributes using the toolkit and their success plots, as shown in Fig.~\ref{fig2:Attribute_analysis_OTB}. 

Based on the overall success rates and precision scores in Table~\ref{re:3}, both CNN-Transformer based trackers and CNN-based trackers have achieved overall high performance. TrTr \cite{zhao2021trtr} tracker showed excellent performance on OTB100 in terms of accuracy and precision by replacing the cross-correlation mechanism of Siamese tracking with a Transformer. Similarly, TrDiMP \cite{bib66} showed competitive success and precision scores by exploiting temporal cues with a CNN-Transformer based architecture. On the other side, CNN-based Siamese trackers: SiamAttn \cite{yu2020deformable} and SiamRN \cite{cheng2021learning} showed high success and precision scores, respectively. 

Compared to other benchmark datasets, the tracking performances of fully-Transformer based trackers are slightly lower than the other two types on OTB100. CNN-Transformer based trackers and CNN-based trackers  showed better performances in OTB100 by capturing and matching convolutional features that represent local-regional cues. Since most of the OTB videos have fewer frames, the appearance of the target remains the same in many sequences. Therefore, CNN-based feature extraction and matching have shown excellent tracking results. On the other hand, the performances of fully-Transformer based approaches mainly rely on their attention mechanism and global feature capturing capabilities, and their performances are slightly limited on OTB100 since most tracking sequences have low-resolution videos with fewer frames.

The attribute-based evaluation on OTB100 showed that the dataset is no longer challenging for recent trackers, based on the results in Table~\ref{re:attribute} and the plots in Fig.~\ref{fig2:Attribute_analysis_OTB}. The fully-Transformer based trackers: MixFormer \cite{bib80},  SparseTT \cite{bib77}, and ProContEXT \cite{lan2022procontext} showed better performance than the CNN-based trackers in many challenging attributes. In particular, almost all of the fully Transformer-based trackers successfully handled the fast motion (FM) and out-of-view (OV) scenarios and outperformed the CNN-based trackers by a large margin, due to their long-range feature capturing capability. On the other side, fully-Transformer trackers showed poor performances in background clutters (BC), and deformation (DEF) frames because of their poor discriminative capabilities in short-range videos. In overall, CNN-Transformer based tracker TrTr \cite{zhao2021trtr} successfully handles all tracking challenges in OTB100 by combining the CNN features with a Transformer architecture.

\subsubsection{Analysis on UAV123 Dataset}
\begin{figure*}
	\centering
	\begin{minipage}[b]{0.245\linewidth}
		\centering
		\centerline{\includegraphics[width=\textwidth]{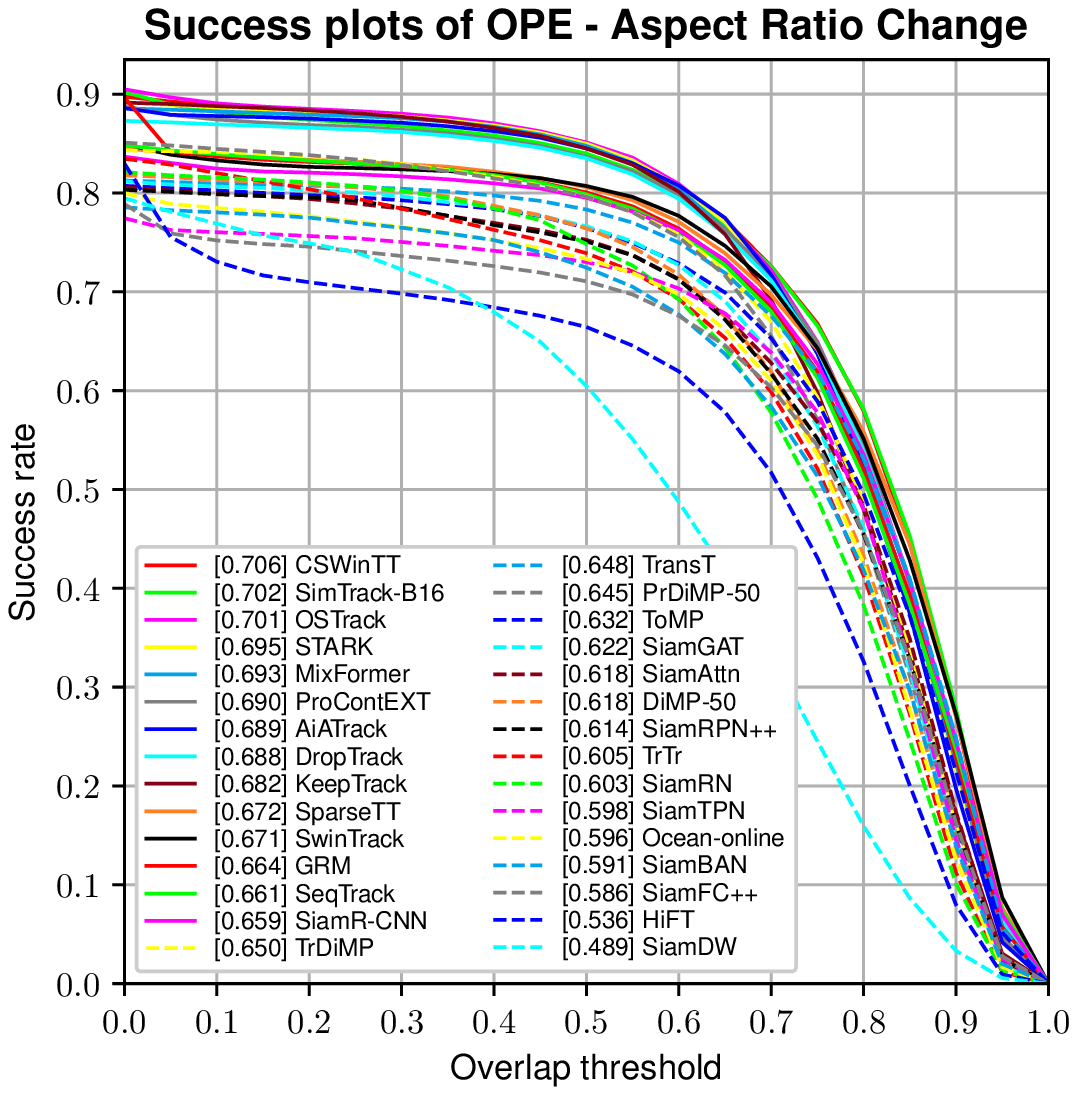}}
	\end{minipage}
	\begin{minipage}[b]{0.245\linewidth}
		\centering
		\centerline{\includegraphics[width=\textwidth]{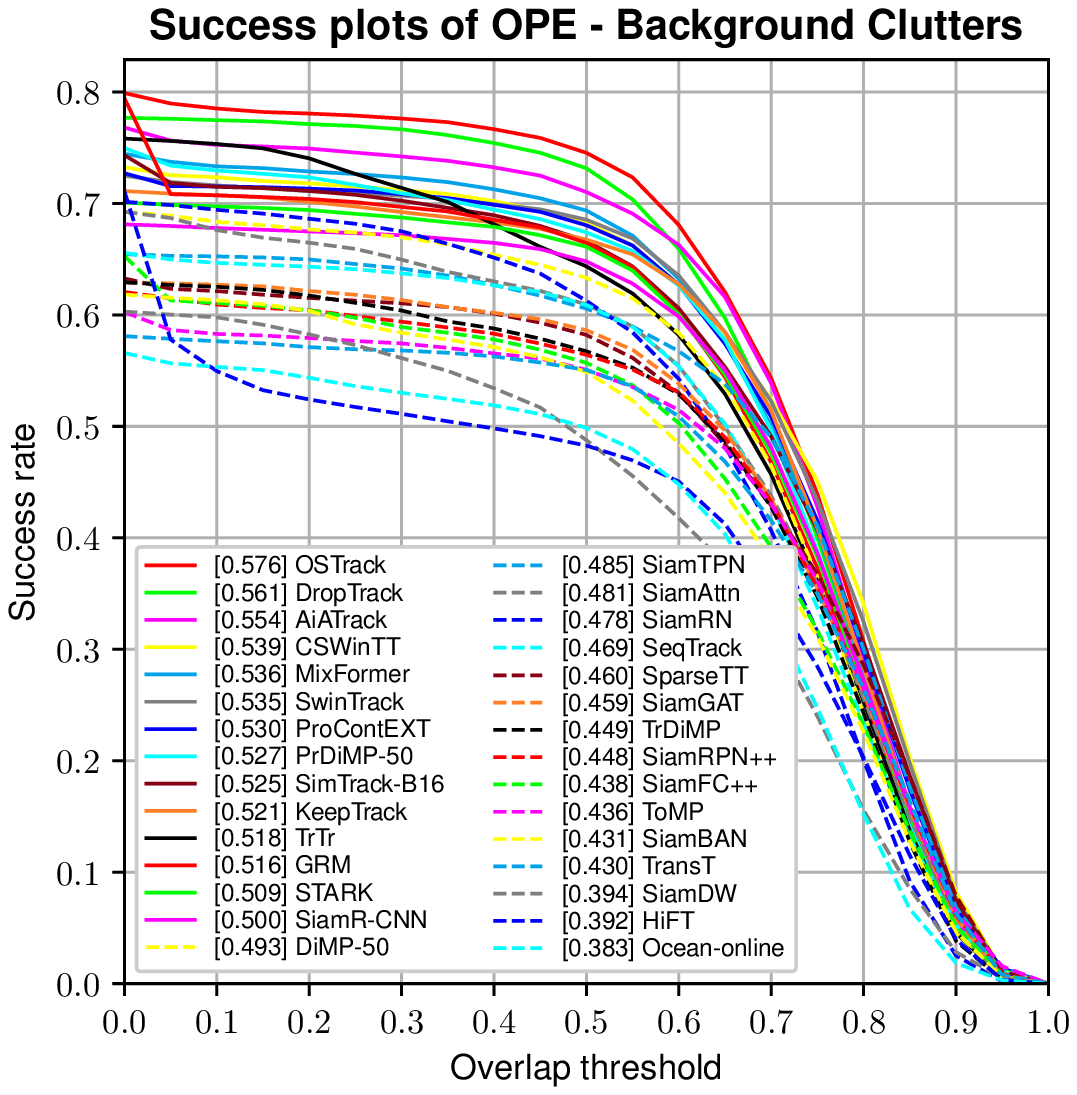}}
	\end{minipage}
	\begin{minipage}[b]{0.245\linewidth}
		\centering
		\centerline{\includegraphics[width=\textwidth]{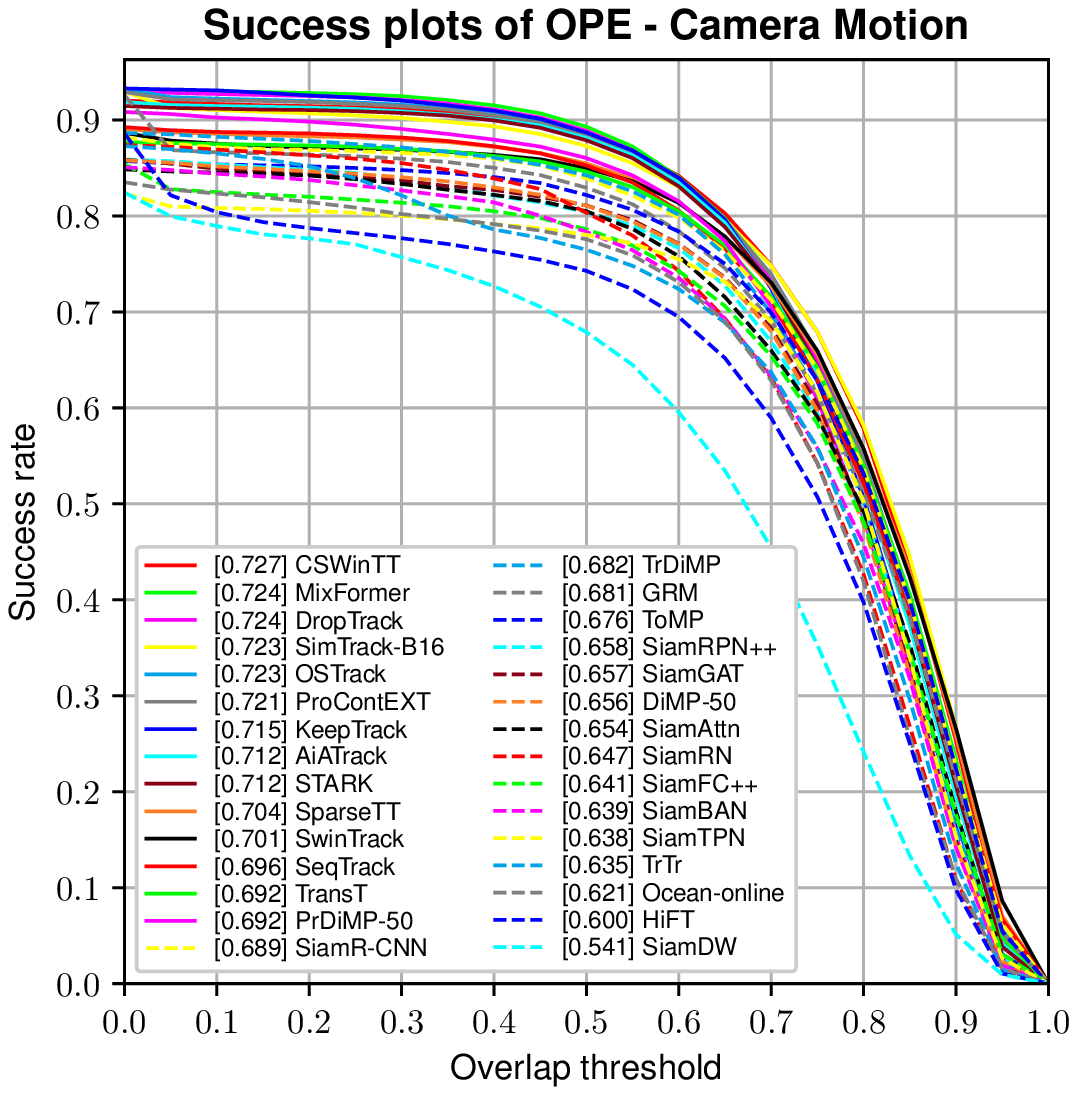}}
	\end{minipage}
	\begin{minipage}[b]{0.245\linewidth}
		\centering
		\centerline{\includegraphics[width=\textwidth]{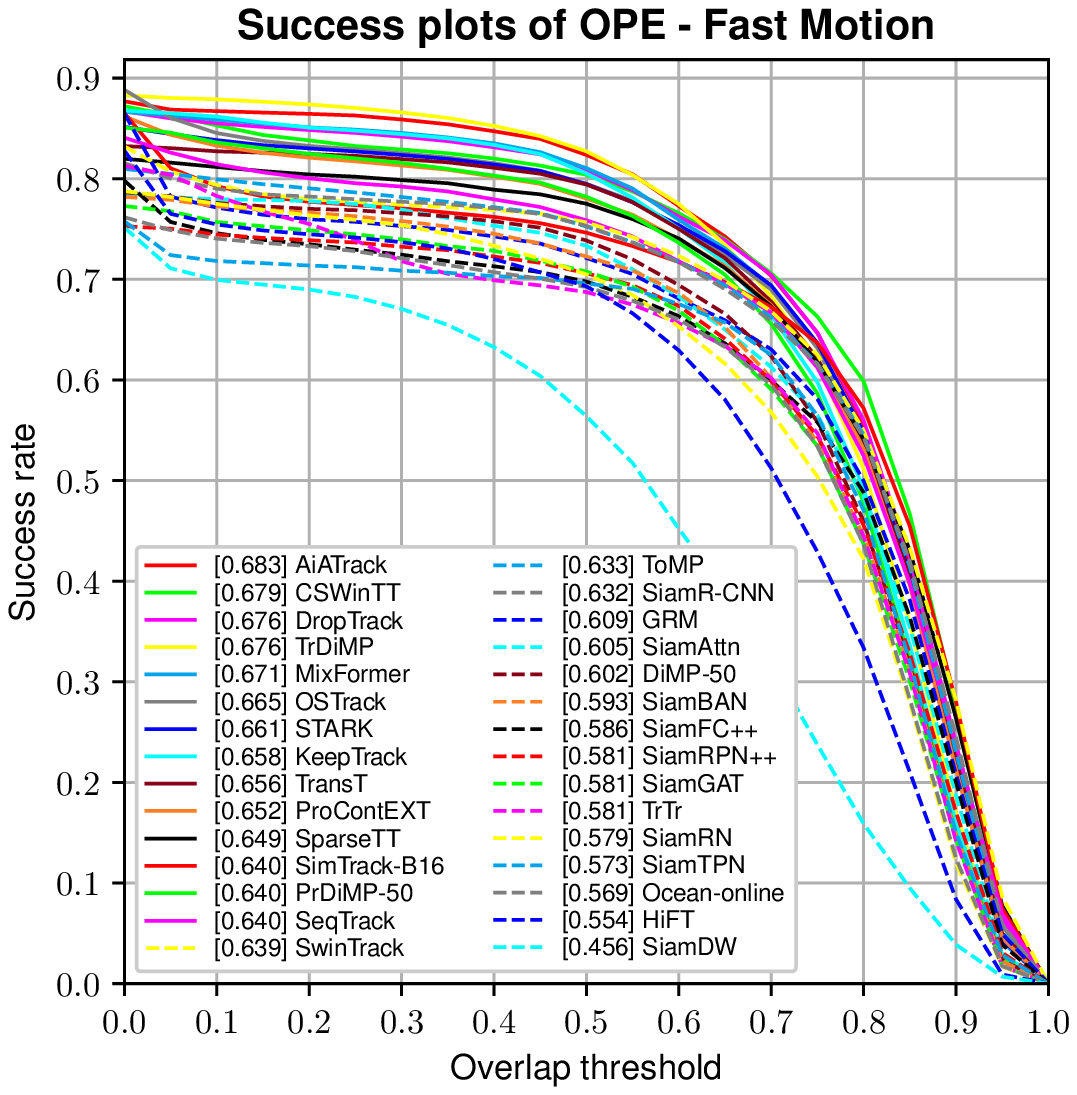}}
	\end{minipage}
	\begin{minipage}[b]{0.245\linewidth}
		\centering
		\centerline{\includegraphics[width=\textwidth]{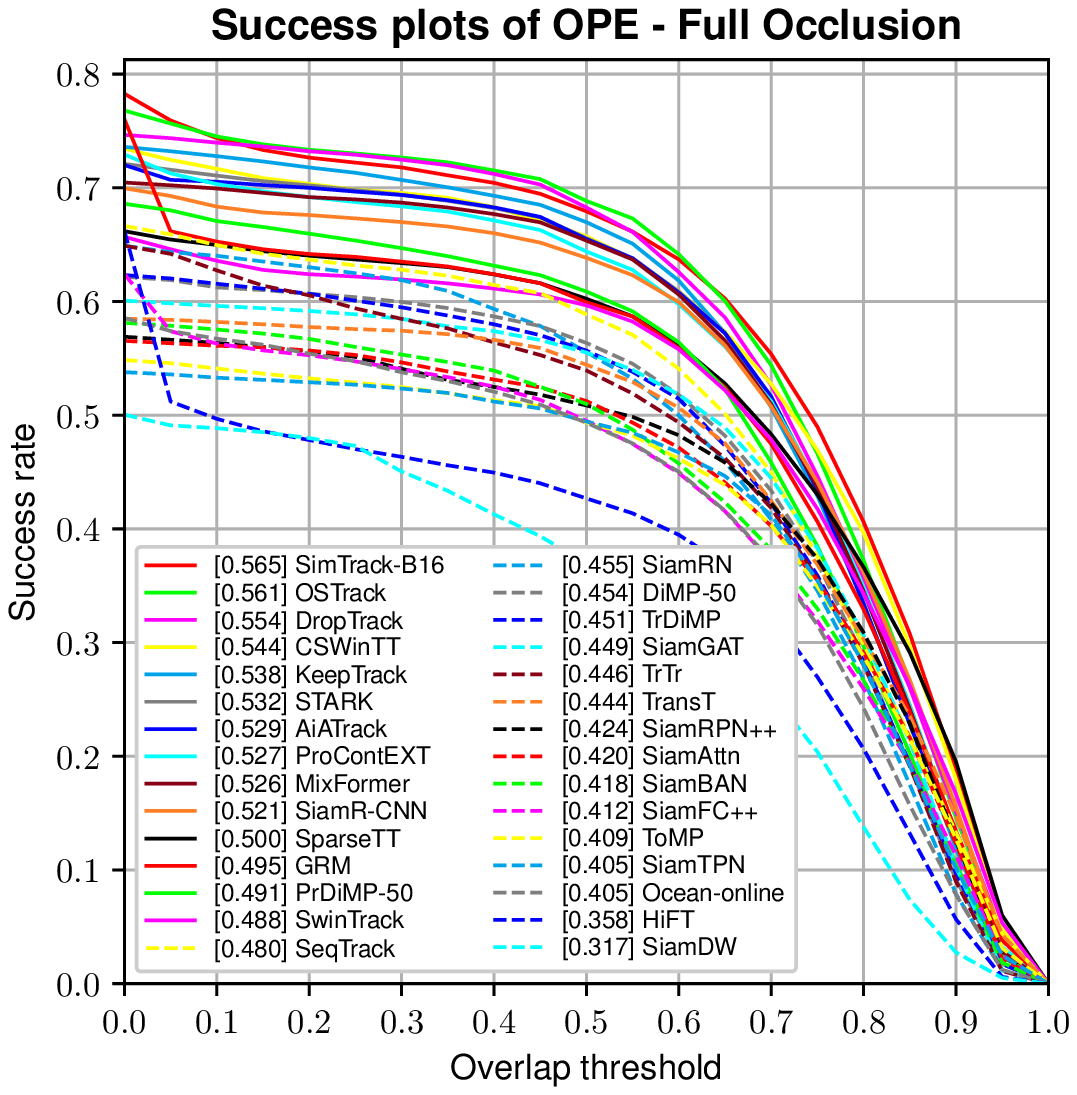}}
	\end{minipage}
	\begin{minipage}[b]{0.245\linewidth}
		\centering
		\centerline{\includegraphics[width=\textwidth]{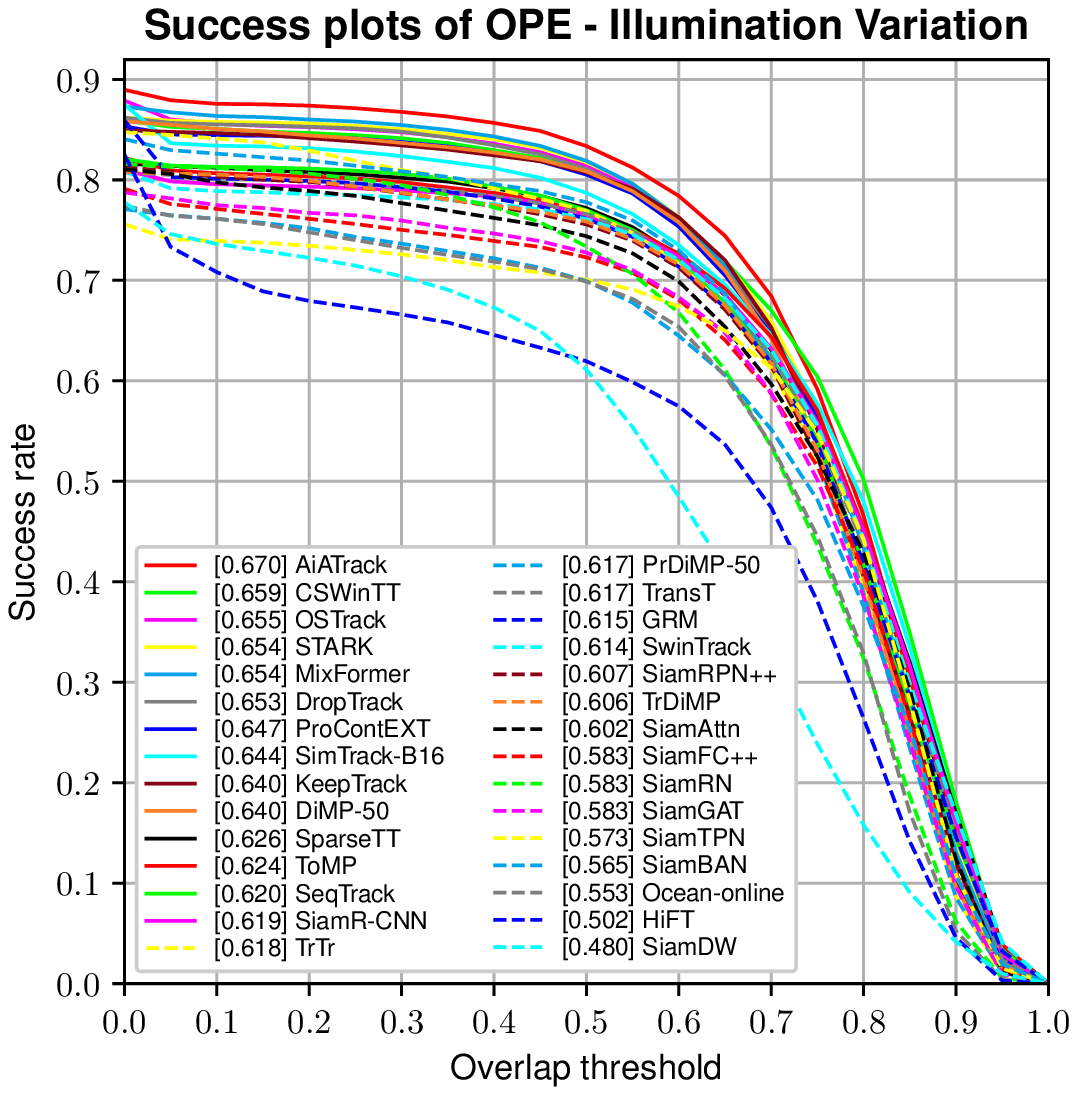}}
	\end{minipage}
	\begin{minipage}[b]{0.245\linewidth}
		\centering
		\centerline{\includegraphics[width=\textwidth]{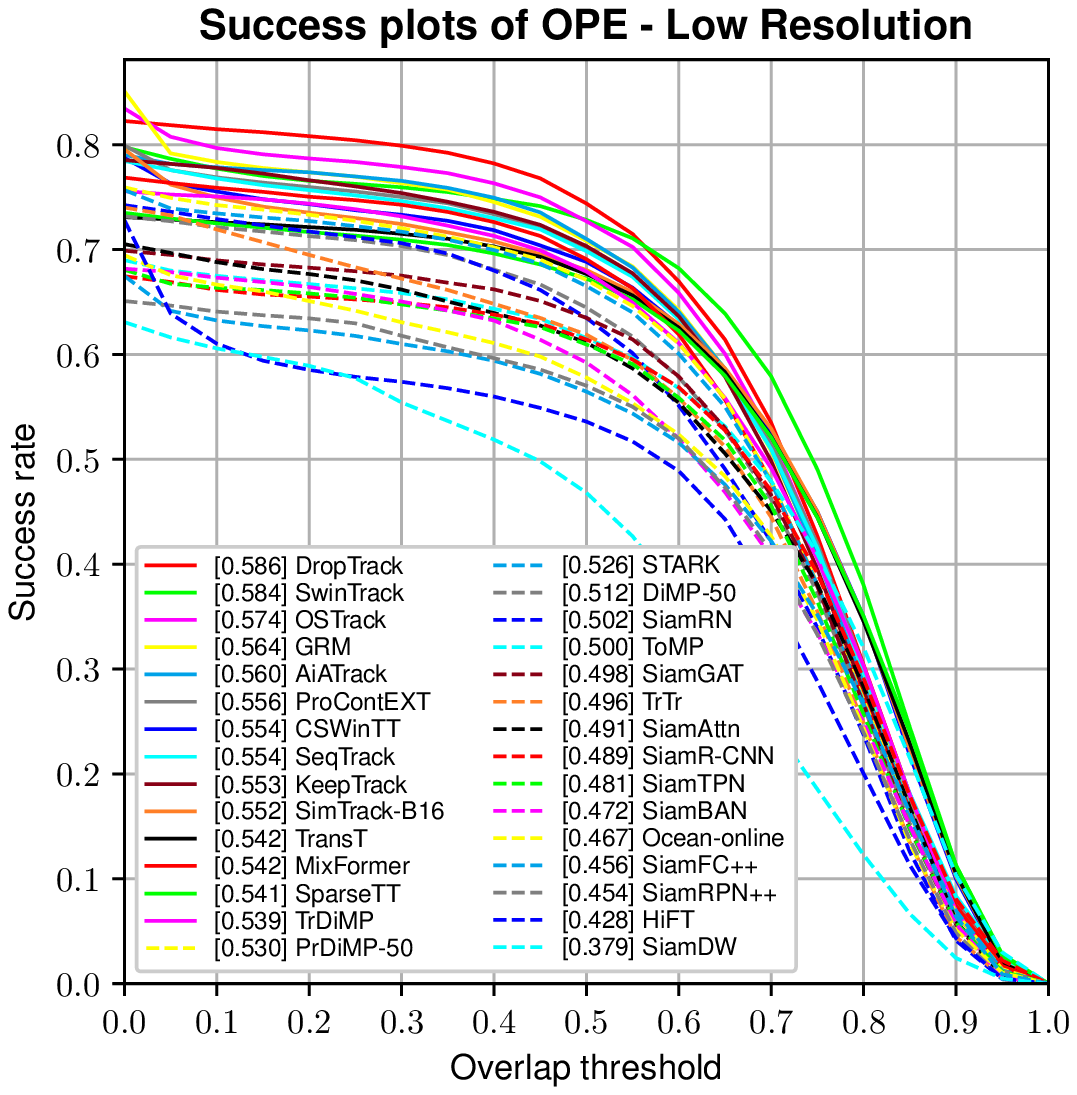}}
	\end{minipage}
	\begin{minipage}[b]{0.245\linewidth}
		\centering
		\centerline{\includegraphics[width=\textwidth]{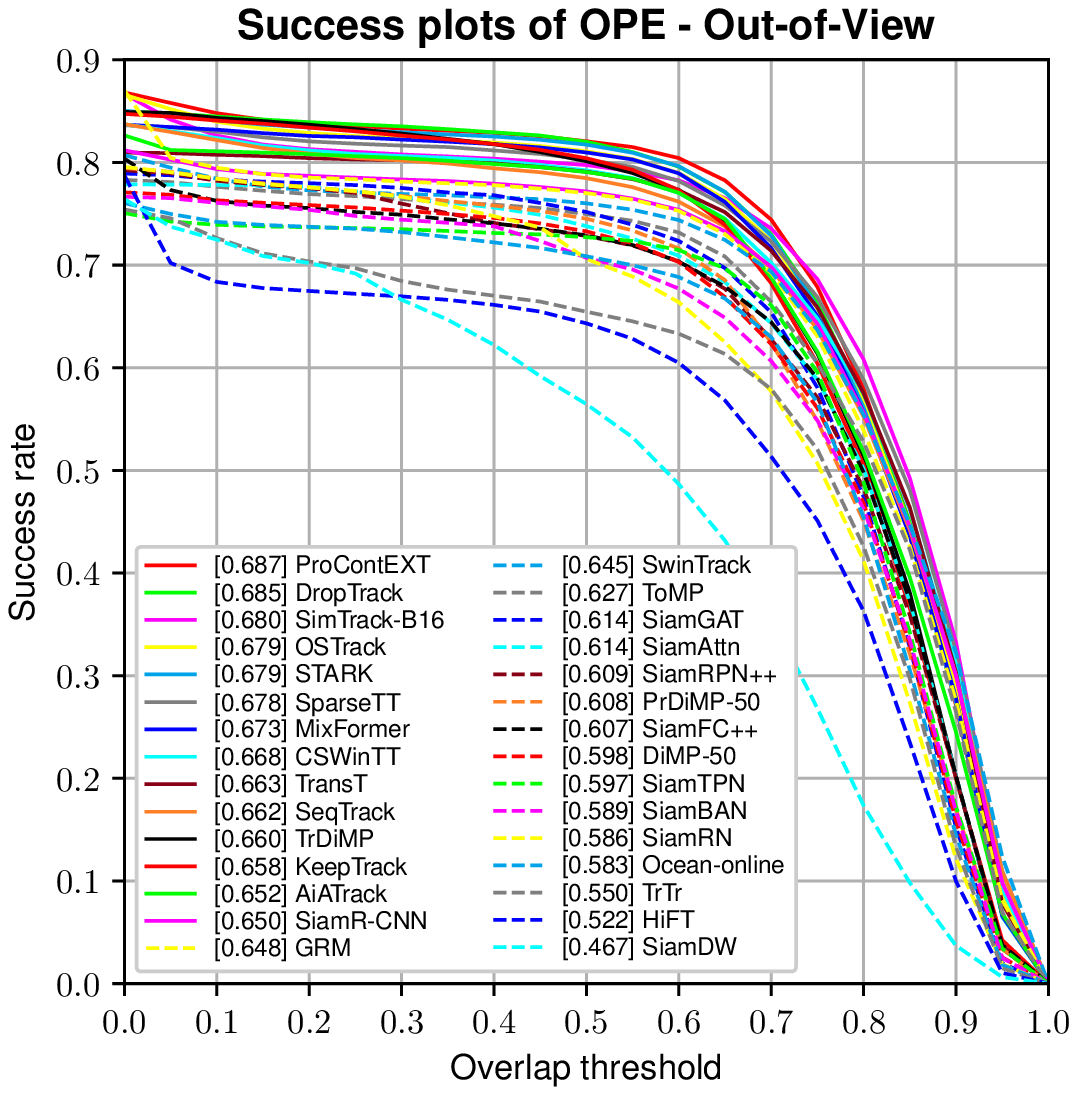}}
	\end{minipage}
	\begin{minipage}[b]{0.245\linewidth}
		\centering
		\centerline{\includegraphics[width=\textwidth]{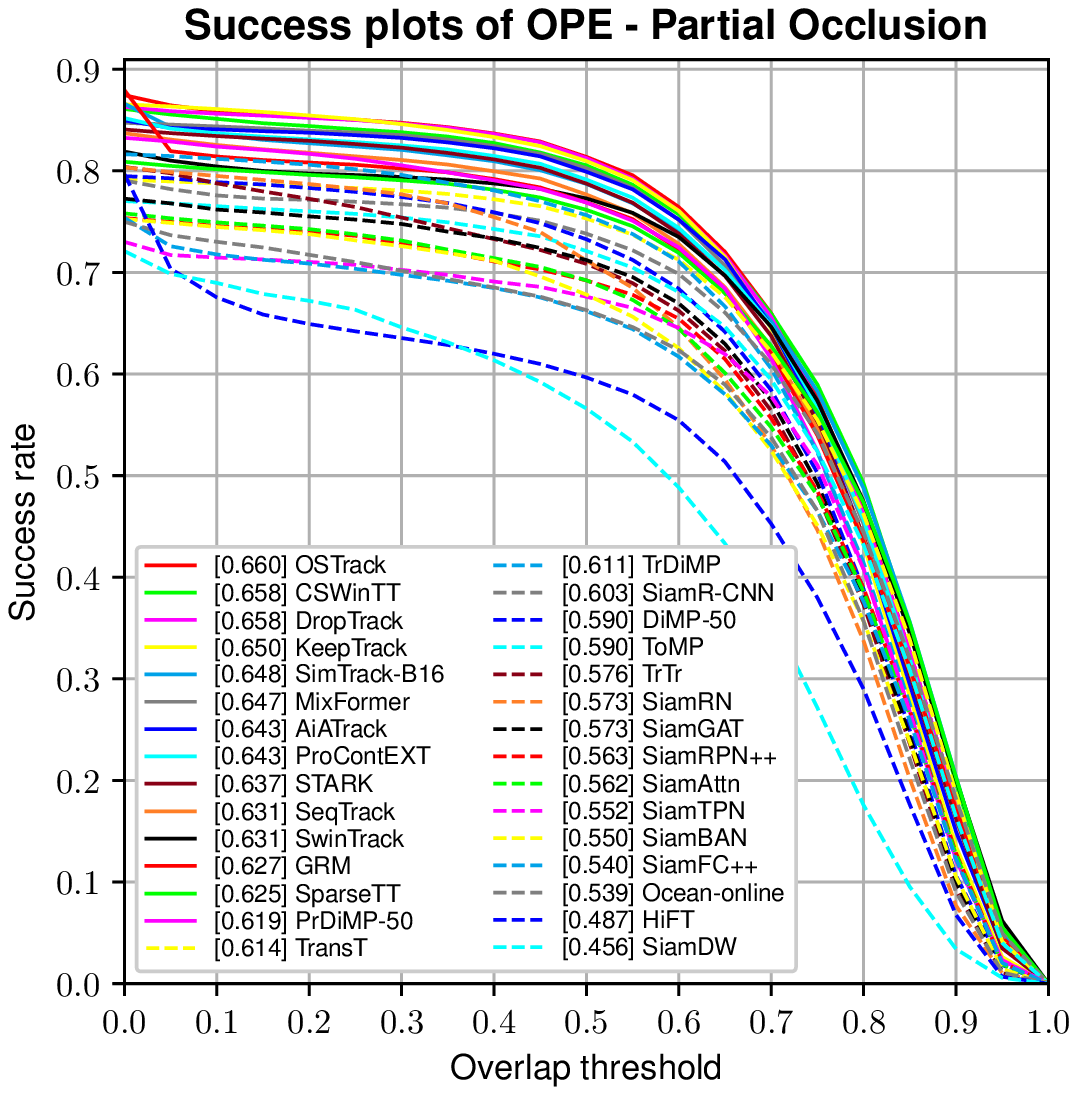}}
	\end{minipage}
	\begin{minipage}[b]{0.245\linewidth}
		\centering
		\centerline{\includegraphics[width=\textwidth]{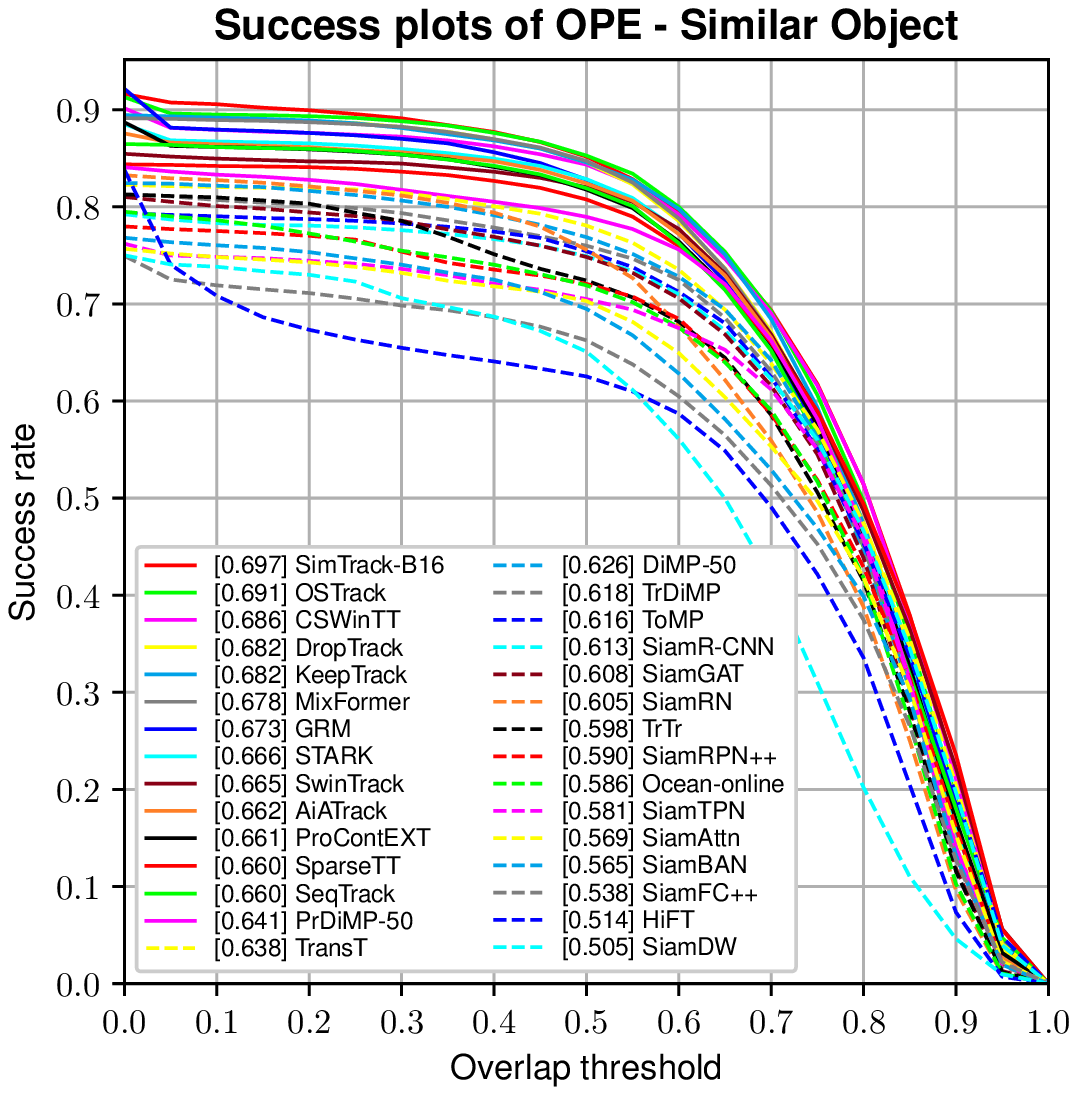}}
	\end{minipage}
	\begin{minipage}[b]{0.245\linewidth}
		\centering
		\centerline{\includegraphics[width=\textwidth]{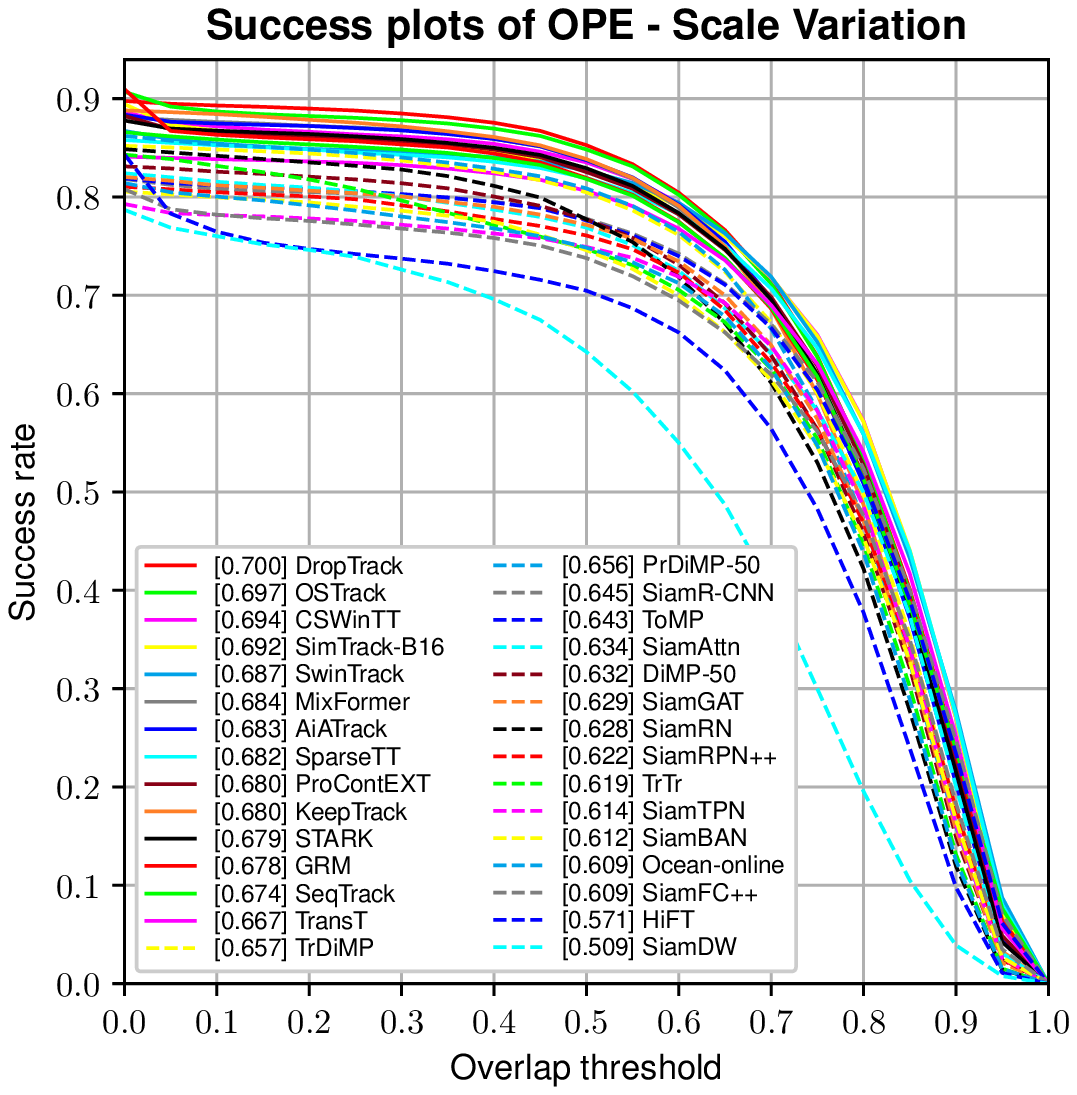}}
	\end{minipage}
	\begin{minipage}[b]{0.245\linewidth}
		\centering
		\centerline{\includegraphics[width=\textwidth]{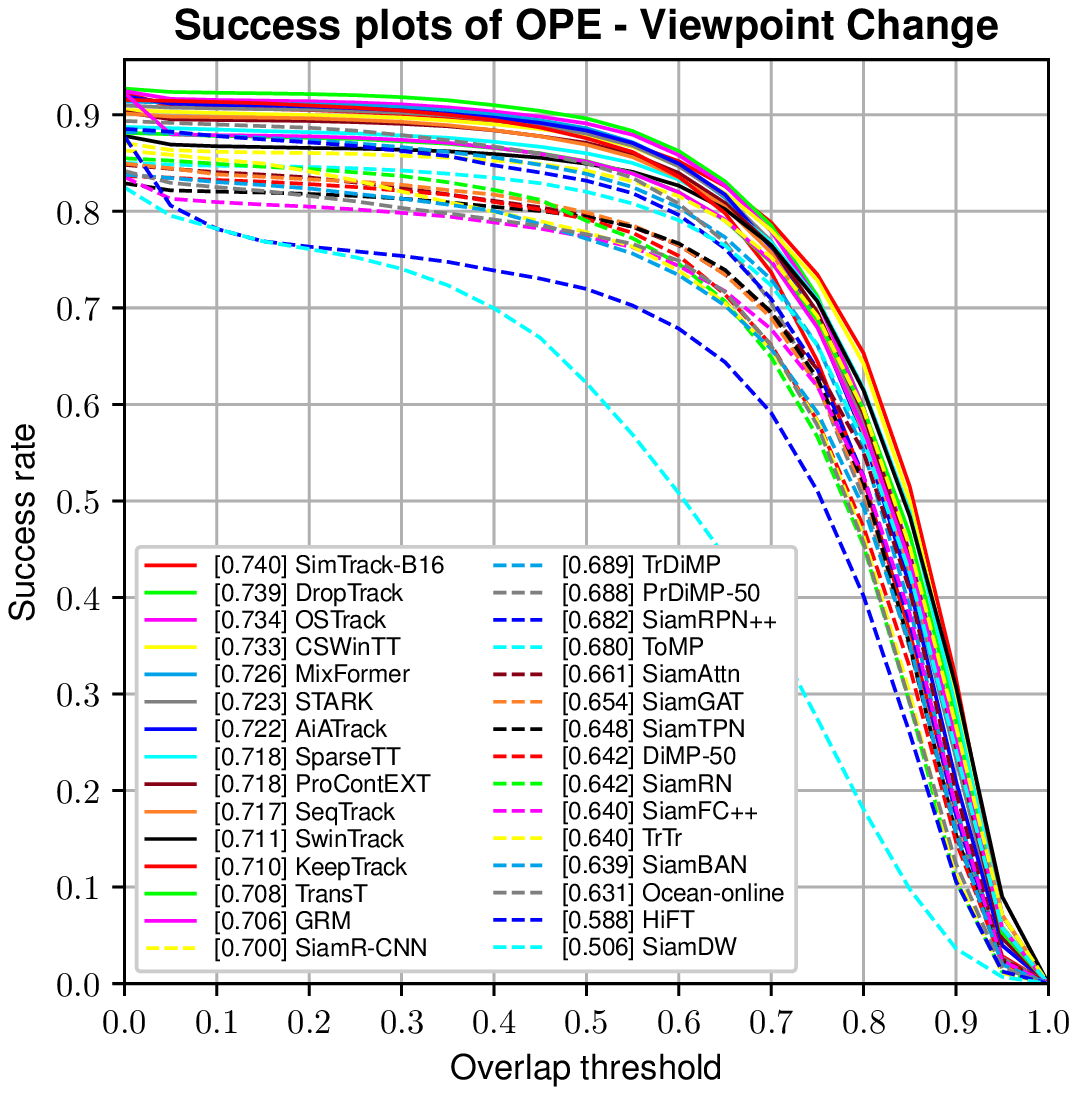}}
	\end{minipage}
	\caption[The success plots of the trackers  for 12 challenging attributes on UAV123 dataset]{The success plots of the trackers  for 12 challenging attributes on UAV123 dataset.}
	\label{fig3:Attribute_analysis_UAV}
\end{figure*}

Tracking a target in the UAV123 dataset is more difficult than in other benchmark datasets, since the target objects are relatively small in aerial tracking sequences. Therefore, the trackers can capture limited visual cues and are unable to rely on powerful appearance models. In addition, tracking a target in UAV123 is more challenging since the target object and camera frequently change position and orientation. We have used the official toolkit of UAV123 to measure the  precision and success scores of the trackers and to conduct the tracking attribute-wise evaluation. 

The One-stream One-stage trackers showed outstanding performances in UAV123 based on the overall precision and success rate scores due to their combined feature learning and future fusion process. Especially, the DropTrack \cite{wu2023dropmae} tracker showed excellent performance in UAV123 with a success rate score of 70.9\% and a precision score of 92.4\% due to its exceptional ability to capture correlated spatial cues. The  OSTrack \cite{bib78} and MixFormer \cite{bib80} approaches are also showed good results in UAV123 dataset. Among the CNN-Transformer based trackers,  CSWinTT \cite{bib67} and AiATrack \cite{bib73} approaches showed better success and precision scores, respectively. CNN-based tracker: KeepTrack \cite{mayer2021learning} obtained third highest precision score because of its distractor object handling capability without relying heavily on an appearance model.

Fully-Transformer based and CNN-Transformer based trackers showed excellent performance compared to the CNN-based trackers in every attribute on the UAV123 benchmark, as shown in Fig.~\ref{fig3:Attribute_analysis_UAV} and Table ~\ref{re:attribute}. 
The OSTrack \cite{bib78} and DropTrack \cite{wu2023dropmae}  trackers successfully handle the background clutter (BC), partial occlusion (POC), out-of-view (OV), and scale variation (SV) scenarios because of their strong discriminative capability and the background feature eliminating technique. On the other hand, CNN-Transformer based trackers: CSWinTT \cite{bib67} and AiATrack \cite{bib73} significantly outperforms fully-Transformer based trackers in fast motion (FM), illumination variation (IV), and aspect ratio change (ARC)  scenarios. 

Based on the experimental analysis, full occlusion (FOC), background clutter (BC), and low resolution (LR) are the most challenging attributes in UAV123 dataset since all the trackers are struggling to capture the strong appearance cues in these scenarios in aerial tracking videos. To summarize, the evaluation of state-of-the-art trackers on the UAV123 dataset showed that their performance was only average. This finding highlights the need for further research and innovation to improve the accuracy and reliability of aerial tracking systems.

\subsubsection{Analysis on LaSOT Dataset}

\begin{figure*}
	\centering
	\begin{minipage}[b]{0.245\linewidth}
		\centering
		\centerline{\includegraphics[width=\textwidth]{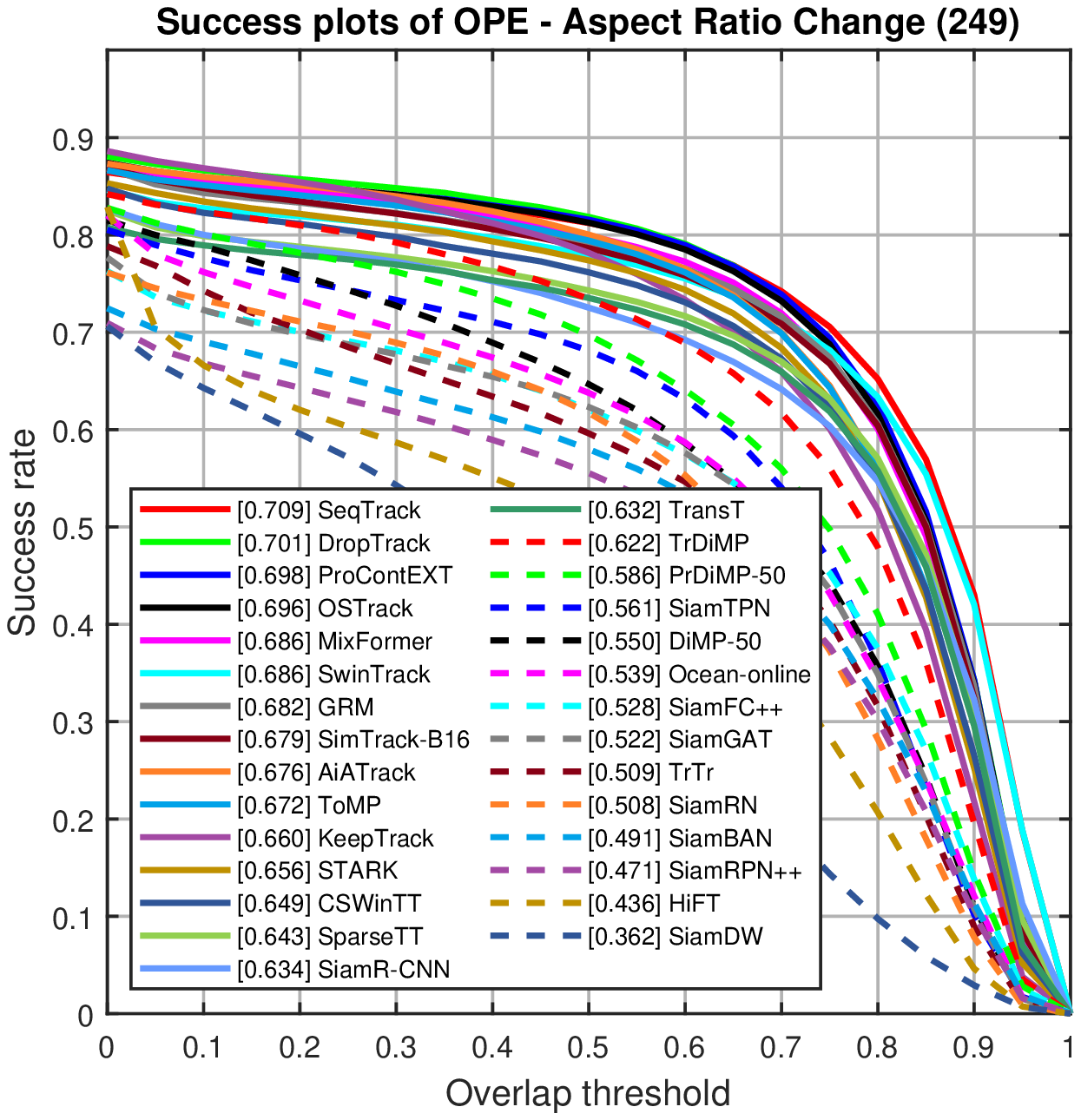}}
	\end{minipage}
	\begin{minipage}[b]{0.245\linewidth}
		\centering
		\centerline{\includegraphics[width=\textwidth]{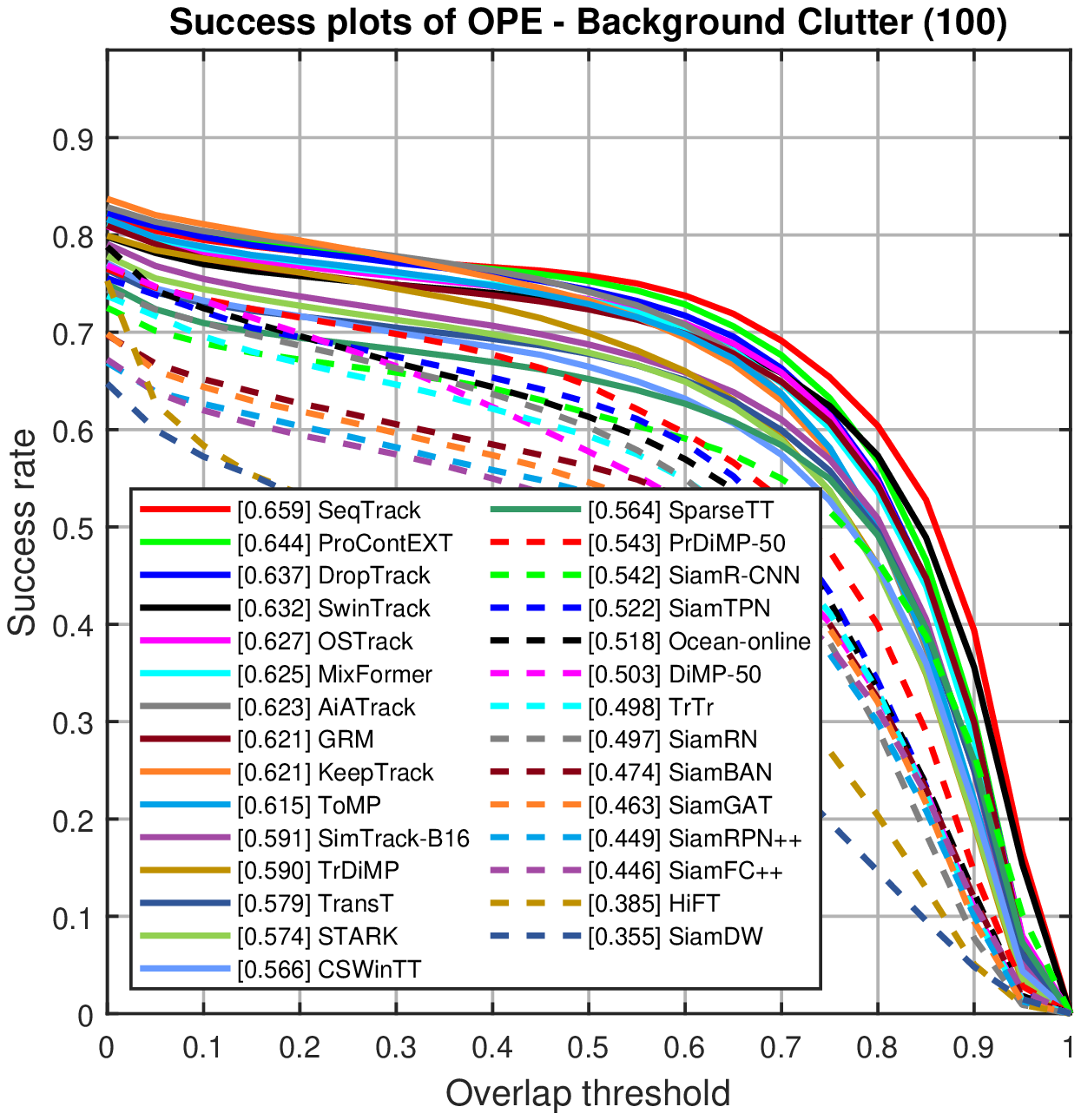}}
	\end{minipage}
	\begin{minipage}[b]{0.245\linewidth}
		\centering
		\centerline{\includegraphics[width=\textwidth]{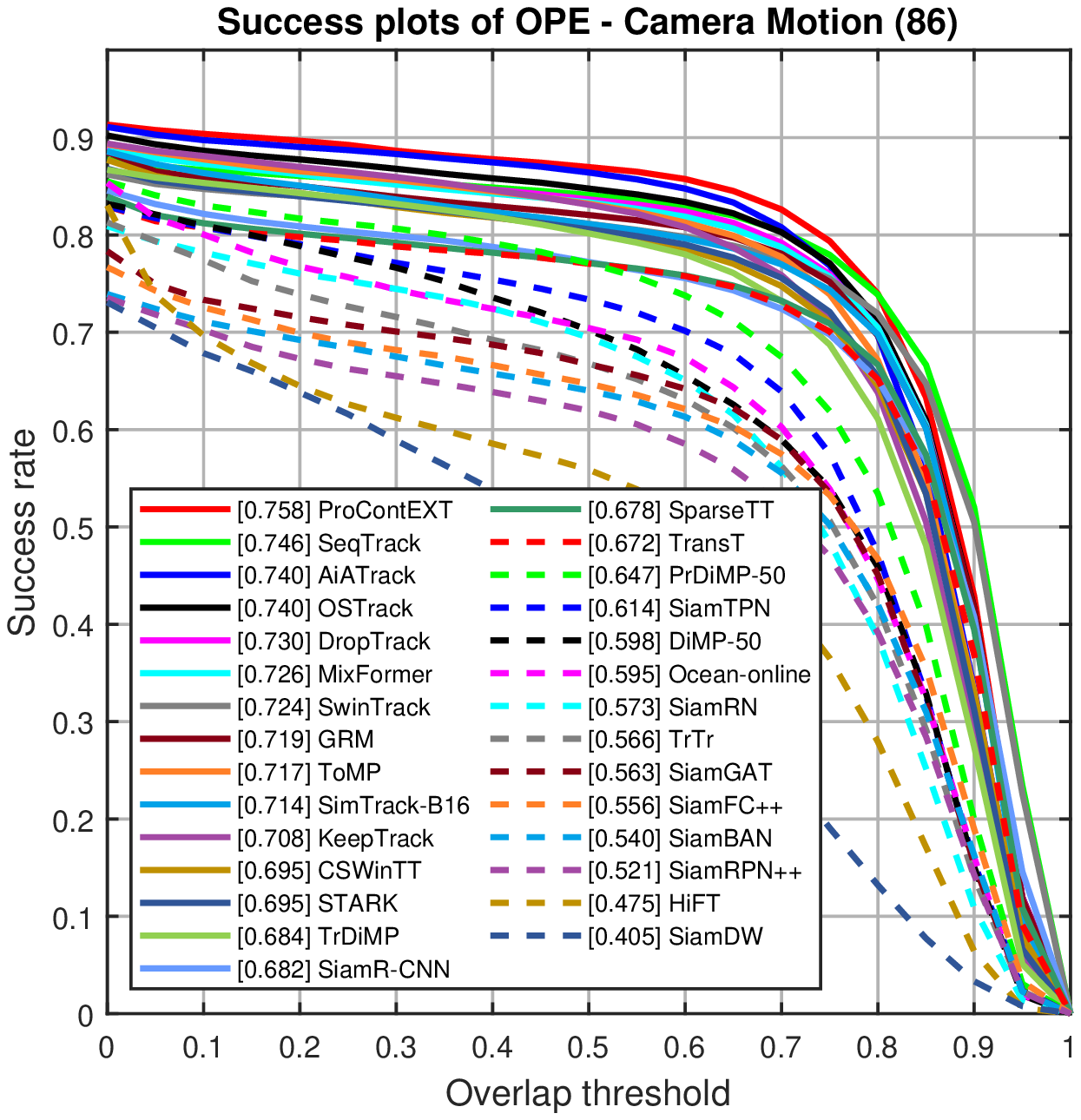}}
	\end{minipage}
	\begin{minipage}[b]{0.245\linewidth}
		\centering
		\centerline{\includegraphics[width=\textwidth]{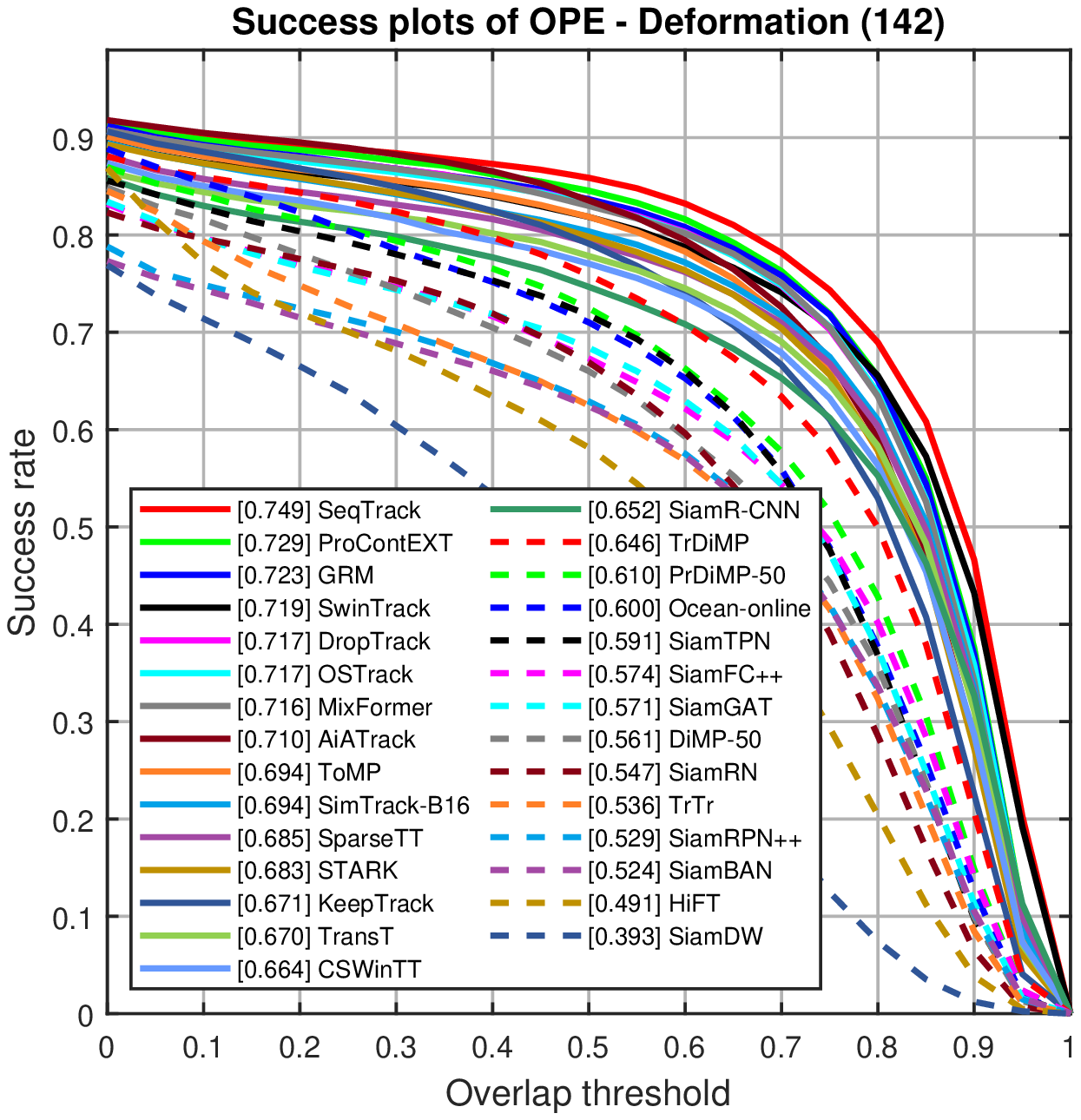}}
	\end{minipage}
	\begin{minipage}[b]{0.245\linewidth}
		\centering
		\centerline{\includegraphics[width=\textwidth]{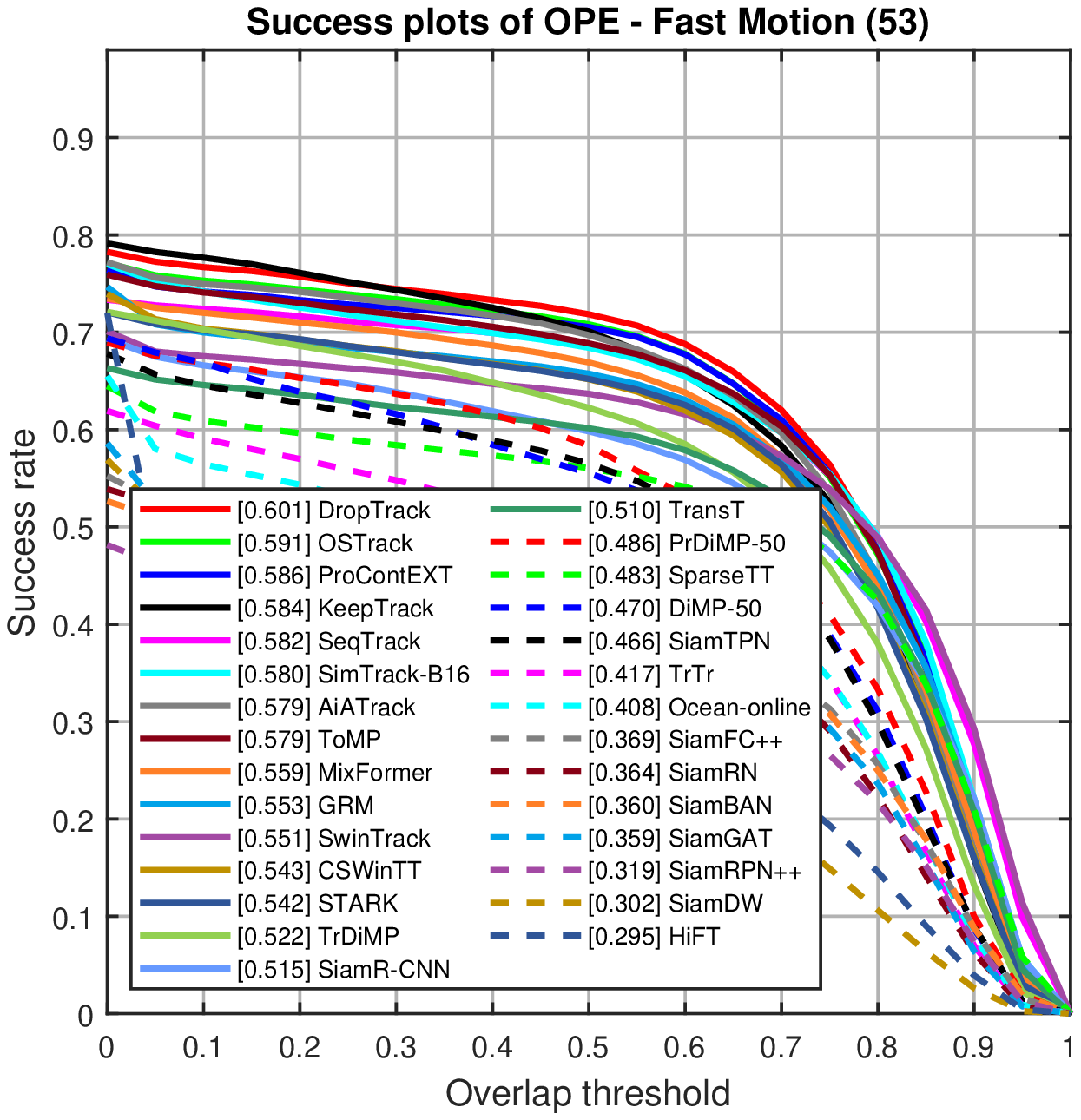}}
	\end{minipage}
	\begin{minipage}[b]{0.245\linewidth}
		\centering
		\centerline{\includegraphics[width=\textwidth]{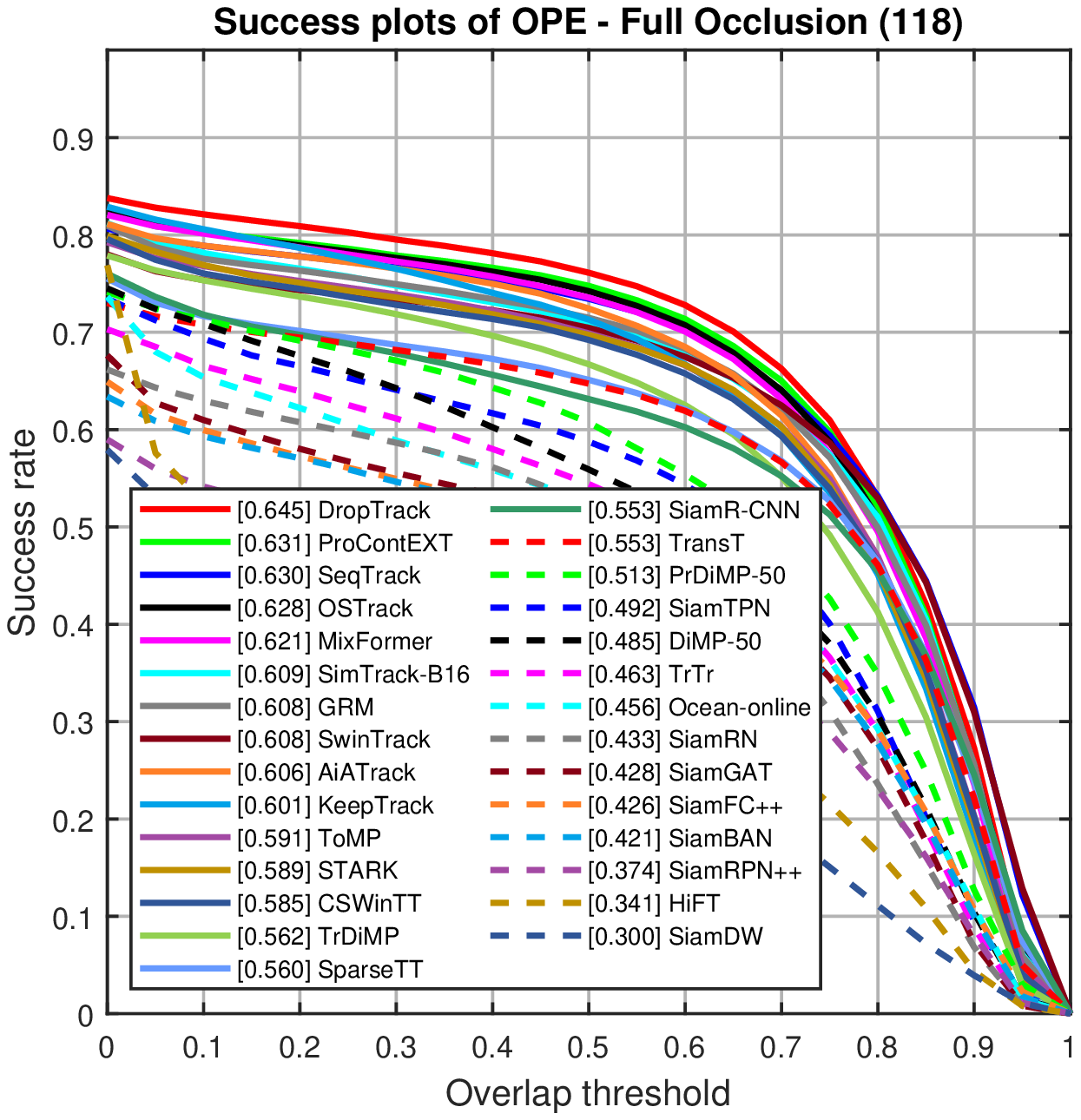}}
	\end{minipage}
	\begin{minipage}[b]{0.245\linewidth}
		\centering
		\centerline{\includegraphics[width=\textwidth]{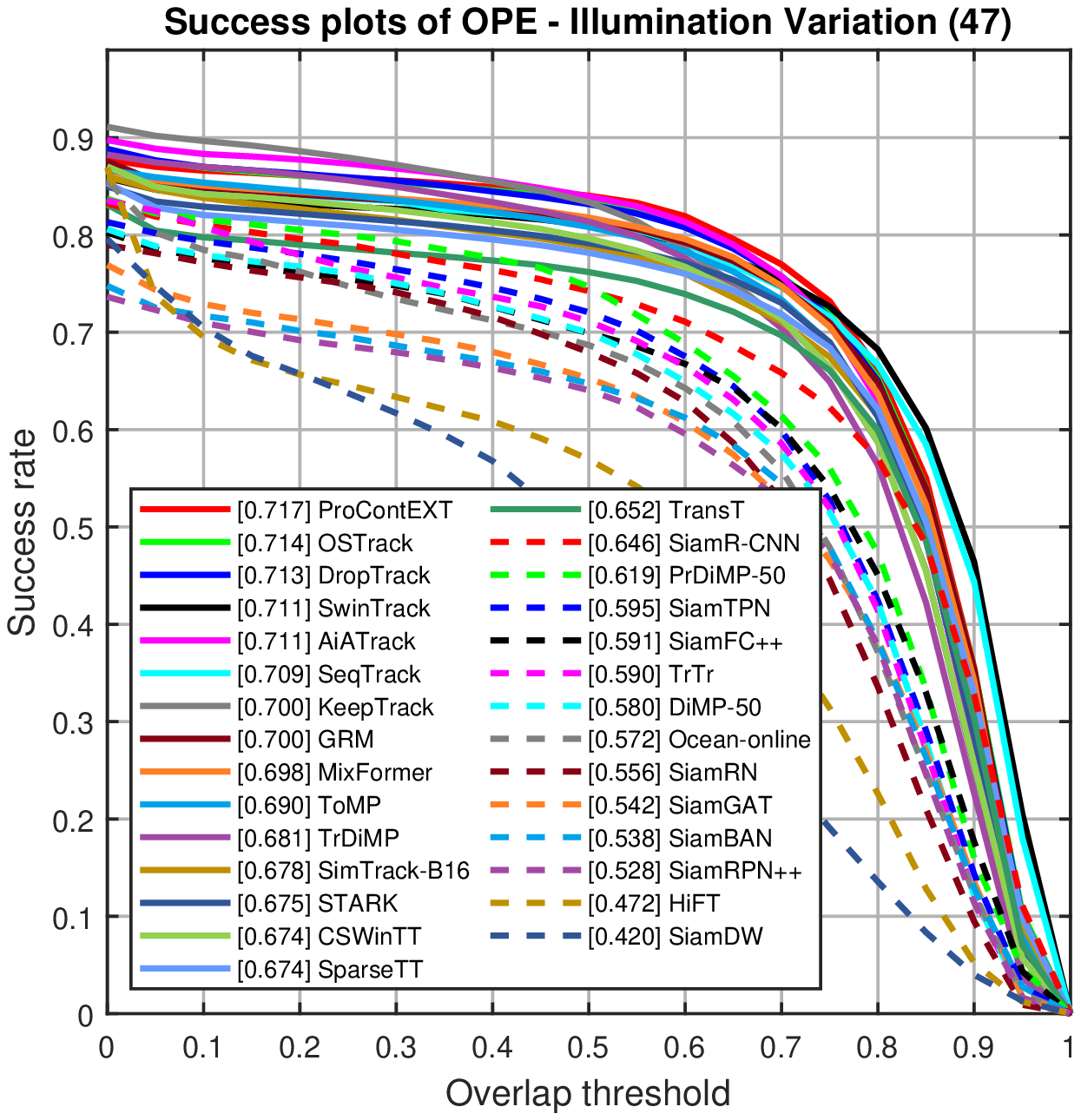}}
	\end{minipage}
	\begin{minipage}[b]{0.245\linewidth}
		\centering
		\centerline{\includegraphics[width=\textwidth]{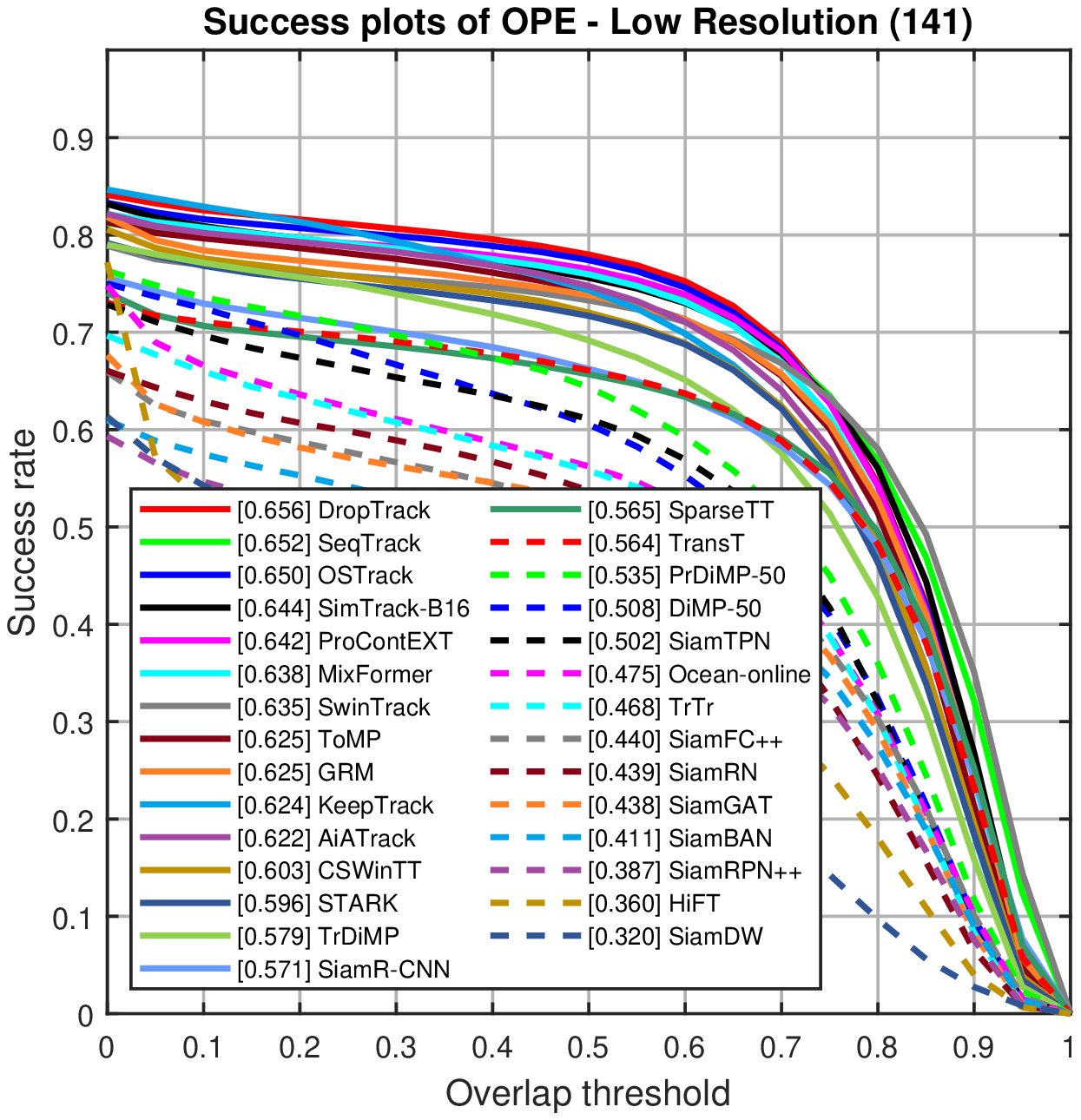}}
	\end{minipage}
	\begin{minipage}[b]{0.245\linewidth}
		\centering
		\centerline{\includegraphics[width=\textwidth]{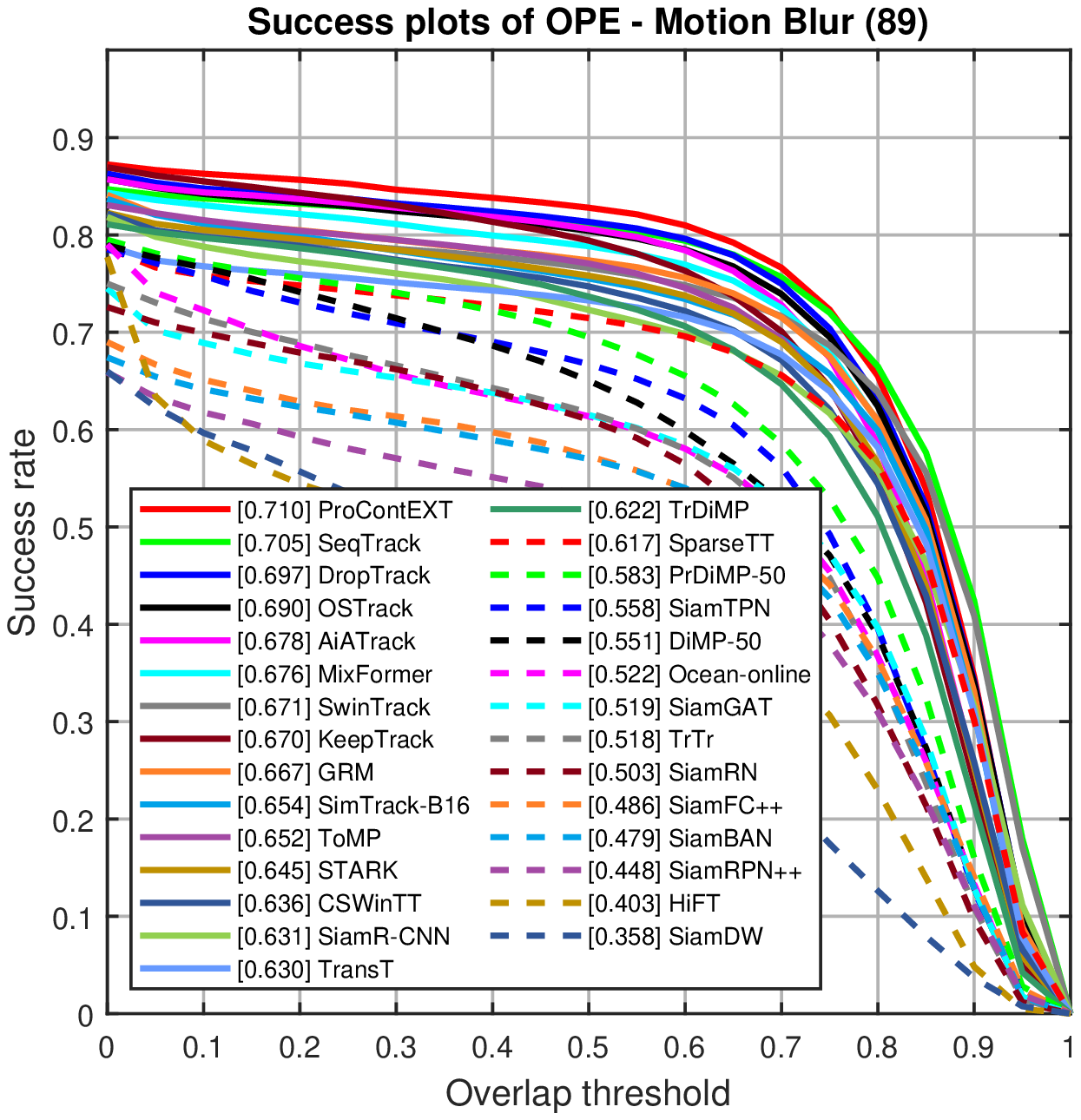}}
	\end{minipage}
	\begin{minipage}[b]{0.245\linewidth}
		\centering
		\centerline{\includegraphics[width=\textwidth]{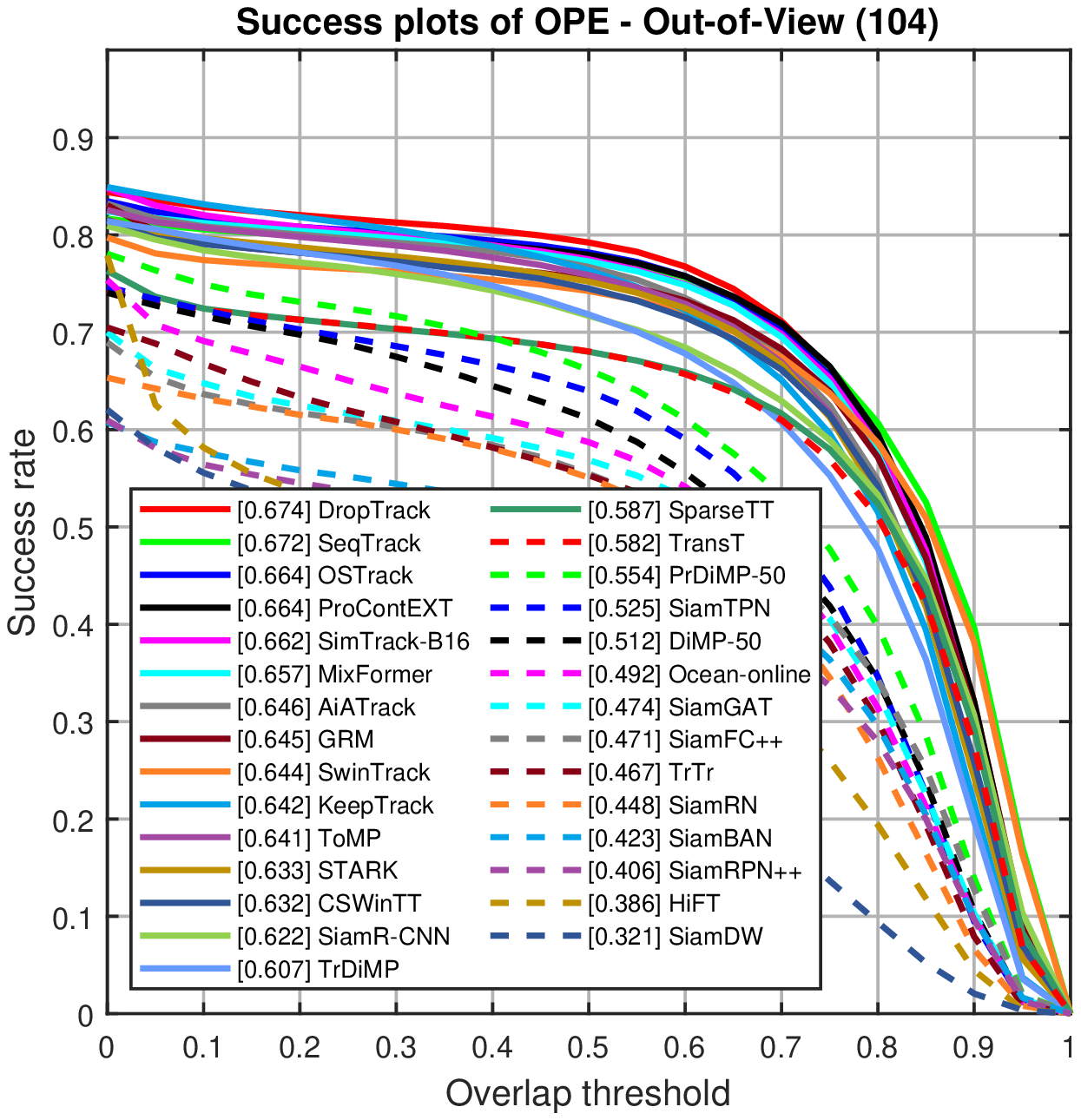}}
	\end{minipage}
	\begin{minipage}[b]{0.245\linewidth}
		\centering
		\centerline{\includegraphics[width=\textwidth]{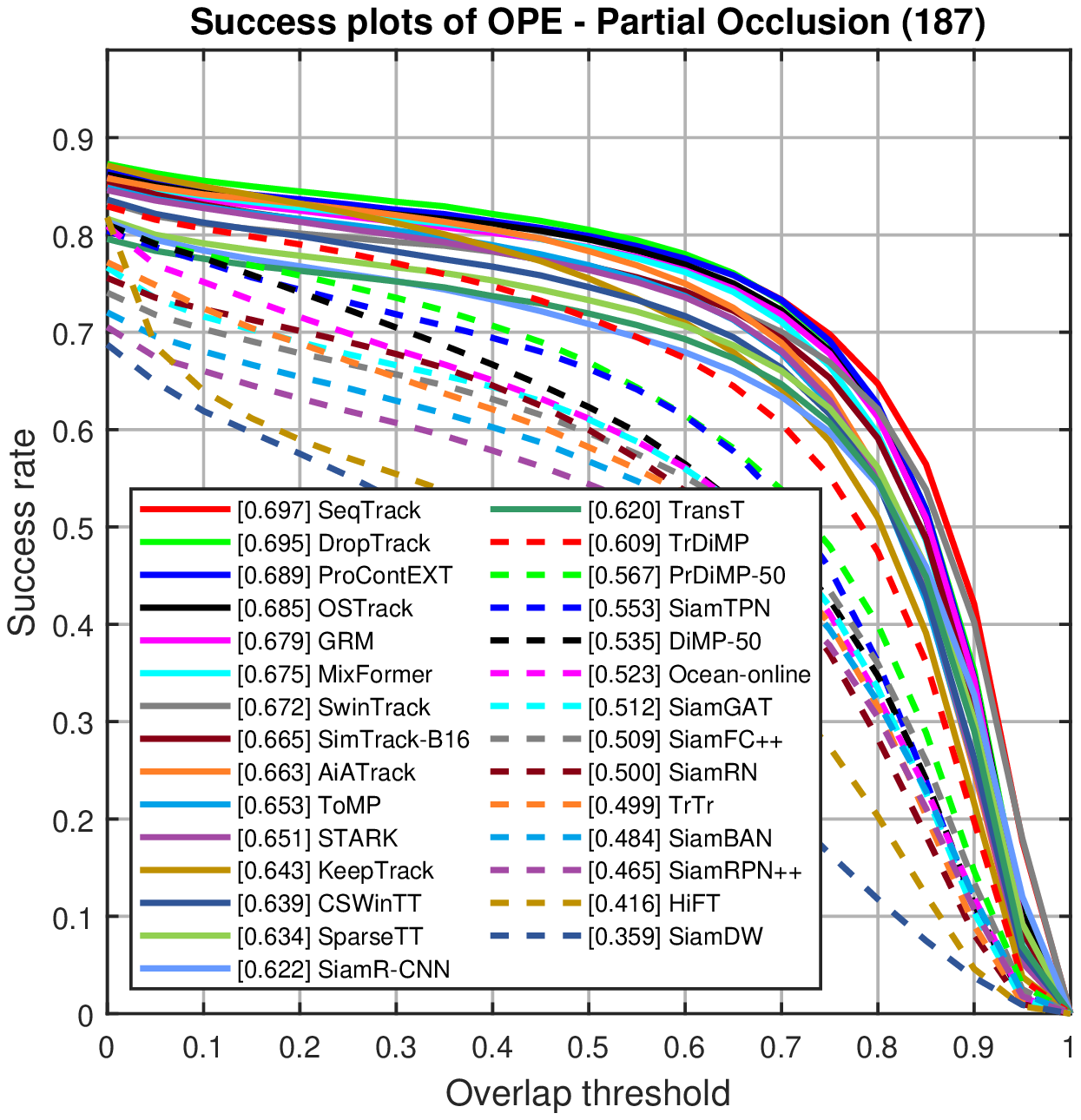}}
	\end{minipage}
	\begin{minipage}[b]{0.245\linewidth}
		\centering
		\centerline{\includegraphics[width=\textwidth]{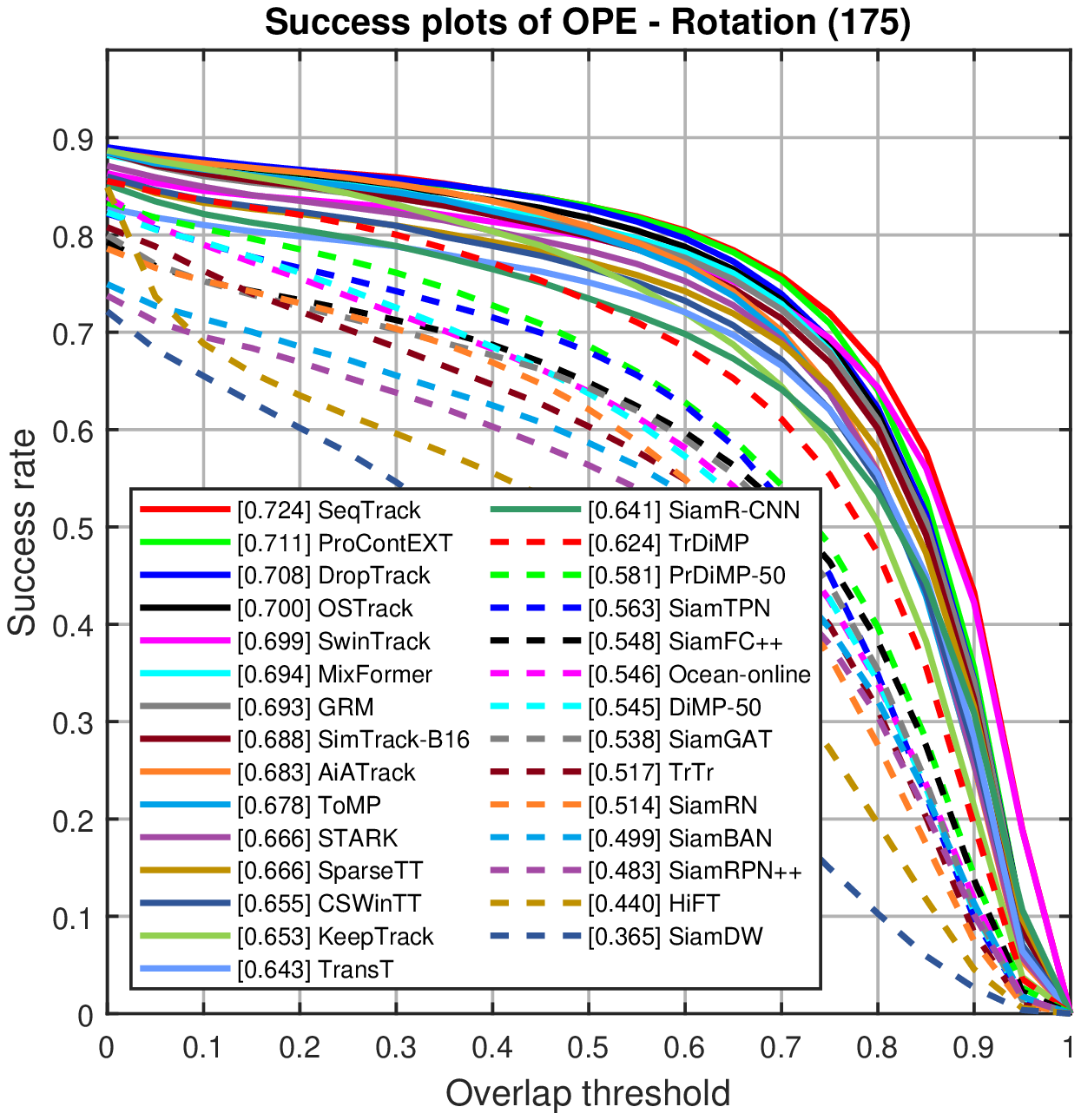}}
	\end{minipage}
	\begin{minipage}[b]{0.245\linewidth}
		\centering
		\centerline{\includegraphics[width=\textwidth]{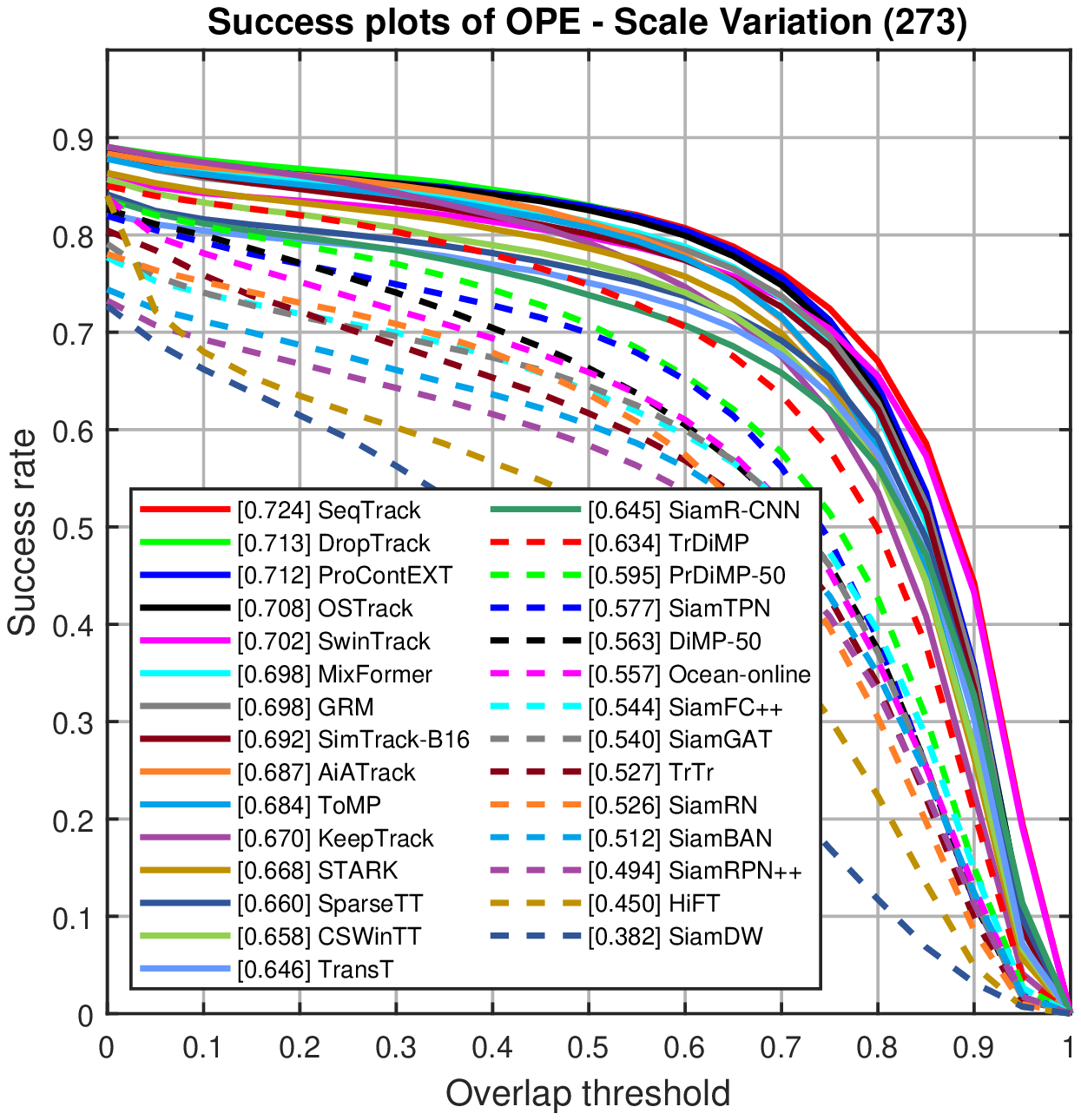}}
	\end{minipage}
	\begin{minipage}[b]{0.245\linewidth}
		\centering
		\centerline{\includegraphics[width=\textwidth]{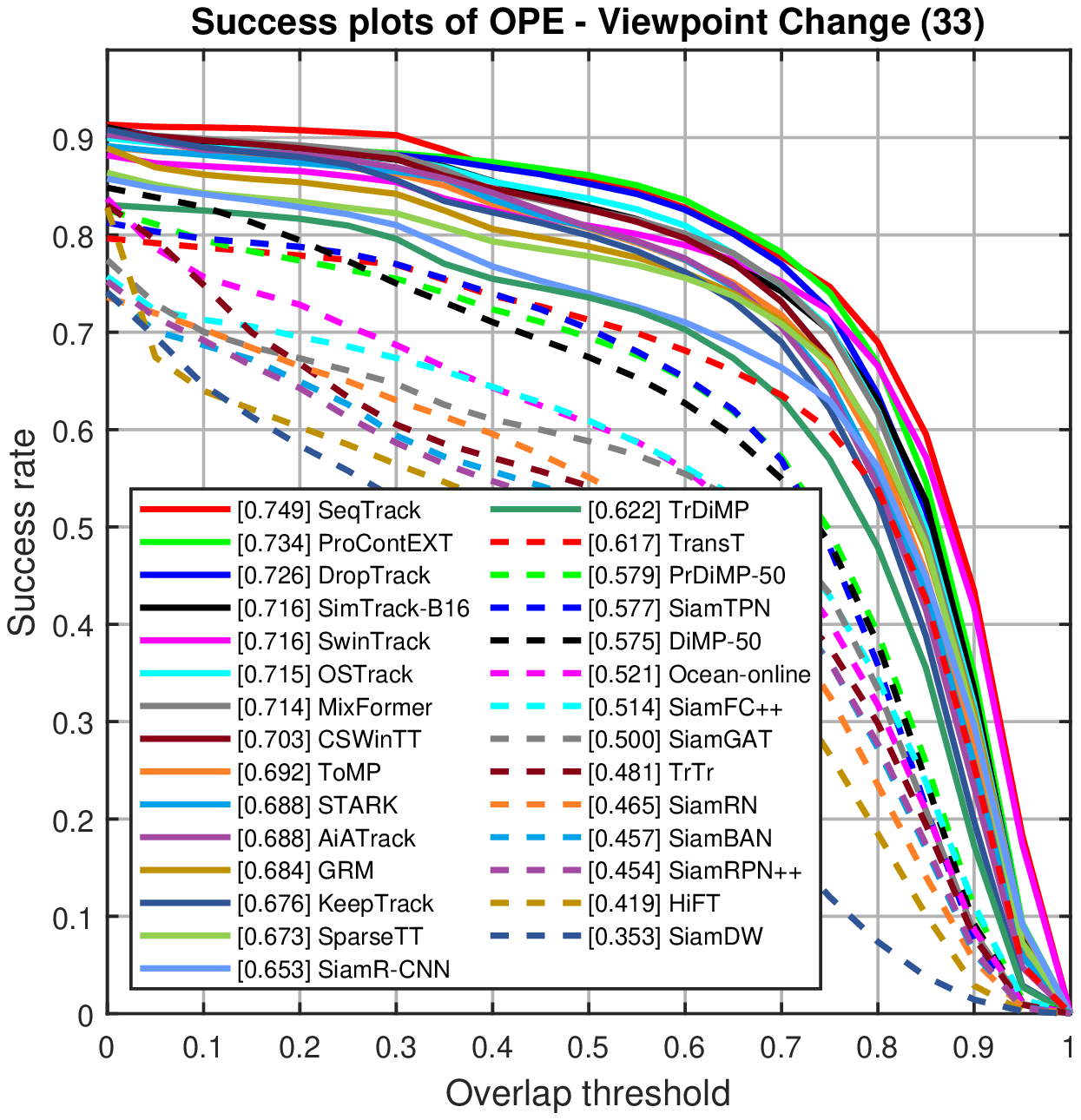}}
	\end{minipage}
	\caption[The success plots comparison of trackers on LaSOT for 14 challenging attributes]{The success plots comparison of trackers on LaSOT for 14 challenging attributes.}
	\label{fig4:Attribute_analysis_LaSOT}
\end{figure*}

Experimental analysis  on LaSOT dataset is important to identify the future direction of VOT since it has long-term tracking sequences with several challenging scenarios. We have utilized the official toolkit of the LaSOT dataset to measure the tracking performance of the approaches. 

As summarized in the overall performance results in Table ~\ref{re:3}, One-stream One-stage  tracker SeqTrack \cite{chen2023seqtrack} showed outstanding performance in LaSOT in terms of success rate, normalized precision, and precision scores since it trained the Transformer by treating tracking as a sequence learning problem rather than template matching. The DropTrack \cite{wu2023dropmae} approach also demonstrated excellent performance in LaSOT due to the correlated spatial feature learning capability of its backbone network. In addition to these trackers, the ProContEXT \cite{lan2022procontext} and OSTrack \cite{bib78} approaches showed good  performances since they efficiently used the information flow between the target template and search region and hence eliminated the background features in feature matching. Overall, all One-stream One-stage fully-Transformer based trackers significantly outperformed other tracker types with a large margin in the LaSOT dataset, as the target-aware Transformer features are better suited for long-term tracking than CNN-based local feature extraction.

Based on the attribute-wise success plots in Fig.~\ref{fig4:Attribute_analysis_LaSOT}, and Table ~\ref{re:attribute}, SeqTrack \cite{chen2023seqtrack} and DropTrack \cite{wu2023dropmae} trackers showed best performance in most of the tracking scenarios in the LaSOT benchmark. In particular, the SeqTrack tracker outperforms the competitors with a large scale in background clutter (BC), deformation (DFE), scale variation (SV), and view point change (VC) scenes because of its sequence learning mechanism. In addition to these trackers, the ProContEXT \cite{lan2022procontext} tracker showed better performance in camera motion (CM), illumination variation (IV), and motion blur (MB) scenarios because of its strong discriminative capability and background feature removal mechanism.

Overall, One-stream One-stage fully-Transformer  trackers showed excellent performance, while CNN-Transformer based trackers showed considerable performance in the LaSOT benchmark. On the other hand, the overall tracking and attribute-wise performances of CNN-based approaches are very limited in the LaSOT dataset, as they fail to include temporal cues and extract target-specific features in search regions. Based on the attribute-wise success rates, fast motion and full occlusion are the most challenging attributes for state-of-the-art trackers. 

\subsubsection{Analysis on TrackingNet Dataset}
TrackingNet dataset has more than 30k training and 511 testing video sequences with 14 million and 225k annotations, respectively. Since the TrackingNet dataset has the videos with a large diversity of resolutions, target object classes, and frame rates, evaluating the tracking performance on this dataset is important for many real-world applications. Similar to LaSOT dataset, we  ranked the trackers in TrackingNet dataset using  success rate, precision score, and normalized precision. Since the test set annotations are not publicly available, we are unable to conduct the attribute-wise comparison in this dataset. 

Similar to the LaSOT benchmark,  the  SeqTrack \cite{chen2023seqtrack} tracker showed superior performance in TrackingNet dataset with 85.5\% of success rate, 89.8\% of normalized precision, and 85.8\% of precision scores. Also, other One-stream One-stage based fully-Transformer trackers:  ProContEXT \cite{lan2022procontext} and DropTrack \cite{wu2023dropmae} approaches also showed competitive tracking performances. Moreover, the GRM \cite{gao2023generalized} and MixFormer \cite{bib80} achieved the third rank in terms of precision and normalized precision, respectively. 

Experimental analysis on TrackingNet dataset showed that fully-Transformer based trackers significantly outperformed other two categories by a large margin. Although CNN-Transformer based trackers: AiATrack \cite{bib73} and ToMP \cite{bib75} obtained considerable competitive performances, tracking accuracy of other CNN-Transformer based trackers are lower. On the other hand, similar to the LaSOT dataset, all CNN-based trackers are struggling to show good performance in TrackingNet dataset. The experimental results clearly show that fully-Transformer based feature extraction and fusion successfully handle real-world tracking scenarios, such as various resolutions and frame rates, better than CNN-based feature extraction and target localization.

\subsubsection{Analysis on GOT-10k dataset}
Evaluating performance on the GOT-10k dataset is essential for VOT approaches, as it has a large diversity of target object classes. This allows for measuring the one-shot tracking performance of the trackers, i.e., the performance when the target object class is not included in the training. Furthermore, the evaluation results of this dataset are not biased towards familiar objects, as the tracking models are only trained on the training set of GOT-10k. We strictly adhere to the evaluation protocols to measure the generalization capability of trackers. Average Overlap and success rate at the threshold of 0.5 ($\mathrm{SR_{0.5}}$) and 0.75 ($\mathrm{SR_{0.75}}$) are used to rank the trackers in GOT-10k dataset. 

Based on the obtained results in Table ~\ref{re:3}, the DropTrack \cite{wu2023dropmae} tracker achieved the highest average overlap score of 75.9\% and $\mathrm{SR_{0.5}}$ (success rate at 0.5 threshold) of 86.6\%. The ProContEXT \cite{lan2022procontext} and SeqTrack \cite{chen2023seqtrack} trackers also demonstrated competitive performances.  Overall, the experimental results showed that all One-stream One-stage fully-Transformer based trackers possess a good generalization capability, as their attention mechanism perform effectively across a diverse range of target object classes. On the other hand, the tracking performances of CNN-based trackers are lower in the GOT-10k dataset, as CNNs are not as effective in handling unfamiliar objects.

\subsection{Tracking Efficiency Analysis}\label{efficiency}
Analyzing tracking efficiency is as important as analyzing tracking performance, as it is crucial for many practical situations and real-world applications.   In this efficiency analysis comparison, we have included all the recent trackers except a few approaches since their source codes and tracking models are not publicly available. To conduct the unbiased comparison, efficiency results of each tracker is obtained by executing their source codes on a computer with a NVIDIA Quadro P4000 GPU and a 64GB of RAM. We did not change the  parameters of the tracking models of these approaches and obtained the efficiency results by evaluating the trackers on the LaSOT benchmark dataset. 

\begin{table*}
	\begin{center}
		\footnotesize
		\caption{Tracking efficiency comparison of the recently proposed state-of-the-art trackers on LaSOT dataset. The search region size, number of parameters (in million), tracking speed (in frame per second), number of floating point operations (FLOPs) of each tracker are listed. The efficiency experiments are conducted on a NVIDIA Quadro P4000 GPU with a 64GB of RAM. The symbols \dag and \ddag are denoting One-stream One-stage trackers, and Two-stream Two-stage trackers, respectively.}\label{re:effiency}%
		\begin{tabular}{l l @{\extracolsep{0.1cm}}*{2}{d{4.3}} l  @{\extracolsep{0.1cm}}*{2}{d{4.3}}}
			\toprule
			&\multirow{2}{*}{\textbf{Trackers}} & \mc{\textbf{Tracking}} & \mc{\textbf{No. of}}& \mc{\textbf{Search}} & \multicolumn{1}{c}{\multirow{2}{*}{\textbf{FLOPs (G)}}}& \mc{\textbf{Success}}\\
			&	 &	 \mc{\textbf{Speed (FPS)}} & \mc{\textbf{Parameters (M)}}		&	\mc{\textbf{Reign Size}}	&	 & \mc{\textbf{ Rate}}\\
			\midrule
			\multirow{9}{*}{\rotatebox[origin=c]{90}{\parbox[c]{2.5cm}{\centering Fully-Transformer based Trackers}}} 
			& SeqTrackL-384 \cite{chen2023seqtrack}\dag & 5.81 & 308.98  & 384 $\times$ 384 & 535.85  & 72.5
			\\
			& DropTrack \cite{wu2023dropmae}\dag & 20.64 & 92.83  & 384 $\times$ 384 & 48.32  &71.5
			\\
			& ProContEXT \cite{lan2022procontext} \dag & 10.44 & 92.83 & 384 $\times$ 384 & 85.27 & 71.4
			\\
			& OSTrack-384 \cite{bib78}\dag & 20.64 & 92.83  & 384 $\times$ 384 & 48.32  & 71.1
			\\
			& SwinTrack-B-384 \cite{bib38}\ddag & 11.92 & 90.96   &384 $\times$ 384 & 61.85 & 70.2 
			\\
			& MixFormer-L \cite{bib80}\dag & 8.02  & 195.40  & 320 $\times$ 320 & 113.02 & 70.0
			\\
			& GRM-B \cite{gao2023generalized}\dag & 36.02 & 99.83  & 256 $\times$ 256 & 30.90  & 69.9
			\\			
			& SimTrack-B/16 \cite{bib79}\dag & 36.34 & 88.64 & 224 $\times$ 224 & 22.26 & 69.3
			\\
			& SparseTT \cite{bib77}\ddag & 30.50  & 46.33 & 289 $\times$ 289 & 9.21		&	66.0
			\\			
			\midrule 
			\multirow{9}{*}{\rotatebox[origin=c]{90}{\parbox[c]{2.5cm}{\centering CNN-Transformer based Trackers}}} & AiATrack \cite{bib73} & 31.22 &17.95	 & 320 $\times$ 320 & 9.45 & 69.0
			\\
			& ToMP-101 \cite{bib75}  & 13.16  & 66.86 & 288 $\times$ 288 & 22.12 & 68.5
			\\
			& STARK-ST101 \cite{bib36} & 17.90 & 47.17  & 320 $\times$ 320 & 18.48 & 67.1
			\\
			& CSWinTT \cite{bib67} & 8.76 &28.23 & 384 $\times$ 384 &  16.45 & 66.2
			\\
			& TransT \cite{bib37}  & 21.15 &23.02 & 256 $\times$ 256 & 16.71 & 64.9
			\\
			& TrDiMP \cite{bib66} & 15.75 & 43.75 &352 $\times$ 352 & 16.59 & 63.9
			\\
			& SiamTPN(Shuffle) \cite{xing2022siamese} & 31.86 &6.17  & 256 $\times$ 256 &  2.09 & 58.1
			\\
			& TrTr-online \cite{zhao2021trtr} & 16.28  &11.93 & 255 $\times$ 255 & 21.23 & 52.7
			\\			
			& HiFT \cite{bib76} & 37.06 &11.07  & 287 $\times$ 287 & 6.47 & 45.1
			\\
			\midrule
			\multirow{11}{*}{\rotatebox[origin=c]{90}{\parbox[c]{3cm}{\centering CNN-based Trackers}}} & KeepTrack \cite{mayer2021learning} & 8.11 & 43.10  & 352 $\times$ 352 & 28.65 & 67.3
			\\
			& PrDiMP-50 \cite{danelljan2020probabilistic} & 18.05	& 43.10	 & 288 $\times$ 288 & 15.51 & 59.9
			\\
			& DiMP-50 \cite{bhat2019learning} & 30.67 &43.10  & 288 $\times$ 288 & 10.35	 &	56.8
			\\
			& Ocean \cite{zhang2020ocean} & 18.60  &37.16	 & 255 $\times$ 255 & 20.26 & 56.0
			\\
			& SiamAttn \cite{yu2020deformable} & 15.72 &140.24 & 255 $\times$ 255	& - & 56.0
			\\
			& SiamFC++ (GoogLeNet) \cite{xu2020siamfc++}& 45.27 &13.89  & 303 $\times$ 303 & - & 54.3
			\\
			& SiamGAT \cite{guo2021graph} & 41.99 &14.23  & 287 $\times$ 287 & 17.28 & 53.9
			\\
			& SiamRN \cite{cheng2021learning} & 6.51 &56.56  & 255 $\times$ 255 & 116.87 & 52.7
			\\						
			&SiamBAN \cite{chen2020siamese} & 23.71 &53.93  & 255 $\times$ 255 & 48.84 & 51.4
			\\		
			& SiamRPN++ \cite{li2019siamrpn++} & 5.17 &53.95 & 255 $\times$ 255 & 48.92 & 49.6
			\\
			& SiamDW (CIRResNet22-FC) \cite{zhang2019deeper}& 52.58 &2.46  & 255 $\times$ 255 & 12.90 & 38.5
			\\			
			\bottomrule
		\end{tabular}
	\end{center}
\end{table*}

We have evaluated the efficiency of the trackers in terms of their tracking speed, number of parameters, and the number of floating point operations (FLOPs) in their tracking model. Tracking speed is important for many real-world applications and hence we have considered it as an important metric in efficiency comparison. It is computed by calculating the average number of frames processed by an approach in a second. Since the reported tracking speeds of the approaches depend on the hardware and implementation platforms, we calculated the tracking speed of the trackers on the same hardware platform using the PyTorch deep learning framework. 

The number of parameters is another efficiency metric in deep learning-based tracking approaches, as models with fewer parameters are more hardware-efficient and can work on small devices such as mobiles and tablets.  Number of parameters of a tracking approach are the total number of arguments passed to the optimizer and in most situations it does not rely on the input size of the tracking approach. We have used the PyTorch default function to measure the total number of parameters of a tracking model. We have considered the number of floating point operations (FLOPs)  as the third  metric to measure the efficiency of a tracking model. Except for fully-Transformer based trackers, the number of FLOPs of an approach depends on the tracking model and corresponding search image size since the target template features are computed only in the first frame of a tracking sequence. In fully-Transformer based trackers, features of the target template are computed in each and every frame, and hence the size of the template influences the FLOPs. Although the number of FLOPs of some trackers is very high, they are still able to track a target at high speed since their models are highly parallel, and nowadays GPUs can handle them successfully. However, these approaches are not suitable for some applications that only run on CPUs and mobile devices.  The overall efficiency results of the trackers, along with their corresponding search image size and success scores on the LaSOT benchmark, are reported in Table ~\ref{re:effiency}.

Based on the obtained efficiency results  in Table ~\ref{re:effiency}, CNN-based trackers showed better performances than the other two types of trackers. In particular, SiamDW \cite{zhang2019deeper} tracker achieved top efficiency results with 52.58 FPS of tracking speed and 2.46 million of parameters by using the light-weighted cropping-inside residual units based CNN backbone.  The SiamGAT \cite{guo2021graph} tracker achieved a tracking speed of 41.99 FPS. Its tracking model has 14.23 million parameters and 14.23 gigaFLOPs for a search image size of $287 \times 287$.  Although most of the recent CNN-based trackers are computationally efficient,  SiamRPN++ \cite{li2019siamrpn++} tracker achieved poor efficiency results with 5.17 FPS of tracking speed since it used a deeper CNN backbone architecture for feature extraction.  The SiamRN \cite{cheng2021learning} approach also showed the second-lowest efficiency results among the CNN-based trackers, with an average tracking speed of 6.51 FPS and 116.87 gigaFLOPs, due to its computationally expensive relation detector module. 

Overall, most of the CNN-Transformer based trackers successfully balance tracking robustness and computational efficiency. In particular, the number of FLOPs of the CNN-Transformer based trackers is considerably lower than the other two categories because they successfully capture robust cues from CNN-based features, even with a lightweight backbone network.  Particularly,  AiATrack \cite{bib73} tracker achieved 31.22 FPS of the average tracking speed and 17.95 million parameters while maintaining 69\% of success score on the LaSOT dataset. Although the AiATrack approach searches the target in a large search region, it has only 9.45 gigaFLOPs because its model update mechanism uses a feature-reusing technique to avoid additional computational costs. Among the CNN-Transformer based trackers, HiFT \cite{bib76} approach achieved top efficiency results with 37.06 FPS of the tracking speed and 11.07 million of parameters while showing a considerable tracking accuracy. Since the HiFT tracker utilized the lightweight AlexNet \cite{bib15} as the backbone feature extraction network, it achieved top efficiency results with fewer FLOPs. Based on its average tracking speed, the CSWinTT \cite{bib67} tracker showed poor results with 8.76 FPS because the cyclic shifting attention mechanism of this approach is computationally expensive.

Although most of the fully Transformer-based trackers are computationally inefficient, GRM-B \cite{gao2023generalized}, SimTrack-B/16 \cite{bib79}, and SparseTT \cite{bib77} trackers considerably balance the tracking robustness and efficiency. Since the GRM tracker searches for the target in a narrower region compared to other fully-Transformer trackers and avoids unnecessary attention computation by eliminating background patches, it successfully balances accuracy and efficiency. The reason behind the better efficiency of the SimTrack approach is its simple and computationally efficient ViT-B/16 backbone Transformer architecture.  Similarly, the SparseTT \cite{bib77} tracking model has a computational load of 9.21 gigaFLOPs since it utilizes the smallest version of the Swin Transformer \cite{bib34} as the backbone network. Also, the SparseTT tracker removed the non-similarity image pairs between target template and search image in the searching process and hence improved the tracking efficiency.

Among the fully-Transformer based approaches, SeqTrack \cite{chen2023seqtrack} achieved lower efficiency results with 5.81 FPS of tracking speed, 308.98 million parameters, and 535.85 gigaFLOPs. Although SeqTrack has shown superior accuracy in several benchmarks, it is computationally expensive since it relies on the large-scale ViT-L model. Additionally, unlike other trackers, SeqTrack uses a large size ($384 \times 384$) of the target template and search region, leading to an increase in the number of floating-point operations.  The MixFormer-L \cite{bib80} tracker also achieved low efficiency results with a tracking speed of 8.02 FPS, 195.40 million parameters, and 113.02 gigaFLOPs. The tracking model of the MixFormer-L approach was created from the wider model of CVT Transformer \cite{bib99} (CVT-W24), and hence it achieved poor results in terms of efficiency.  Although the ProContEXT \cite{lan2022procontext} tracker is a follow-up work of OSTrack \cite{bib78} tracking approach, it can able to track a target at the half speed of the OSTrack since its context-aware self-attention module is computationally expensive and hence the number of FLOPs are doubled.  

In summary, CNN-based trackers achieve better results in terms of tracking speed, number of parameters, and number of floating-point operations. However, their tracking robustness is poorer than the other two categories. On the other hand, although fully-Transformer based trackers show excellent tracking robustness, their efficiency is lower than the other two types. Number of floating point operations of the fully-Transformer based approaches are generally higher than the CNN-Transformer based and CNN-based trackers because of their attention mechanism. CNN-Transformer based trackers successfully balance tracking robustness and efficiency by combining CNN-based feature extraction and Transformer-based feature fusion.

\section{Discussions}\label{sec6}

In this section, we discuss the summary of the findings based on the literature review and experimental survey, as well as the future directions of Transformer-based single object tracking.

\subsection{Summary of Findings}\label{findings}
This survey study focused on analyzing the literature and performances of a subset of visual object trackers that use Transformers in their tracking pipeline. We covered different types of Transformer-based trackers and analyzed their individual performances based on how they addressed the tracking challenges. In addition, we compared the performances of Transformer trackers with state-of-the-art CNN-based trackers to demonstrate how they have surpassed them by a significant margin within a short period of time. 

Before the introduction of the Transformer in object tracking, CNN-based trackers dominated the tracking world. Specifically, Siamese-based approaches achieved a considerable balance between tracking robustness and efficiency in benchmark datasets by treating object tracking as a template matching problem. However, since  CNN-based Siamese tracking approaches primarily rely on the correlation operation, which is a local linear matching process, their performance is limited in challenging tracking scenarios. Additionally, based on the results of our experimental analysis on the large-scale LaSOT benchmark, the long-term tracking capability of CNN-based trackers is very limited. This limitation arises from their disregard for temporal cues in the template matching process. Furthermore, attribute-wise experimental results showed that CNN-based trackers still struggle to track a target in fully occluded, rotated, viewpoint-changed, and scale-varied scenarios due to their poor target discriminative and feature matching capabilities. While all CNN-based trackers have demonstrated poor tracking robustness in recently developed challenging datasets, they still exhibit excellent performance in the OTB100 dataset. This notable distinction arises from the fact that the targets in the OTB100 dataset do not undergo substantial appearance alterations, unlike those in the other challenging datasets.

The Transformer was initially introduced to the single-object tracking community as a module with a CNN backbone, and these approaches are referred to as CNN-Transformer based trackers in this study. Researchers replaced the correlation operation of Siamese tracking approaches with a Transformer architecture in CNN-Transformer based trackers, utilizing  attention mechanism. Based on the results of this study, CNN-Transformer based trackers successfully balanced tracking robustness and efficiency in benchmark datasets. Although their tracking robustness is good in short-term and aerial tracking sequences, their long-term tracking capabilities are considerably lower than fully-Transformer based approaches. Compared to CNN-based trackers, CNN-Transformer based approaches successfully utilize the backbone CNN architecture for feature extraction, and they show excellent performances even with smaller pre-trained CNN models. However, our attribute-wise experimental analysis reveals that CNN-Transformer based trackers exhibit limited performance in tracking scenarios involving full occlusion, low resolution, out-of-view targets, and illumination variation since their inability to capture the temporal cues and target-specific cues using CNN features.

Recent VOT approaches fully rely on Transformer architectures, leveraging their global feature learning capabilities in object tracking. Based on their model architecture, we classify the literature of fully-Transformer approaches into two categories: Two-stream Two-stage trackers and One-stream One-stage trackers. Two-stream Two-stage trackers perform feature extraction and fusion in two distinguishable stages, utilizing two identical Transformer network branches and another Transformer network, respectively. On the other hand, One-stream One-stage trackers use a single pipeline of a Transformer network. Based on our experiential evaluation on challenging benchmark datasets, fully-Transformer based trackers significantly outperform other approaches by a wide margin while maintaining moderate efficiency scores. 

Our experimental comparison study clearly shows that One-stream One-stage fully-Transformer trackers significantly outperform other types of trackers and are expected to dominate the single object tracking community for the next couple of years. Since the feature extraction and fusion are conducted together by a single Transformer network architecture in these trackers, the information flow between the features of target template and search region patches is efficiently utilized. As a result, target-specific discriminative features are enhanced, while unnecessary background and distractor features are eliminated. Therefore, the One-stream One-stage  trackers successfully handle long-term tracking scenarios, even when the target undergoes severe appearance changes and full occlusion.

Based on our experimental study, we have found that most One-stream One-stage fully-Transformer trackers exhibit poor efficiency compared to CNN-based trackers in terms of FLOPs, tracking speed, and the number of parameters. The primary reason for this is that ViT based Transformers employ a self-attention mechanism that calculates attention between all patches, resulting in a quadratic computational complexity in the input sequence. Additionally, certain trackers such as SeqTrack \cite{chen2023seqtrack}, which utilize encoders and decoders for self and cross-attention, exhibit very poor computational efficiency.    In addition to that reason, most Transformer trackers search for the target in a larger search region compared to CNN-based trackers, which results in slower tracking speeds.  

The experimental results of this study clearly indicate that the performance of recent trackers on the OTB100 dataset has reached saturation, as evident from the overall performances and attribute-wise results. As a result, some recent trackers have already observed this fact and started avoiding the evaluation of their performances on OTB100. We have also observed that while the tracking accuracy of some trackers is excellent on certain challenging benchmark datasets, their performances are only average on the GOT-10k dataset since these trackers tend to be biased towards familiar target objects. These findings justify the need for a massive tracking dataset comprising long tracking sequences and a large number of non-overlapping training and testing target object classes, which will truly evaluate the performance of future trackers.

The increasing number of Transformer-based approaches proposed this year, along with their exceptional performances, clearly indicates that Transformers have replaced CNNs in single object tracking. These trackers are expected to continue impacting the tracking community in the coming years, as they have introduced novel possibilities and perspectives, such as treating object tracking as a sequence learning problem rather than the template matching. The influence of Transformer-based language models on the field has further inspired researchers to propose similar architectures for object tracking. Moreover, recent advancements in self-supervised learning-based mask autoencoders have motivated the development of more robust trackers that effectively address data redundancy issues and capture more generalized representations. Consequently, it is evident that Transformer-based trackers are positioned to dominate the field of single object tracking in the future.

\subsection{Future Directions of Transformer Tracking}\label{future}
Our experimental survey clearly shows that One-stream One-stage fully-Transformer based trackers significantly outperform other Transformer-based trackers by a large margin. Despite their impressive performance on challenging benchmark datasets, there are still several issues that need to be addressed and require further attention in future work. Taking this into account, we provide some recommendations for future directions by exclusively considering the One-stream One-stage fully-Transformer based trackers.

\textbf{Designing a new spatio-temporal Transformer architecture for object tracking to capture the spatial and temporal cues together:} All of the current Transformer-based trackers utilize the Vision Transformer (ViT) \cite{bib33} or a variant of ViT as their backbone and fine-tune it for tracking. These Transformer backbone networks were originally designed for image recognition or object classification tasks, primarily focusing on capturing spatial relationships within individual images. However, they lack the inherent capability to capture temporal cues between a sequence of continuous reference frames, which is crucial for object tracking. This limitation stems from the fundamental difference between object classification and tracking tasks. To overcome this limitation, there is a need to design a novel spatio-temporal Transformer architecture specifically tailored for tracking tasks, which effectively captures spatial and temporal cues together. These Transformers can be  capture spatial cues by computing attention between image patches within a frame and temporal cues by computing attention between patches in adjacent frames. Similar to the recently proposed spatio-temporal Transformers in other tasks \cite{chappa2023spartan, aksan2021spatio}, a novel spatio-temporal Transformer could be employed to capture spatial-temporal cues from the continuous reference frame sequence, leading to the attainment of more robust results.

\textbf{Improving computational efficiency through utilization of lightweight Transformer architectures, quantization techniques, and feature reusing techniques:} Although fully-Transformer based trackers have shown outstanding tracking robustness, their computational efficiencies are considerably poor and hence not suitable for many real-world applications. Recently, several lightweight Transformers have been proposed \cite{luo2022towards, mehta2020delight, hatamizadeh2023fastervit}, demonstrating excellent efficiency scores while maintaining accuracy in various computer vision tasks. By utilizing these lightweight Transformers in VOT, computational costs could be improved. In addition, similar to the work of \cite{li2022patch}, utilizing proper quantization techniques could reduce the computational complexity of Transformers without compromising performance.  Moreover, the current Transformer trackers process the template frame patches in each and every frame using an encoder, resulting in increased computational costs. To address this issue, an efficient feature reusing mechanism could be employed to reduce the model's computational complexity.

\textbf{Utilizing self-supervised learning based masked autoencoder pre-trained models to enhance tracking performance:} Recent Transformer trackers \cite{bib78, lan2022procontext, gao2023generalized, chen2023seqtrack} have enhanced their tracking accuracy by utilizing self-supervised learning based masked autoencoder pre-trained models \cite{he2022masked} to initialize the tracker encoder. These pre-trained models helped the tracker extract more discriminative features by focusing on relevant features and discarding irrelevant or noisy information. In addition, the trackers that utilized these pre-trained models showed better performance for unseen target objects due to the generalization capability of the masked autoencoders. Due to these facts, self-supervised learning-based masked autoencoder pre-trained models could be further investigated as a potential future direction for object tracking. Additionally, since current masked autoencoders are primarily designed to capture spatial cues within an image for image recognition tasks, there is a need to develop a self-supervised pre-trained model for object tracking that can effectively capture both spatial and temporal cues. However, it is important to note that developing these powerful pre-trained mask autoencoder models requires significant computational resources and can be expensive for real-time tracking tasks. Therefore, it is crucial to adopt an efficient fine-tuning mechanism, similar to LoRA \cite{hu2022lora}, to reduce the computational complexity without compromising tracking performance.

\textbf{Enriching the accuracy of fully-Transformer trackers to track a small target object with less appearance cues:} The experimental results on the UAV123 dataset show that fully-Transformer based approaches struggle to track small target objects with limited appearance cues, as other types of trackers outperform them in attribute-wise comparisons. In particular, their success scores are poor when tracking small target objects in full occlusion, background clutter, and low resolution scenarios. Since the patch-level attention mechanism of the Transformer fails to capture the correct appearance cues of a small target object, fully-Transformer based trackers exhibit limited performance. Similar to the CSWinTT tracker \cite{bib67}, incorporating hierarchical architecture with window-level and target size-aware attention mechanism could improve the accuracy of fully-Transformer trackers when tracking small target objects. 

\textbf{Improving the local feature learning capability of the trackers through a target-specific patch merging technique:} Transformers, while effective at capturing long-range global structures in an image, have considerably poor local feature learning capabilities due to their division of the input image into equally sized patches, which results in the loss of information about local features such as edges and lines. To enhance the local feature learning capability of Transformers, a recent approach called Tokens-To-Token Vision Transformer (T2T-ViT) \cite{yuan2021tokens} aggregates neighboring patches into a single patch. Since many Transformer trackers \cite{bib78, lan2022procontext, bib79} employ the ViT \cite{bib33} backbone with hard divided input patches, their target localization capabilities could be enhanced by incorporating a patch merging technique similar to that employed in the T2T-ViT model. By utilizing a target-specific patch merging technique, it is possible to preserve the local features of the target and improve overall tracking accuracy.   

\textbf{Enhancing the token selection mechanism to overcome background interference and distractors:}  In Transformer tracking approaches, the target template and search region patches are converted into tokens, and the attentions between these tokens are computed using a Transformer architecture. Since most trackers fail to remove background and distractor tokens and compute attention between all tokens, their performance is reduced. Additionally, due to unnecessary attention computation, the efficiency of these trackers is also reduced.  Although some trackers \cite{bib78, lan2022procontext, gao2023generalized} incorporate a mechanism to remove background tokens, further investigation and enhancement of the token selection mechanism is warranted. A more advanced token selection technique, which incorporates information from previous frames, has the potential to significantly enhance tracking accuracy and efficiency.

\textbf{Fast motion, fully occlusion, and low resolution are the major challenges in Transformer tracking:} Based on our evaluation results on various benchmark datasets, we have identified that Transformer trackers struggle to demonstrate satisfactory performance in fast motion, severe occlusion, and low-resolution scenes. Although fully-Transformer based approaches have shown considerable improvement in these scenarios, our evaluation on the challenging LaSOT dataset reveals that their success scores are lower in frames with fast motion, full occlusion, and low resolution, with the top-performing trackers achieving only 60.1\%, 64.5\%, and 65.6\%, respectively. To handle these challenges, several approaches could be taken such as enlarging the search region with a distractor-aware mechanism could handle fast-moving targets while reducing the impact of distractor objects. Additionally, including a target re-detection scheme in occlusion scenarios could improve the tracking robustness, and including the temporal cues in tracking could handle low resolution situations.

\section{Conclusion}\label{sec7}
In this study, we conducted a survey on Transformer tracking approaches. We analyzed the literature on Transformer trackers and classified them into three types: CNN-Transformer trackers, One-stream One-stage fully-Transformer based trackers, and Two-stream Two-stage fully-Transformer based trackers. We present the literature of the 26 Transformer trackers in this paper based on how they addressed different tracking challenges.

In the second phase of this study, we experimentally evaluated the tracking robustness and computational efficiency of the Transformer tracking approaches and compared their performances with CNN-based trackers. In total, we evaluated 37 trackers in our experiments. The experimental results on challenging benchmark datasets demonstrated that One-stream One-stage fully-Transformer based trackers are the state-of-the-art approaches. Also, we found that CNN-Transformer based trackers successfully maintained a balance between robustness and efficiency. In the end, we provide future directions for Transformer tracking.

\bibliography{sn-bibliography}

\begin{thebibliography}{100}
\expandafter\ifx\csname url\endcsname\relax
  \def\url#1{\texttt{#1}}\fi
\expandafter\ifx\csname urlprefix\endcsname\relax\def\urlprefix{URL }\fi
\expandafter\ifx\csname href\endcsname\relax
  \def\href#1#2{#2} \def\path#1{#1}\fi

\bibitem{ciaparrone2020deep}
G.~Ciaparrone, F.~L. S{\'a}nchez, S.~Tabik, L.~Troiano, R.~Tagliaferri,
  F.~Herrera, Deep learning in video multi-object tracking: A survey,
  Neurocomputing 381 (2020) 61--88.
\newblock \href {https://doi.org/https://doi.org/10.1016/j.neucom.2019.11.023}
  {\path{doi:https://doi.org/10.1016/j.neucom.2019.11.023}}.

\bibitem{kalake2021analysis}
L.~Kalake, W.~Wan, L.~Hou, Analysis based on recent deep learning approaches
  applied in real-time multi-object tracking: A review, IEEE Access 9 (2021)
  32650--32671.
\newblock \href {https://doi.org/10.1109/ACCESS.2021.3060821}
  {\path{doi:10.1109/ACCESS.2021.3060821}}.

\bibitem{bib1}
S.~Jha, C.~Seo, E.~Yang, G.~P. Joshi, Real time object detection and
  trackingsystem for video surveillance system, Multimed. Tools Appl. 80 (2021)
  3981--3996.

\bibitem{bib2}
S.~Bhakar, D.~P. Bhatt, V.~S. Dhaka, Y.~K. Sarma, A review on classifications
  of tracking systems in augmented reality, J. Phys.: Conf. Ser. 2161~(1)
  (2022) 012077.

\bibitem{bib3}
C.~Premachandra, S.~Ueda, Y.~Suzuki, Detection and tracking of moving objects
  at road intersections using a 360-degree camera for driver assistance and
  automated driving, IEEE Access 8 (2020) 135652--135660.

\bibitem{bib4}
R.~Pereira, G.~Carvalho, L.~Garrote, U.~J. Nunes, Sort and {Deep-SORT} based
  multi-object tracking for mobile robotics: Evaluation with new data
  association metrics, Appl. Sci. 12~(3) (2022) 1319.

\bibitem{bib5}
R.~Chandrakar, R.~Raja, R.~Miri, U.~Sinha, A.~K.~S. Kushwaha, H.~Raja, Enhanced
  the moving object detection and object tracking for traffic surveillance
  using {RBF-FDLNN} and {CBF} algorithm, Expert Syst. Appl. 191 (2022) 116306.

\bibitem{bib6}
B.~T. Naik, M.~F. Hashmi, Z.~W. Geem, N.~D. Bokde, {DeepPlayer-Track}: Player
  and referee tracking with jersey color recognition in soccer, IEEE Access 10
  (2022) 32494--32509.

\bibitem{bib7}
M.~Wallner, D.~Steininger, V.~Widhalm, M.~Sch{\"o}rghuber, C.~Beleznai, {RGB-D}
  railway platform monitoring and scene understanding for enhanced passenger
  safety, in: Proc. Int. Conf. Pattern Recognit., Springer, 2021, pp. 656--671.

\bibitem{bib8}
Y.~Liu, C.~Sivaparthipan, A.~Shankar, Human--computer interaction based visual
  feedback system for augmentative and alternative communication, Int. J.
  Speech Technol. 25~(2) (2022) 305--314.

\bibitem{bib9}
B.~Liu, J.~Huang, L.~Yang, C.~Kulikowsk, Robust tracking using local sparse
  appearance model and k-selection, in: Proc. IEEE Conf. Comput. Vis. Pattern
  Recognit., 2011, pp. 1313--1320.

\bibitem{bib10}
S.~Zhang, X.~Wang, Human detection and object tracking based on histograms of
  oriented gradients, in: Proc. Ninth Int. Conf. natural comput., 2013, pp.
  1349--1353.

\bibitem{bib11}
L.~Bertinetto, J.~Valmadre, S.~Golodetz, O.~Miksik, P.~H. Torr, Staple:
  Complementary learners for real-time tracking, in: Proc. IEEE Conf. Comput.
  Vis. Pattern Recognit., 2016, pp. 1401--1409.

\bibitem{li2013survey}
X.~Li, W.~Hu, C.~Shen, Z.~Zhang, A.~Dick, A.~V.~D. Hengel, A survey of
  appearance models in visual object tracking, ACM Trans. Intell. Syst.
  Technol. 4~(4) (2013) 1--48.

\bibitem{online2015}
T.~Kokul, A.~Ramanan, U.~Pinidiyaarachchi, Online multi-person
  tracking-by-detection method using {ACF} and particle filter, in: Proc. IEEE
  Int. Conf. Intell. Comput. Inf. Syst., 2015, pp. 529--536.

\bibitem{zhu2021single}
K.~Zhu, X.~Zhang, G.~Chen, X.~Tan, P.~Liao, H.~Wu, X.~Cui, Y.~Zuo, Z.~Lv,
  Single object tracking in satellite videos: Deep siamese network
  incorporating an interframe difference centroid inertia motion model, Remote
  Sens. 13~(7) (2021) 1298.

\bibitem{bib13}
L.~Wang, T.~Liu, G.~Wang, K.~L. Chan, Q.~Yang, Video tracking using learned
  hierarchical features, IEEE Trans. Image Process. 24~(4) (2015) 1424--1435.

\bibitem{zhao2021deep}
H.~Zhao, G.~Yang, D.~Wang, H.~Lu, Deep mutual learning for visual object
  tracking, Pattern Recognit. 112 (2021) 107796.

\bibitem{gate2017}
T.~Kokul, C.~Fookes, S.~Sridharan, A.~Ramanan, U.~Pinidiyaarachchi, Gate
  connected convolutional neural network for object tracking, in: Proc. IEEE
  Int. Conf. Image Process., 2017, pp. 2602--2606.

\bibitem{bib15}
A.~Krizhevsky, I.~Sutskever, G.~E. Hinton, {ImageNet} classification with deep
  convolutional neural networks, Commun. ACM 60~(6) (2017) 84--90.

\bibitem{bib40}
S.~M. Marvasti-Zadeh, L.~Cheng, H.~Ghanei-Yakhdan, S.~Kasaei, Deep learning for
  visual tracking: A comprehensive survey, IEEE Trans. Intell. Transp. Syst.
  23~(5) (2022) 3943--3968.

\bibitem{walid2021real}
W.~Walid, M.~Awais, A.~Ahmed, G.~Masera, M.~Martina, Real-time implementation
  of fast discriminative scale space tracking algorithm, J. Real-Time Image
  Proc. 18~(6) (2021) 2347--2360.

\bibitem{xu2021adaptive}
T.~Xu, Z.~Feng, X.-J. Wu, J.~Kittler, Adaptive channel selection for robust
  visual object tracking with discriminative correlation filters, Int. J.
  Comput. Vis. 129~(5) (2021) 1359--1375.

\bibitem{bib19}
L.~Bertinetto, J.~Valmadre, J.~F. Henriques, A.~Vedaldi, P.~H. Torr,
  Fully-convolutional siamese networks for object tracking, in: Proc. Eur.
  Conf. Comput. Vis. Workshops, Springer, 2016, pp. 850--865.

\bibitem{bib20}
K.~Chen, W.~Tao, Once for all: a two-flow convolutional neural network for
  visual tracking, IEEE Trans. Circuits Syst. Video Technol. 28~(12) (2017)
  3377--3386.

\bibitem{bib21}
J.~Bromley, I.~Guyon, Y.~LeCun, E.~S{\"a}ckinger, R.~Shah, Signature
  verification using a "{Siamese}" time delay neural network, in: Proc.
  Advances Neural Inf. Process. Syst., Vol.~6, Morgan-Kaufmann, 1993.

\bibitem{bib22}
B.~Li, J.~Yan, W.~Wu, Z.~Zhu, X.~Hu, High performance visual tracking with
  siamese region proposal network, in: Proc. IEEE Conf. Comput. Vis. Pattern
  Recognit., 2018, pp. 8971--8980.

\bibitem{bib23}
Z.~Zhu, Q.~Wang, B.~Li, W.~Wu, J.~Yan, W.~Hu, Distractor-aware siamese networks
  for visual object tracking, in: Proc. Eur. Conf. Comput. Vis., 2018, pp.
  101--117.

\bibitem{li2019siamrpn++}
B.~Li, W.~Wu, Q.~Wang, F.~Zhang, J.~Xing, J.~Yan, {SiamRPN++}: Evolution of
  siamese visual tracking with very deep networks, in: Proc. IEEE/CVF Conf.
  Comput. Vis. Pattern Recognit., 2019, pp. 4282--4291.

\bibitem{bib25}
D.~Guo, J.~Wang, Y.~Cui, Z.~Wang, S.~Chen, {SiamCAR}: Siamese fully
  convolutional classification and regression for visual tracking, in: Proc.
  IEEE/CVF Conf. Comput. Vis. Pattern Recognit., 2020, pp. 6269--6277.

\bibitem{chen2020siamese}
Z.~Chen, B.~Zhong, G.~Li, S.~Zhang, R.~Ji, Siamese box adaptive network for
  visual tracking, in: Proc. IEEE/CVF Conf. Comput. Vis. Pattern Recognit.,
  2020, pp. 6668--6677.

\bibitem{voigtlaender2020siam}
P.~Voigtlaender, J.~Luiten, P.~H. Torr, B.~Leibe, Siam {R-CNN} visual tracking
  by re-detection, in: Proc. IEEE/CVF Conf. Comput. Vis. Pattern Recognit.,
  2020, pp. 6578--6588.

\bibitem{guo2021graph}
D.~Guo, Y.~Shao, Y.~Cui, Z.~Wang, L.~Zhang, C.~Shen, Graph attention tracking,
  in: Proc. IEEE/CVF Conf. Comput. Vis. Pattern Recognit., 2021, pp.
  9543--9552.

\bibitem{cheng2021learning}
S.~Cheng, B.~Zhong, G.~Li, X.~Liu, Z.~Tang, X.~Li, J.~Wang, Learning to filter:
  Siamese relation network for robust tracking, in: Proc. IEEE/CVF Conf.
  Comput. Vis. Pattern Recognit., 2021, pp. 4421--4431.

\bibitem{bib29}
M.~Ondra{\v{s}}ovi{\v{c}}, P.~Tar{\'a}bek, Siamese visual object tracking: A
  survey, IEEE Access 9 (2021) 110149--110172.

\bibitem{bib30}
A.~Vaswani, N.~Shazeer, N.~Parmar, J.~Uszkoreit, L.~Jones, A.~N. Gomez,
  {\L}.~Kaiser, I.~Polosukhin, Attention is all you need, in: Proc. Advances
  Neural Inf. Process. Syst., Vol.~30, Curran Associates, Inc., 2017.

\bibitem{bib33}
A.~Dosovitskiy, L.~Beyer, A.~Kolesnikov, D.~Weissenborn, X.~Zhai,
  T.~Unterthiner, M.~Dehghani, M.~Minderer, G.~Heigold, S.~Gelly, et~al.,
  \href{https://openreview.net/forum?id=YicbFdNTTy}{An image is worth 16x16
  words: Transformers for image recognition at scale}, in: Proc. Int. Conf.
  Learn. Represen., 2021.
\newline\urlprefix\url{https://openreview.net/forum?id=YicbFdNTTy}

\bibitem{bib35}
N.~Carion, F.~Massa, G.~Synnaeve, N.~Usunier, A.~Kirillov, S.~Zagoruyko,
  End-to-end object detection with transformers, in: Proc. Eur. Conf. Comput.
  Vis., Springer, 2020, pp. 213--229.

\bibitem{chen2021crossvit}
C.-F.~R. Chen, Q.~Fan, R.~Panda, {CrossViT}: Cross-attention multi-scale vision
  transformer for image classification, in: Proc. IEEE/CVF Int. Conf. Comput.
  Vis., 2021, pp. 357--366.

\bibitem{strudel2021segmenter}
R.~Strudel, R.~Garcia, I.~Laptev, C.~Schmid, Segmenter: Transformer for
  semantic segmentation, in: Proc. IEEE/CVF Int. Conf. Comput. Vis., 2021, pp.
  7262--7272.

\bibitem{guo2021pct}
M.-H. Guo, J.-X. Cai, Z.-N. Liu, T.-J. Mu, R.~R. Martin, S.-M. Hu, {PCT}: Point
  cloud transformer, Comput. Visual Media 7 (2021) 187--199.

\bibitem{han2022survey}
K.~Han, Y.~Wang, H.~Chen, X.~Chen, J.~Guo, Z.~Liu, Y.~Tang, A.~Xiao, C.~Xu,
  Y.~Xu, et~al., A survey on vision transformer, IEEE Trans. Pattern Anal.
  Mach. Intell. 45~(1) (2023) 87--110.

\bibitem{khan2022transformers}
S.~Khan, M.~Naseer, M.~Hayat, S.~W. Zamir, F.~S. Khan, M.~Shah, Transformers in
  vision: A survey, ACM Comput. Surv. 54~(10s) (2022) 1--41.

\bibitem{islam2022recent}
K.~Islam, Recent advances in vision transformer: A survey and outlook of recent
  work, arXiv preprint arXiv:2203.01536 (2022).

\bibitem{bib66}
N.~Wang, W.~Zhou, J.~Wang, H.~Li, Transformer meets tracker: Exploiting
  temporal context for robust visual tracking, in: Proc. IEEE/CVF Conf. Comput.
  Vis. Pattern Recognit., 2021, pp. 1571--1580.

\bibitem{bib37}
X.~Chen, B.~Yan, J.~Zhu, D.~Wang, X.~Yang, H.~Lu, Transformer tracking, in:
  Proc. IEEE/CVF Conf. Comput. Vis. Pattern Recognit., 2021, pp. 8126--8135.

\bibitem{bib36}
B.~Yan, H.~Peng, J.~Fu, D.~Wang, H.~Lu, Learning spatio-temporal transformer
  for visual tracking, in: Proc. IEEE/CVF Int. Conf. Comput. Vis., 2021, pp.
  10448--10457.

\bibitem{zhao2021trtr}
M.~Zhao, K.~Okada, M.~Inaba, {TrTr}: Visual tracking with transformer, arXiv
  preprint arXiv:2105.03817 (2021).

\bibitem{bib74}
B.~Yu, M.~Tang, L.~Zheng, G.~Zhu, J.~Wang, H.~Feng, X.~Feng, H.~Lu,
  High-performance discriminative tracking with transformers, in: Proc.
  IEEE/CVF Int. Conf. Comput. Vis., 2021, pp. 9856--9865.

\bibitem{bib76}
Z.~Cao, C.~Fu, J.~Ye, B.~Li, Y.~Li, {HiFT}: Hierarchical feature transformer
  for aerial tracking, in: Proc. IEEE/CVF Int. Conf. Comput. Vis., 2021, pp.
  15457--15466.

\bibitem{bib75}
C.~Mayer, M.~Danelljan, G.~Bhat, M.~Paul, D.~P. Paudel, F.~Yu, L.~Van~Gool,
  Transforming model prediction for tracking, in: Proc. IEEE/CVF Conf. Comput.
  Vis. Pattern Recognit., 2022, pp. 8731--8740.

\bibitem{xing2022siamese}
D.~Xing, N.~Evangeliou, A.~Tsoukalas, A.~Tzes, Siamese transformer pyramid
  networks for real-time {UAV} tracking, in: Proc. IEEE/CVF Winter Conf. Appl.
  Comput. Vis., 2022, pp. 2139--2148.

\bibitem{ma2022unified}
F.~Ma, M.~Z. Shou, L.~Zhu, H.~Fan, Y.~Xu, Y.~Yang, Z.~Yan, Unified transformer
  tracker for object tracking, in: Proc. IEEE/CVF Conf. Comput. Vis. Pattern
  Recognit., 2022, pp. 8781--8790.

\bibitem{zhong2022correlation}
M.~Zhong, F.~Chen, J.~Xu, G.~Lu, Correlation-based transformer tracking, in:
  Proc. Int. Conf. Artif. Neural Networks, Springer, 2022, pp. 85--96.

\bibitem{bib67}
Z.~Song, J.~Yu, Y.-P.~P. Chen, W.~Yang, Transformer tracking with cyclic
  shifting window attention, in: Proc. IEEE/CVF Conf. Comput. Vis. Pattern
  Recognit., 2022, pp. 8791--8800.

\bibitem{bib73}
S.~Gao, C.~Zhou, C.~Ma, X.~Wang, J.~Yuan, {AiATrack}: Attention in attention
  for transformer visual tracking, in: Proc. Eur. Conf. Comput. Vis., Springer,
  2022, pp. 146--164.

\bibitem{xie2021learning}
F.~Xie, C.~Wang, G.~Wang, W.~Yang, W.~Zeng, Learning tracking representations
  via dual-branch fully transformer networks, in: Proc. IEEE/CVF Int. Conf.
  Comput. Vis., 2021, pp. 2688--2697.

\bibitem{bib38}
L.~Lin, H.~Fan, Z.~Zhang, Y.~Xu, H.~Ling, {SwinTrack}: A simple and strong
  baseline for transformer tracking, in: Proc. Advances Neural Inf. Process.
  Syst., Vol.~35, Curran Associates, Inc., 2022, pp. 16743--16754.

\bibitem{bib77}
Z.~Fu, Z.~Fu, Q.~Liu, W.~Cai, Y.~Wang, {SparseTT}: Visual tracking with sparse
  transformers, in: Proc. 31st Int. Joint Conf. Artif. Intell., 2022, pp.
  905--912.

\bibitem{bib80}
Y.~Cui, C.~Jiang, L.~Wang, G.~Wu, {MixFormer}: End-to-end tracking with
  iterative mixed attention, in: Proc. IEEE/CVF Conf. Comput. Vis. Pattern
  Recognit., 2022, pp. 13608--13618.

\bibitem{bib79}
B.~Chen, P.~Li, L.~Bai, L.~Qiao, Q.~Shen, B.~Li, W.~Gan, W.~Wu, W.~Ouyang,
  Backbone is all your need: a simplified architecture for visual object
  tracking, in: Proc. Eur. Conf. Comput. Vis., Springer, 2022, pp. 375--392.

\bibitem{bib78}
B.~Ye, H.~Chang, B.~Ma, S.~Shan, X.~Chen, Joint feature learning and relation
  modeling for tracking: A one-stream framework, in: Proc. Eur. Conf. Comput.
  Vis., Springer, 2022, pp. 341--357.

\bibitem{lan2022procontext}
J.-P. Lan, Z.-Q. Cheng, J.-Y. He, C.~Li, B.~Luo, X.~Bao, W.~Xiang, Y.~Geng,
  X.~Xie, {ProContEXT}: Exploring progressive context transformer for tracking,
  in: IEEE Int. Conf. on Acoust Speech Signal Process.(ICASSP), 2023, pp. 1--5.

\bibitem{yang2023bandt}
K.~Yang, H.~Zhang, J.~Shi, J.~Ma, {BANDT}: A border-aware network with
  deformable transformers for visual tracking, IEEE Trans. Consum. Electron.
  (2023).
\newblock \href {https://doi.org/10.1109/TCE.2023.3251407}
  {\path{doi:10.1109/TCE.2023.3251407}}.

\bibitem{wu2023dropmae}
Q.~Wu, T.~Yang, Z.~Liu, B.~Wu, Y.~Shan, A.~B. Chan, {DropMAE}: Masked
  autoencoders with spatial-attention dropout for tracking tasks, in: Proc.
  IEEE/CVF Conf. Comput. Vis. Pattern Recognit., 2023, pp. 14561--14571.

\bibitem{xie2023videotrack}
F.~Xie, L.~Chu, J.~Li, Y.~Lu, C.~Ma, {VideoTrack}: Learning to track objects
  via video transformer, in: Proc. IEEE/CVF Conf. Comput. Vis. Pattern
  Recognit., 2023, pp. 22826--22835.

\bibitem{gao2023generalized}
S.~Gao, C.~Zhou, J.~Zhang, Generalized relation modeling for transformer
  tracking, in: Proc. IEEE/CVF Conf. Comput. Vis. Pattern Recognit., 2023, pp.
  18686--18695.

\bibitem{wei2023autoregressive}
X.~Wei, Y.~Bai, Y.~Zheng, D.~Shi, Y.~Gong, Autoregressive visual tracking, in:
  Proc. IEEE/CVF Conf. Comput. Vis. Pattern Recognit., 2023, pp. 9697--9706.

\bibitem{chen2023seqtrack}
X.~Chen, H.~Peng, D.~Wang, H.~Lu, H.~Hu, {SeqTrack}: Sequence to sequence
  learning for visual object tracking, in: Proc. IEEE/CVF Conf. Comput. Vis.
  Pattern Recognit., 2023, pp. 14572--14581.

\bibitem{bib39}
X.~Li, W.~Hu, C.~Shen, Z.~Zhang, A.~Dick, A.~V.~D. Hengel, A survey of
  appearance models in visual object tracking, ACM Trans. Intell. Syst.
  Technol. 4~(4) (2013) 1--48.

\bibitem{bib41}
P.~Li, D.~Wang, L.~Wang, H.~Lu, Deep visual tracking: Review and experimental
  comparison, Pattern Recognit. 76 (2018) 323--338.

\bibitem{bib53}
A.~Yilmaz, O.~Javed, M.~Shah, Object tracking: A survey, ACM Comput. Surv.
  38~(4) (2006) 13--es.

\bibitem{bib54}
R.~Pflugfelder, An in-depth analysis of visual tracking with siamese neural
  networks, arXiv preprint arXiv:1707.00569 (2017).

\bibitem{bib56}
B.~Deori, D.~M. Thounaojam, A survey on moving object tracking in video, Int.
  J. Inf. Theory 3~(3) (2014) 31--46.

\bibitem{bib57}
Z.~Soleimanitaleb, M.~A. Keyvanrad, A.~Jafari, Object tracking methods: A
  review, in: 9th Int. Conf. Comput. Knowl. Eng., 2019, pp. 282--288.

\bibitem{bib58}
Y.~Zhang, T.~Wang, K.~Liu, B.~Zhang, L.~Chen, Recent advances of single-object
  tracking methods: A brief survey, Neurocomputing 455 (2021) 1--11.

\bibitem{bib59}
Y.~Wu, J.~Lim, M.-H. Yang, Object tracking benchmark, IEEE Trans. Pattern Anal.
  Mach. Intell. 37~(9) (2015) 1834--1848.
\newblock \href {https://doi.org/10.1109/TPAMI.2014.2388226}
  {\path{doi:10.1109/TPAMI.2014.2388226}}.

\bibitem{smeulders2013visual}
A.~W. Smeulders, D.~M. Chu, R.~Cucchiara, S.~Calderara, A.~Dehghan, M.~Shah,
  Visual tracking: An experimental survey, IEEE Trans. Pattern Anal. Mach.
  Intell. 36~(7) (2013) 1442--1468.

\bibitem{cannons2008review}
K.~Cannons, A review of visual tracking, Dept. Comput. Sci. Eng. York Univ.,
  Toronto, ON, Canada, Tech. Rep. CSE-2008-07, 242 (2008).

\bibitem{chen2022visual}
F.~Chen, X.~Wang, Y.~Zhao, S.~Lv, X.~Niu, Visual object tracking: A survey,
  Comput. Vis. Image Understanding 222 (2022) 103508.

\bibitem{javed2022visual}
S.~Javed, M.~Danelljan, F.~S. Khan, M.~H. Khan, M.~Felsberg, J.~Matas, Visual
  object tracking with discriminative filters and siamese networks: A survey
  and outlook, IEEE Trans. Pattern Anal. Mach. Intell. 45~(5) (2022)
  6552--6574.

\bibitem{bib42}
H.~Fan, L.~Lin, F.~Yang, P.~Chu, G.~Deng, S.~Yu, H.~Bai, Y.~Xu, C.~Liao,
  H.~Ling, {LaSOT}: A high-quality benchmark for large-scale single object
  tracking, in: Proc. IEEE/CVF Conf. Comput. Vis. Pattern Recognit., 2019, pp.
  5374--5383.

\bibitem{bib62}
L.~Huang, X.~Zhao, K.~Huang, {GOT-10k}: A large high-diversity benchmark for
  generic object tracking in the wild, IEEE Trans. Pattern Anal. Mach. Intell.
  43~(5) (2019) 1562--1577.

\bibitem{bib72}
M.~Muller, A.~Bibi, S.~Giancola, S.~Alsubaihi, B.~Ghanem, {TrackingNet}: A
  large-scale dataset and benchmark for object tracking in the wild, in: Proc.
  Eur. Conf. Comput. Vis., 2018, pp. 300--317.

\bibitem{mueller2016benchmark}
M.~Mueller, N.~Smith, B.~Ghanem, A benchmark and simulator for {UAV} tracking,
  in: Proc. Eur. Conf. Comput. Vis., Springer, 2016, pp. 445--461.

\bibitem{bib61}
M.~Kristan, et~al., The visual object tracking {VOT2015} challenge results, in:
  Proc. IEEE Int. Conf. Comput. Vis. Workshops, 2015, pp. 1--23.

\bibitem{VOT2016}
S.~Hadfield, R.~Bowden, K.~Lebeda, The visual object tracking {VOT2016}
  challenge results, Lecture Notes in Computer Science 9914 (2016) 777--823.

\bibitem{VOT2017}
M.~Kristan, A.~Leonardis, J.~Matas, M.~Felsberg, R.~Pflugfelder, et~al., The
  visual object tracking {VOT2017} challenge results, in: Proc. IEEE/CVF Int.
  Conf. Comput. Vis. Workshops, 2017, pp. 1949--1972.
\newblock \href {https://doi.org/10.1109/ICCVW.2017.230}
  {\path{doi:10.1109/ICCVW.2017.230}}.

\bibitem{kristan2018sixth}
M.~Kristan, et~al., The sixth visual object tracking {VOT2018} challenge
  results, in: Proc. Eur. Conf. Comput. Vis. Workshops, 2018, pp. 1--52.

\bibitem{hendrycks2016gaussian}
D.~Hendrycks, K.~Gimpel, Gaussian error linear units ({GELUs}), arXiv preprint
  arXiv:1606.08415 (2016).

\bibitem{bib34}
Z.~Liu, Y.~Lin, Y.~Cao, H.~Hu, Y.~Wei, Z.~Zhang, S.~Lin, B.~Guo, {Swin
  Transformer}: Hierarchical vision transformer using shifted windows, in:
  Proc. IEEE/CVF Int. Conf. Comput. Vis., 2021, pp. 10012--10022.

\bibitem{bib99}
H.~Wu, B.~Xiao, N.~Codella, M.~Liu, X.~Dai, L.~Yuan, L.~Zhang, {CvT}:
  Introducing convolutions to vision transformers, in: Proc. IEEE/CVF Int.
  Conf. Comput. Vis., 2021, pp. 22--31.

\bibitem{bib97}
X.~Chu, Z.~Tian, B.~Zhang, X.~Wang, C.~Shen,
  \href{https://openreview.net/forum?id=3KWnuT-R1bh}{Conditional positional
  encodings for vision transformers}, in: Proc. 11th Int. Conf. Learn.
  Represen., 2023.
\newline\urlprefix\url{https://openreview.net/forum?id=3KWnuT-R1bh}

\bibitem{bib98}
K.~Han, A.~Xiao, E.~Wu, J.~Guo, C.~Xu, Y.~Wang, Transformer in transformer, in:
  Proc. Advances Neural Inf. Process. Syst., Vol.~34, 2021, pp. 15908--15919.

\bibitem{yuan2022volo}
L.~Yuan, Q.~Hou, Z.~Jiang, J.~Feng, S.~Yan, {VOLO}: Vision outlooker for visual
  recognition, IEEE Trans. Pattern Anal. Mach. Intell. 45~(5) (2023)
  6575--6586.

\bibitem{zhang2019deeper}
Z.~Zhang, H.~Peng, Deeper and wider siamese networks for real-time visual
  tracking, in: Proc. IEEE/CVF Conf. Comput. Vis. Pattern Recognit., 2019, pp.
  4591--4600.

\bibitem{bhat2019learning}
G.~Bhat, M.~Danelljan, L.~V. Gool, R.~Timofte, Learning discriminative model
  prediction for tracking, in: Proc. IEEE/CVF Int. Conf. Comput. Vis., 2019,
  pp. 6182--6191.

\bibitem{yu2020deformable}
Y.~Yu, Y.~Xiong, W.~Huang, M.~R. Scott, Deformable siamese attention networks
  for visual object tracking, in: Proc. IEEE/CVF Conf. Comput. Vis. Pattern
  Recognit., 2020, pp. 6728--6737.

\bibitem{xu2020siamfc++}
Y.~Xu, Z.~Wang, Z.~Li, Y.~Yuan, G.~Yu, {SiamFC++}: Towards robust and accurate
  visual tracking with target estimation guidelines, in: Proc. AAAI Conf.
  Artif. Intell., Vol.~34, 2020, pp. 12549--12556.

\bibitem{zhang2020ocean}
Z.~Zhang, H.~Peng, J.~Fu, B.~Li, W.~Hu, Ocean: Object-aware anchor-free
  tracking, in: Proc. Eur. Conf. Comput. Vis., Springer, 2020, pp. 771--787.

\bibitem{danelljan2020probabilistic}
M.~Danelljan, L.~V. Gool, R.~Timofte, Probabilistic regression for visual
  tracking, in: Proc. IEEE/CVF Int. Conf. Comput. Vis., 2020, pp. 7183--7192.

\bibitem{mayer2021learning}
C.~Mayer, M.~Danelljan, D.~P. Paudel, L.~Van~Gool, Learning target candidate
  association to keep track of what not to track, in: Proc. IEEE/CVF Int. Conf.
  Comput. Vis., 2021, pp. 13444--13454.

\bibitem{Zhao2023CVPR}
H.~Zhao, D.~Wang, H.~Lu, Representation learning for visual object tracking by
  masked appearance transfer, in: Proc. IEEE/CVF Conf. Comput. Vis. Pattern
  Recognit., 2023, pp. 18696--18705.

\bibitem{lin2014microsoft}
T.-Y. Lin, M.~Maire, S.~Belongie, J.~Hays, P.~Perona, D.~Ramanan,
  P.~Doll{\'a}r, C.~L. Zitnick, Microsoft {COCO}: Common objects in context,
  in: Proc. Eur. Conf. Comput. Vis., Springer, 2014, pp. 740--755.

\bibitem{real2017youtube}
E.~Real, J.~Shlens, S.~Mazzocchi, X.~Pan, V.~Vanhoucke,
  {YouTube-BoundingBoxes}: A large high-precision human-annotated data set for
  object detection in video, in: Proc. IEEE/CVF Conf. Comput. Vis. Pattern
  Recognit., 2017, pp. 5296--5305.

\bibitem{zhudeformable}
X.~Zhu, W.~Su, L.~Lu, B.~Li, X.~Wang, J.~Dai,
  \href{https://openreview.net/forum?id=gZ9hCDWe6ke}{Deformable {DETR}:
  Deformable transformers for end-to-end object detection}, in: Proc. Int.
  Conf. Learn. Represen., 2021.
\newline\urlprefix\url{https://openreview.net/forum?id=gZ9hCDWe6ke}

\bibitem{he2016deep}
K.~He, X.~Zhang, S.~Ren, J.~Sun, Deep residual learning for image recognition,
  in: Proc. IEEE/CVF Conf. Comput. Vis. Pattern Recognit., 2016, pp. 770--778.

\bibitem{hu2022transformer}
X.~Hu, H.~Liu, Y.~Hui, X.~Wu, J.~Zhao, Transformer feature enhancement network
  with template update for object tracking, Sensors 22~(14) (2022) 5219.

\bibitem{he2022masked}
K.~He, X.~Chen, S.~Xie, Y.~Li, P.~Doll{\'a}r, R.~Girshick, Masked autoencoders
  are scalable vision learners, in: Proc. IEEE/CVF Conf. Comput. Vis. Pattern
  Recognit., 2022, pp. 16000--16009.

\bibitem{bib71}
Y.~Wu, J.~Lim, M.-H. Yang, Online object tracking: A benchmark, in: Proc.
  IEEE/CVF Conf. Comput. Vis. Pattern Recognit., 2013, pp. 2411--2418.

\bibitem{chappa2023spartan}
N.~V. Chappa, P.~Nguyen, A.~H. Nelson, H.-S. Seo, X.~Li, P.~D. Dobbs, K.~Luu,
  Spartan: Self-supervised spatiotemporal transformers approach to group
  activity recognition, in: Proc. IEEE/CVF Conf. Comput. Vis. Pattern
  Recognit., 2023, pp. 5157--5167.

\bibitem{aksan2021spatio}
E.~Aksan, M.~Kaufmann, P.~Cao, O.~Hilliges, A spatio-temporal transformer for
  3d human motion prediction, in: Int. Conf. 3D Vision (3DV), IEEE, 2021, pp.
  565--574.

\bibitem{luo2022towards}
G.~Luo, Y.~Zhou, X.~Sun, Y.~Wang, L.~Cao, Y.~Wu, F.~Huang, R.~Ji, Towards
  lightweight transformer via group-wise transformation for vision-and-language
  tasks, IEEE Trans. Image Process. 31 (2022) 3386--3398.

\bibitem{mehta2020delight}
S.~Mehta, M.~Ghazvininejad, S.~Iyer, L.~Zettlemoyer, H.~Hajishirzi,
  \href{https://openreview.net/forum?id=ujmgfuxSLrO}{Delight: Deep and
  light-weight transformer}, in: Proc. Int. Conf. Learn. Represen., 2021.
\newline\urlprefix\url{https://openreview.net/forum?id=ujmgfuxSLrO}

\bibitem{hatamizadeh2023fastervit}
A.~Hatamizadeh, G.~Heinrich, H.~Yin, A.~Tao, J.~M. Alvarez, J.~Kautz,
  P.~Molchanov, Fastervit: Fast vision transformers with hierarchical
  attention, arXiv preprint arXiv:2306.06189 (2023).

\bibitem{li2022patch}
Z.~Li, L.~Ma, M.~Chen, J.~Xiao, Q.~Gu, Patch similarity aware data-free
  quantization for vision transformers, in: Proc. Eur. Conf. Comput. Vis.,
  Springer, 2022, pp. 154--170.

\bibitem{hu2022lora}
E.~J. Hu, yelong shen, P.~Wallis, Z.~Allen-Zhu, Y.~Li, S.~Wang, L.~Wang,
  W.~Chen, \href{https://openreview.net/forum?id=nZeVKeeFYf9}{Lo{RA}: Low-rank
  adaptation of large language models}, in: Proc. Int. Conf. Learn. Represen.,
  2022.
\newline\urlprefix\url{https://openreview.net/forum?id=nZeVKeeFYf9}

\bibitem{yuan2021tokens}
L.~Yuan, Y.~Chen, T.~Wang, W.~Yu, Y.~Shi, Z.-H. Jiang, F.~E. Tay, J.~Feng,
  S.~Yan, Tokens-to-token vit: Training vision transformers from scratch on
  imagenet, in: Proc. IEEE/CVF Int. Conf. Comput. Vis., 2021, pp. 558--567.

\end{thebibliography}

\end{document}